\documentclass{article}

\usepackage{arxiv}
\usepackage{epstopdf}
\usepackage[utf8]{inputenc} 
\usepackage[T1]{fontenc}    
\usepackage{hyperref}       
\usepackage{url}            
\usepackage{booktabs}       
\usepackage{amsfonts}       
\usepackage{nicefrac}       
\usepackage{microtype}      
\usepackage{lipsum}

\usepackage{epsfig}
\usepackage{amssymb}
\usepackage{subfigure,float}
\usepackage[numbers, sort&compress]{natbib}

\title{A Generative Model based Adversarial Security of Deep Learning and Linear Classifier Models}

\author{
  Ferhat Ozgur Catak\\
  Simula Research laboratory\\
  Oslo, Norway \\
  \texttt{ozgur@simula.no} \\
   \And
 Samed Sivaslioglu \\
  TUBITAK BILGEM, Kocaeli, Turkey\\
  \texttt{samedsivaslioglu@gmail.com} \\
  \AND
  Kevser Sahinbas \\
  Department of Management Information System \\
  Istanbul Medipol University \\
  Istanbul, Turkey \\
   \texttt{ksahinbas@medipol.edu.tr} \\
}

\begin{document}
\maketitle

\begin{abstract}
In recent years, machine learning algorithms have been applied widely in various fields such as health, transportation, and the autonomous car. With the rapid developments of deep learning techniques, it is critical to take the security concern into account for the application of the algorithms. While machine learning offers significant advantages in terms of the application of algorithms, the issue of security is ignored. Since it has many applications in the real world, security is a vital part of the algorithms. In this paper, we have proposed a mitigation method for adversarial attacks against machine learning models with an autoencoder model that is one of the generative ones. The main idea behind adversarial attacks against machine learning models is to produce erroneous results by manipulating trained models. We have also presented the performance of autoencoder models to various attack methods from deep neural networks to traditional algorithms by using different methods such as non-targeted and targeted attacks to multi-class logistic regression, a fast gradient sign method, a targeted fast gradient sign method and a basic iterative method attack to neural networks for the MNIST dataset.
\end{abstract}


\section{Introduction}
With the help of artificial intelligence technology, machine learning has been widely used in classification, decision making, voice and face recognition, games, financial assessment, and other fields \cite{paper18, paper17}. The machine learning methods consider player’s choices in the animation industry for games and analyze diseases to contribute to the decision-making mechanism \cite{paper20,paper19,paper21, paper31}. With the successful implementations of machine learning, attacks on the machine learning process and counter-attack methods and incrementing robustness of learning have become hot research topics in recent years \cite{paper22,paper23,paper24,paper41,paper42}. The presence of negative data samples or an attack on the model can lead to producing incorrect results in the predictions and classifications even in the advanced models. 

It is more challenging to recognize the attack because of using big data in machine learning applications compared to other cybersecurity fields. Therefore, it is essential to create components for machine learning that are resistant to this type of attack. In contrast, recent works have conducted in this area and demonstrated that the resistance is not very robust to attacks \cite{paper1,paper2}. These methods have shown success against a specific set of attack methods and have generally failed to provide complete and generic protection\cite{paper3}.

Previous methods have shown success against a specific set of attack methods and have generally failed to provide complete and generic protection [14]. This field has been spreading rapidly, and, in this field, lots of dangers have attracted increasing attention from escaping the filters of unwanted and phishing e-mails, to poisoning the sensor data of a car or aircraft that drives itself \cite{paper25,paper26}. Disaster scenarios can occur if any precautions are not taken in these systems \cite{paper27}.

The main contribution of this work is to explore the autoencoder based generative models against adversarial machine learning attacks to the models. Adversarial Machine Learning has been used to study these attacks and reduce their effects \cite{paper4,paper40}. Previous works point out the fundamental equilibrium to design the algorithms and to create new algorithms and methods that are resistant and robust against attacks that will negatively affect this balance. However, most of these works have been implemented successfully for specific situations. In Section \ref{sec:prelim}, we present some applications of these works. 

This work aims to propose a method that not only presents a generic resistance to specific attack methods but also provides robustness to machine learning models in general. Our goal is to find an effective method that can be used by model trainers. For this purpose, we have processed the data with autoencoder before reaching to the machine learning model.  In our previous works \citep{ourpaper1,ourpaper2} we applied generative model based mitigation approach for the deep learning model attacks. 

We have used non-targeted and targeted attacks to multiclass logistic regression machine learning models for observing the change and difference between attack methods as well as various attack methods to neural networks such as fast gradient sign method (FGSM), targeted fast gradient sign method (T-FGSM) and basic iterative method (BIM). We have selected MNIST dataset that consists of numbers from people’s handwriting to provide people to understand and see changes in the data.

The study is organized as follows. In Section \ref{sec:relworks}, we first present the related works. In Section \ref{sec:prelim}, we introduce several adversarial attack types, environments, and autoencoder. In Section \ref{sec:model}, we present selection of autoencoder model, activation function and tuning parameters. In Section \ref{sec:experimentswithmnist}, we provide some observation on the robustness of autoencoder for adversarial machine learning with different machine learning algorithms and models. In Section \ref{sec:conclusion}, we conclude this study.

\section{Related Work}\label{sec:relworks}
In recent years, with the increase of the machine learning attacks, various studies have been proposed to create defensive measures against these attacks. Data sterility and learning endurance are recommended as countermeasures in defining a machine learning process \cite{paper4}. Most of the studies in these fields have been focused on specific adversarial attacks and generally, presented the theoretical discussion of adversarial machine learning area \cite{paper28,paper29}. 

Bo Li and Yevgeniy Vorobeychik present binary domains and classifications. In their work, the approach starts with mixed-integer linear programming (MILP) with constraint generation and gives suggestions on top of this. They also use the Stackelberg game multi-adversary model algorithm and the other algorithm that feeds back the generated adversarial examples to the training model, which is called as RAD (Retraining with Adversarial Examples) \cite{paper5}. On the other hand, their work is particular and works only in specific methods, even though it is presented as a general protection method. They have proposed a method that implements successful results. Similarly, Xiao et al. provide a method to increase the speed of resistance training against the rectified linear unit (RELU) \cite{paper6}. They use weight sparsity and RELU stability for robust verification. It can be said that their methodology does not provide a general approach.

Yu et al. propose a study that can evaluate the neural network's features under hostile attacks. In their study, the connection between the input space and hostile examples is presented. Also, the connection between the network strength and the decision surface geometry as an indicator of the hostile strength of the neural network is shown. By extending the loss surface to decision surface and other various methods, they provide adversarial robustness by decision surface. The geometry of the decision surface cannot be demonstrated most of the time, and there is no explicit decision boundary between correct or wrong prediction. Robustness can be increased by constructing a good model, but it can change with attack intensity \cite{paper7}.

Mardy et al. investigate artificial neural networks resistant with adversity and increase accuracy rates with different methods, mainly with optimization and prove that there can be more robust machine learning models \cite{paper3}.

Pinto et al. provide a method to solve this problem with the supported learning method. In their study, they formulate learning as a zero-sum, minimax objective function. They present machine learning models that are more resistant to disturbances are hard to model during the training and are better affected by changes in training and test conditions. They generalize reinforced learning on machine learning models. They propose a "Robust Adversarial Reinforced Learning" (RARL), where they train an agent to operate in the presence of a destabilizing adversary that applies disturbance forces to the system. However, in their work, Robust Adversarial Reinforced Learning may overfit itself, and sometimes it can miss predicting without any adversarial being in presence \cite{paper8}.

Carlini and Wagner propose a model that the self-logic and the strength of the machine learning model with a strong attack can be affected. They prove that these types of attacks can often be used to evaluate the effectiveness of potential defenses. They propose defensive distillation as a general-purpose procedure to increase robustness \cite{paper1}. 

Harding et al. similarly investigate the effects of hostile samples produced from targeted and non-targeted attacks in decision making. They provide that non-targeted samples are more effective than targeted samples in human perception and categorization of decisions \cite{paper9}.

Bai et al. present a convolutional autoencoder model with the adversarial decoders to automate the generation of adversarial samples. They produce adversary examples by a convolutional autoencoder model. They use pooling computations and sampling tricks to achieve these results. After this process, an adversarial decoder automates the generation of adversarial samples. Adversarial sampling is useful, but it cannot provide adversarial robustness on its own, and sampling tricks are too specific \cite{paper12}.

Sahay et al. propose FGSM attack and use an autoencoder to denoise the test data.  They have also used an autoencoder to denoise the test data, which is trained with both corrupted and healthy data. Then they reduce the dimension of the denoised data. These autoencoders are specifically designed to compress data effectively and reduce dimensions. Hence, it may not be wholly generalized, and training with corrupted data requires a lot of adjustments to get better test results \cite{paper13}. 

I-Ting Chen et al. also provide with FGSM attack on denoising autoencoders. They analyze the attacks from the perspective that attacks can be applied stealthily. They use autoencoders to filter data before applied to the model and compare it with the model without an autoencoder filter. They use autoencoders mainly focused on the stealth aspect of these attacks and used them specifically against FGSM with specific parameters \cite{paper15}. 

Gondim-Ribeiro et al. propose autoencoders attacks. In their work, they attack 3 types of autoencoders: Simple variational autoencoders, convolutional variational autoencoders, and DRAW (Deep Recurrent AttentiveWriter). They propose to scheme an attack on autoencoders. As they accept that "No attack can both convincingly reconstruct the target while keeping the distortions on the input imperceptible.". This method cannot be used to achieve robustness against adversarial attacks \cite{paper14}.

Table \ref{tbl:relwork} shows the strength and the weakness of the each paper.


\begin{table*}[htbp!] 
	\scriptsize
	\label{tbl:relwork}
	\caption{Related Work Summary} 
	\begin{tabular}{|p{4cm}|p{6cm}|p{6cm}|} 
		\hline  
		\textbf{Research Study}& \textbf{Strength} & \textbf{Weakness} \\ \hline  
		Adversarial Machine Learning \cite{paper4} & Introduces the emerging field of Adversarial Machine Learning. & Discusses the countermeasures against attacks without suggesting a method. \\ \hline 	
		Evasion-Robust \newline Classification on Binary Domains \cite{paper5} & Demonstrates some methods that can be used on Binary Domains, which are based on MILP. &  Very specific about the robustness, even though it is presented as a general method. \\ \hline  
		Training for Faster Adversarial Robustness Verification via Inducing ReLU \newline Stability \cite{paper6} & Using weight sparsity and RELU stability for robust verification. & Does not provide a general approach, or universality as it is suggested in paper. \\ \hline  
		Interpreting Adversarial Robustness: A View from Decision Surface in Input Space \cite{paper7} & By extending the loss surface to decision surface and other various methods, they provide adversarial robustness by decision surface. & The geometry of the decision surface cannot be shown most of the times and there is no explicit decision boundary between correct or wrong prediction. Robustness can be increased by constructing a good model but it can change with attack intensity. \\ \hline  
		Robust Adversarial \newline Reinforcement Learning \cite{paper8} & They have tried to generalize reinforced learning on machine learning models. They suggested a Robust Adversarial Reinforced Learning (RARL) where they have trained an agent to operate in the presence of a destabilizing adversary that applies disturbance forces to the system. & Robust Adversarial Reinforced Learning may overfit itself and sometimes it may mispredict without any adversarial being in presence. \\ \hline  
		Alleviating Adversarial Attacks via Convolutional Autoencoder \cite{paper12}& They have produced adversary examples via a convolutional autoencoder model. Pooling computations and sampling tricks are used. Then an adversarial decoder automate the generation of adversarial samples. & Adversarial sampling is useful but it cannot provide adversarial robustness on its own. Sampling tricks are also too specified. \\ \hline  
		Combatting Adversarial Attacks through Denoising and Dimensionality Reduction: A Cascaded Autoencoder Approach \cite{paper13} & They have used an autoencoder to denoise the test data which is trained with both corrupted and normal data. Then they reduce the dimension of the denoised data. & Autoencoders specifically designed to compress data effectively and reduce dimensions. Therefore it may not be completely generalized and training with corrupted data requires a lot of adjustments for test results. \\ \hline 
		A Comparative Study of Autoencoders against Adversarial Attacks \cite{paper15} & They have used autoencoders to  filter data before applying into the model and compare it with the model without autoencoder filter. & They have used autoencoders mainly focused on the stealth aspect of these attacks and use them specifically against FGSM with specific parameters. \\ \hline 	
		Adversarial Attacks on Variational Autoencoders \cite{paper14}
		& They propose a scheme to attack on autoencoders and validate experiments to three autoencoder models: Simple, convolutional and DRAW (Deep Recurrent Attentive Writer). 
		& As they have accepted "No attack can both convincingly reconstruct the target while keeping the distortions on the input imperceptible.". it cannot provide robustness against  adversarial attacks. \\ \hline
		Understanding Autoencoders with Information Theoretic Concepts \cite{paper10}
		&  They examine data processing inequality with stacked autoencoders and two types of information planes with autoencoders. They have analyzed DNNs learning from a joint geometric and information theoretic perspective, thus emphasizing the role that pair-wise mutual information plays important role in understanding DNNs with autoencoders.  
		& The accurate and tractable estimation of information quantities from large data seems to be a problem due to Shannon's definition and other information theories are hard to estimate, which severely limits its powers to analyze machine learning algorithms. \\ \hline  
		Adversarial Attacks and Defences Competition \cite{paper11}
		& Google Brain organized NIPS 2017 to accelerate research on adversarial examples and robustness of machine learning classifiers.  Alexey Kurakin and Ian Goodfellow et al. present some of the structure and organization of the competition and the solutions developed by several of the top-placing teams. 
		& We experimented with the proposed methods of this competition bu these methods do not  provide a generalized solution for the  robustness against adversarial machine learning model attacks.  \\ \hline
		Explaining And \newline Harnessing Adversarial Examples \cite{paper16}
		& Ian Goodfellow et al. makes considerable observations about Gradient-based optimization and introduce FGSM.
		& Models may mislead for the efficiency of optimization. The paper focuses explicitly on identifying similar types of problematic points in the model. \\ \hline
	\end{tabular}
\end{table*}

\section{Preliminaries} \label{sec:prelim}
In this section, we consider attack types, data poisoning attacks, model attacks, attack environments, and autoencoder. 
\subsection{Attack Types}

Machine Learning attacks can be categorized into data poisoning attacks and model attacks. The difference between the two attacks lies in the influencing type. Data poisoning attacks mainly focus on influencing the data, while model evasion attacks influencing the model for desired attack outcomes. Both attacks aim to disrupt the machine learning structure, evasion from filters, causing wrong predictions, misdirection, and other problems for the machine learning process. In this paper, we mainly focus on machine learning model attacks.

\subsubsection{Data Poisoning Attacks}

According to machine learning methods, algorithms are trained and tested with datasets. Data poisoning in machine learning algorithms has a significant impact on a dataset and can cause problems for algorithm and confusion for developers. With poisoning the data, adversaries can compromise the whole machine learning process. Hence, data poisoning can cause problems in machine learning algorithms. 

\subsubsection{Model Attacks}

Machine learning model attacks have been applied mostly in adversarial attacks, and evasion attacks being have been used most extensively in this category. For spam emails, phishing attacks, and executing malware code, adversaries apply model evasion attacks. There are also some benefits to adversaries in misclassification and misdirection. In this type of attack, the attacker does not change training data but disrupts or changes its data and diverse this data from the training dataset or make this data seem safe. This study mainly concentrates on model attacks.

\subsection{Attack Environments}
There are two significant threat models for adversarial attacks: the white-box and black-box models.

\subsubsection{White Box Attacks}

Under the white-box setting, the internal structure, design, and application of the tested item are accessible to the adversaries. In this model, attacks are based on an analysis of the internal structure. It is also known as open box attacks. Programming knowledge and application knowledge are essential. White-box tests provide a comprehensive assessment of both internal and external vulnerabilities and are the best choice for computational tests. 

\subsubsection{Black Box Attacks}

In the black-box model, internal structure and software testing are secrets to the adversaries. It is also known as behavioral attacks. In these tests, the internal structure does not have to be known by the tester. They provide a comprehensive assessment of errors. Without changing the learning process, black box attacks provide changes to be observed as external effects on the learning process rather than changes in the learning algorithm. In this study, the main reason behind the selection of this method is the observation of the learning process.


\subsection{Autoencoder}

\begin{figure}[h]
	\centering
	\includegraphics[width=1.0\linewidth]{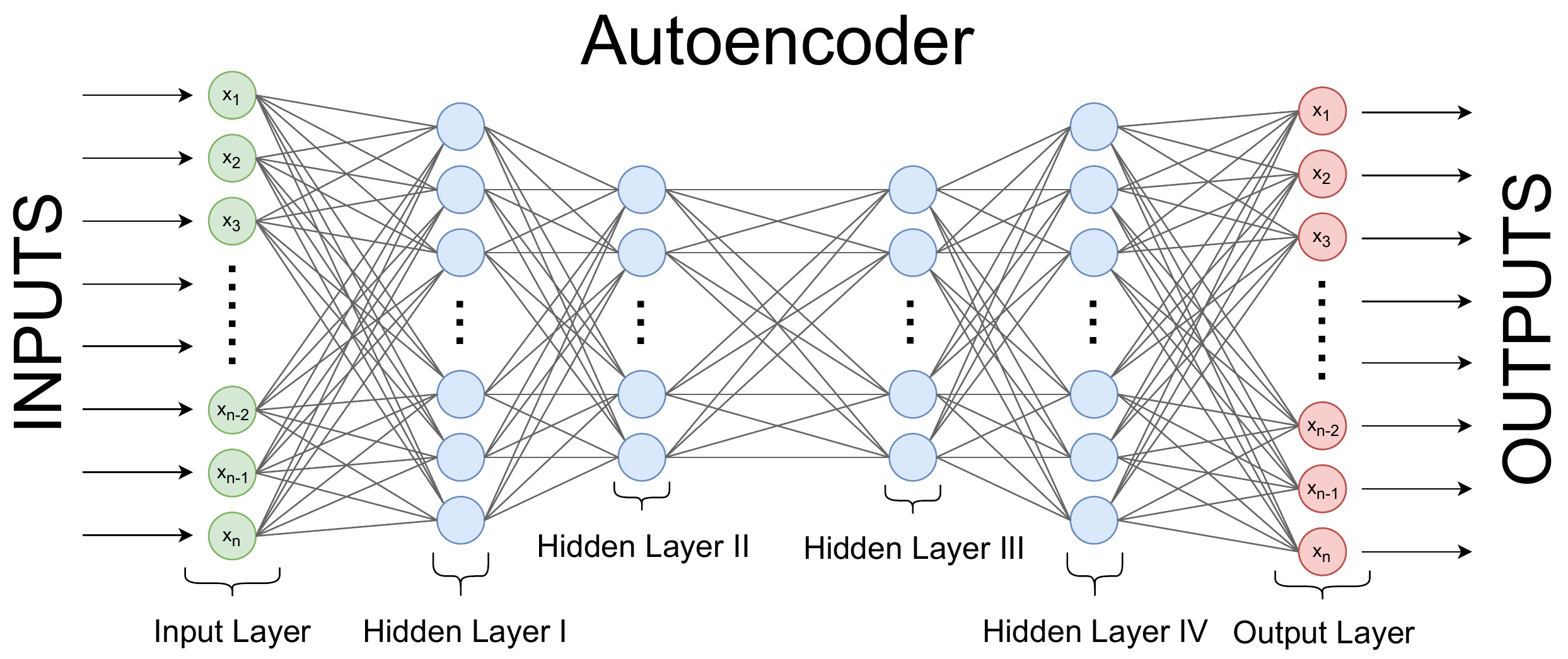}
	\caption{Autoencoder Layer Structure}
	\label{fig:autoencoderLayers}
\end{figure}

An autoencoder neural network is an unsupervised learning algorithm that takes inputs and sets target values to be equals of the input values \cite{paper10}. Autoencoders are generative models that apply backpropagation. They can work without the results of these inputs. While the use of a learning model is in the form of \verb|model.fit(X,Y)|, autoencoders work as \verb|model.fit(X,X)|. The autoencoder works with the ID function to get the output $\mathbf{x}$ that corresponds to $\mathbf{x}$  entries. The identity function seems to be a particularly insignificant function to try to learn; however, there is an interesting structure related to the data, putting restrictions such as limiting the number of hidden units on the network\cite{paper10}. They are neural networks which work as neural networks with an input layer, hidden layers and an output layer but instead of predicting $Y$ as in  \verb|model.fit(X,Y)|, they reconstruct $X$ as in \verb|model.fit(X,X)|. Due to this reconstruction being unsupervised, autoencoders are unsupervised learning models. This structure consists of an encoder and a decoder part. We will define the encoding transition as $\phi$ and decoding transition as $\psi$.

$\phi: X \rightarrow F$ \\
$\psi: F \rightarrow X$ \\
$\phi,\psi = argmin_{\phi,\psi} || X - (\psi  \circ \phi)X||^{2} $

With one hidden layer, encoder will take the input $x \in \mathbb{R}^d =  \chi $ and map it to $ h \in \mathbb{R}^p = F$.
The $h$ below is referred to as latent variables. $\sigma$ is an activation function such as ReLU or sigmoid which were used in this study\cite{paper32,paper33}. $b$ is bias vector, $W$ is weight matrix which both are usually initialized randomly then updated iteratively through training\cite{paper30}. 

$ h = \sigma (Wx + b) $

After the encoder transition is completed, decoder transition maps $h$ to reconstruct $x'$.

$ x' = \sigma'(W'h + b') $ where $\sigma'$, $W'$, $b'$ of decoder are unrelated to $\sigma$, $W$, $b$ of encoder. Loss of autoencoders are trained to be minimal, showed as $L$ below.

$ L (x,x') = || x-x'||^2 = || x-\sigma'(W'(\sigma(Wx+b))+b')||^2 $ 

So the loss function  shows the reconstruction errors, which need to be minimal. After some iterations with input training set $x$ is averaged.

In conclusion,  autoencoders can be seen as  neural networks that reconstruct inputs instead of predicting them. In this paper, we will use them to reconstruct our dataset inputs.

\section{System Model} \label{sec:model}

This section presents the selection of autoencoder model, activation function, and tuning parameters.

\subsection{Creating Autoencoder Model}

\begin{figure}[htbp!]
	\centering
	\includegraphics[width=1.0\linewidth]{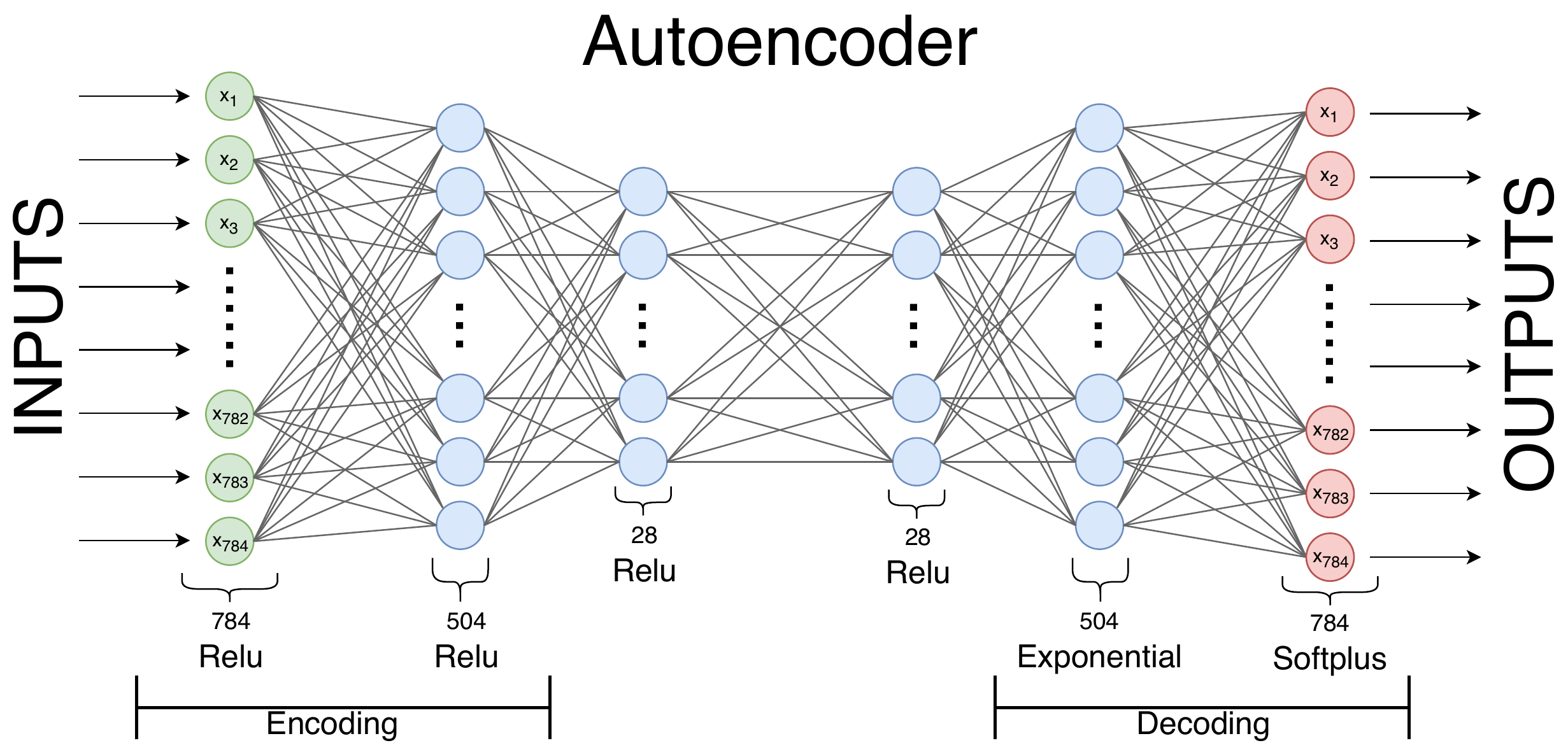}
	\caption{Autoencoder Activation Functions. Note that layer sizes given according to the dataset which is MNIST dataset}
	\label{fig:optimizedAE}
\end{figure}
\begin{figure}[htbp!]
	\centering
	\includegraphics[width=1.0\linewidth]{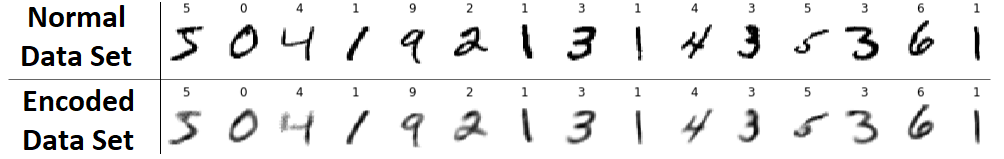}
	\caption{Normal and Encoded Data Set of MNIST}
	\label{fig:autoencodedMnist}
\end{figure}
\begin{figure}[htbp!]
	\centering
	\includegraphics[width=1.0\linewidth]{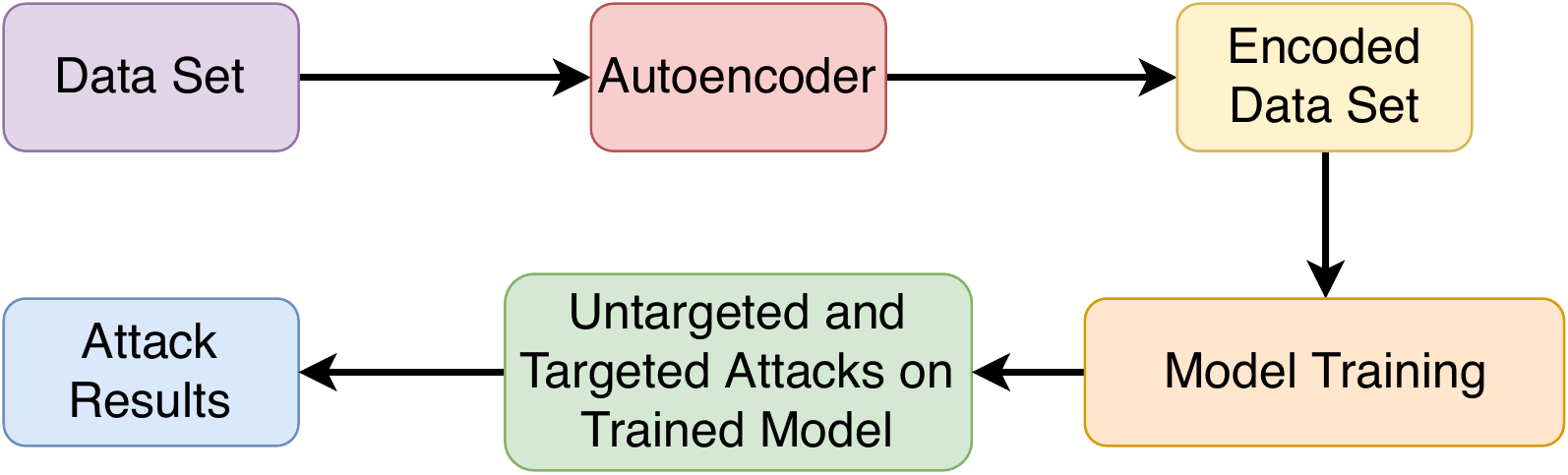}
	\caption{Process Diagram}
	\label{fig:diyagram}
\end{figure}

In this paper, we have selected the MNIST dataset to observe changes easily. Therefore, the size of the layer structure in the autoencoder model is selected as 28 and multipliers to match the MNIST datasets, which represents the numbers by 28 to 28 matrixes. Figure \ref{fig:optimizedAE} presents the structure of matrixes. The modified MNIST data with autoencoder is presented in Figure \ref{fig:autoencodedMnist}. In the training of the model, the encoded data is used instead of using the MNIST datasets directly. As a training method, a multi-class logistic regression method is selected, and attacks are applied to this model. We train autoencoder for 35 epochs. Figure \ref{fig:diyagram} provides the process diagram.

\subsection{Activation Function Selection}

\begin{figure}[hbt!]
	\centering
	\subfigure[Relu Loss History]{\includegraphics[width=0.48\linewidth]{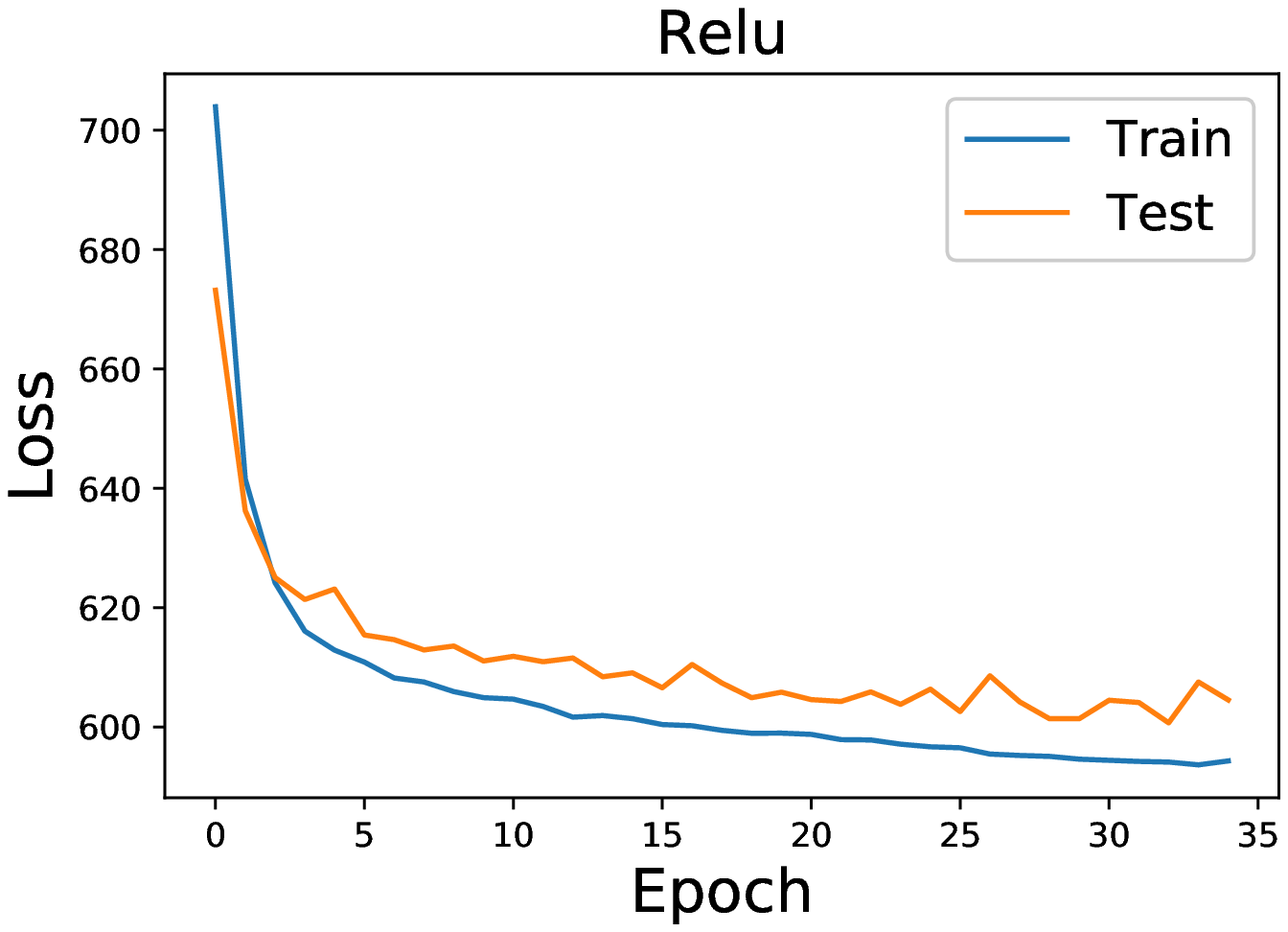}}
	\centering
	\subfigure[Sigmoid Loss History]{\includegraphics[width=0.48\linewidth]{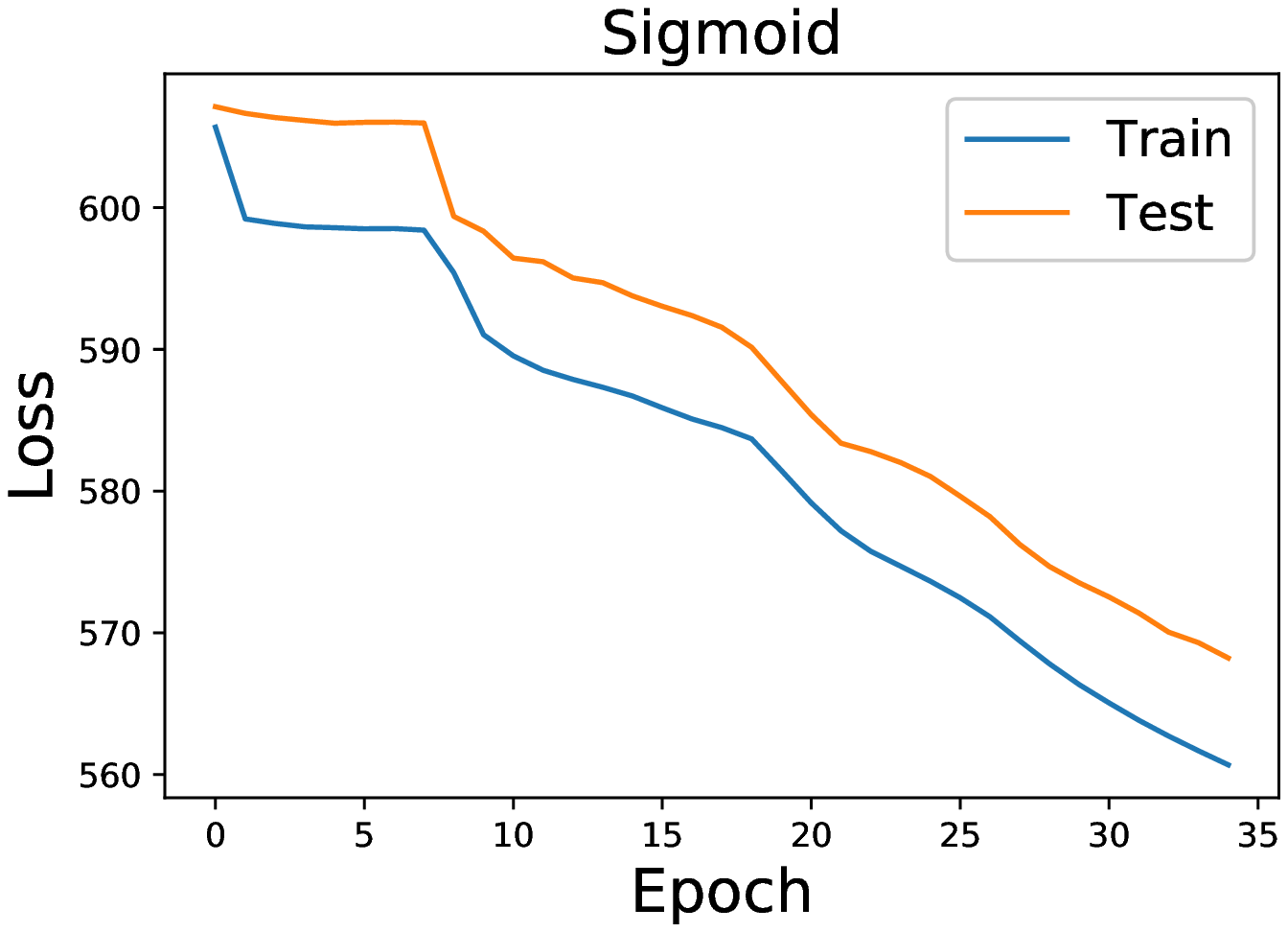}}
	\centering
	\subfigure[Softsign Loss History]{\includegraphics[width=0.48\linewidth]{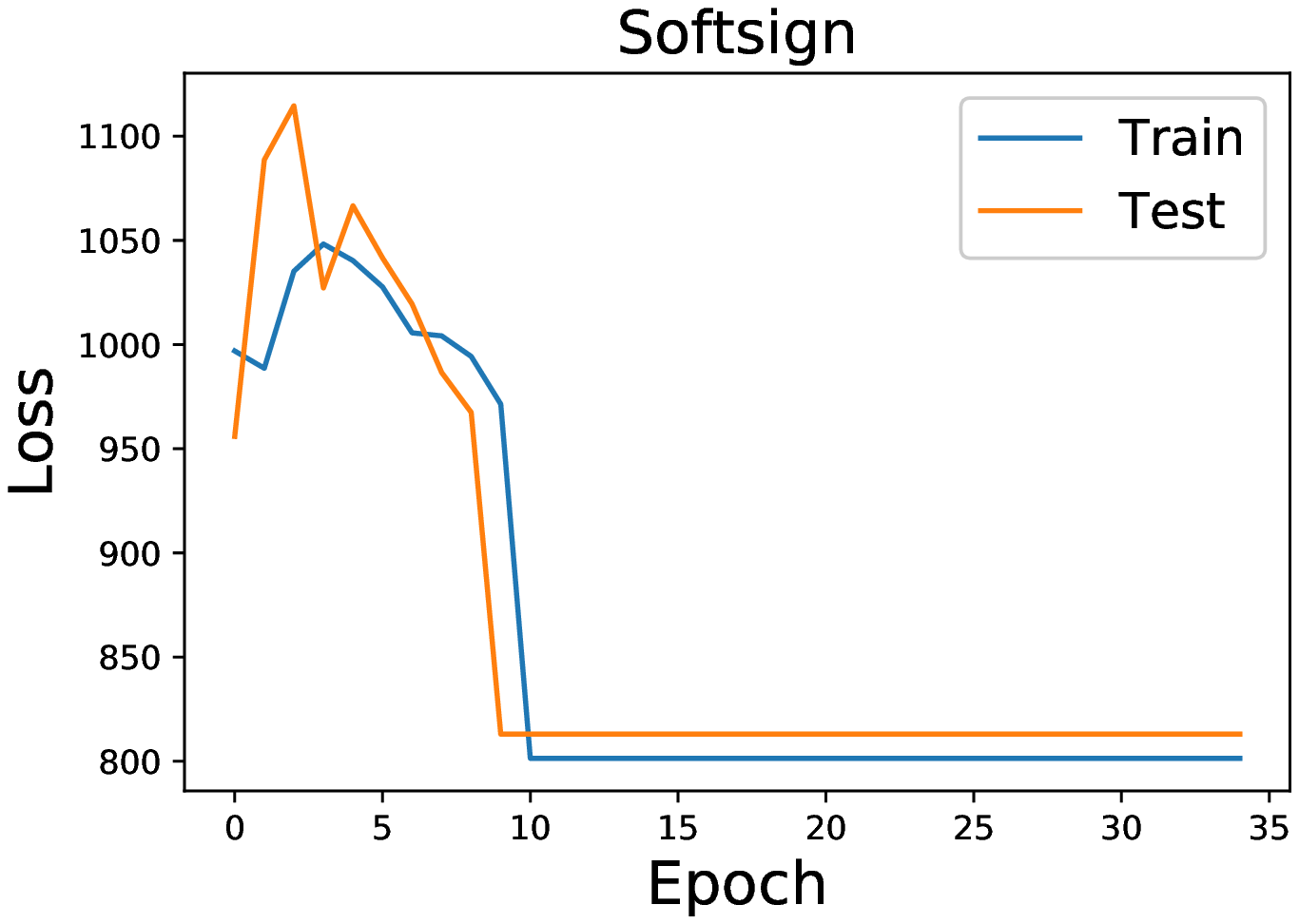}}
	\centering
	\subfigure[Tanh Loss History]{\includegraphics[width=0.48\linewidth]{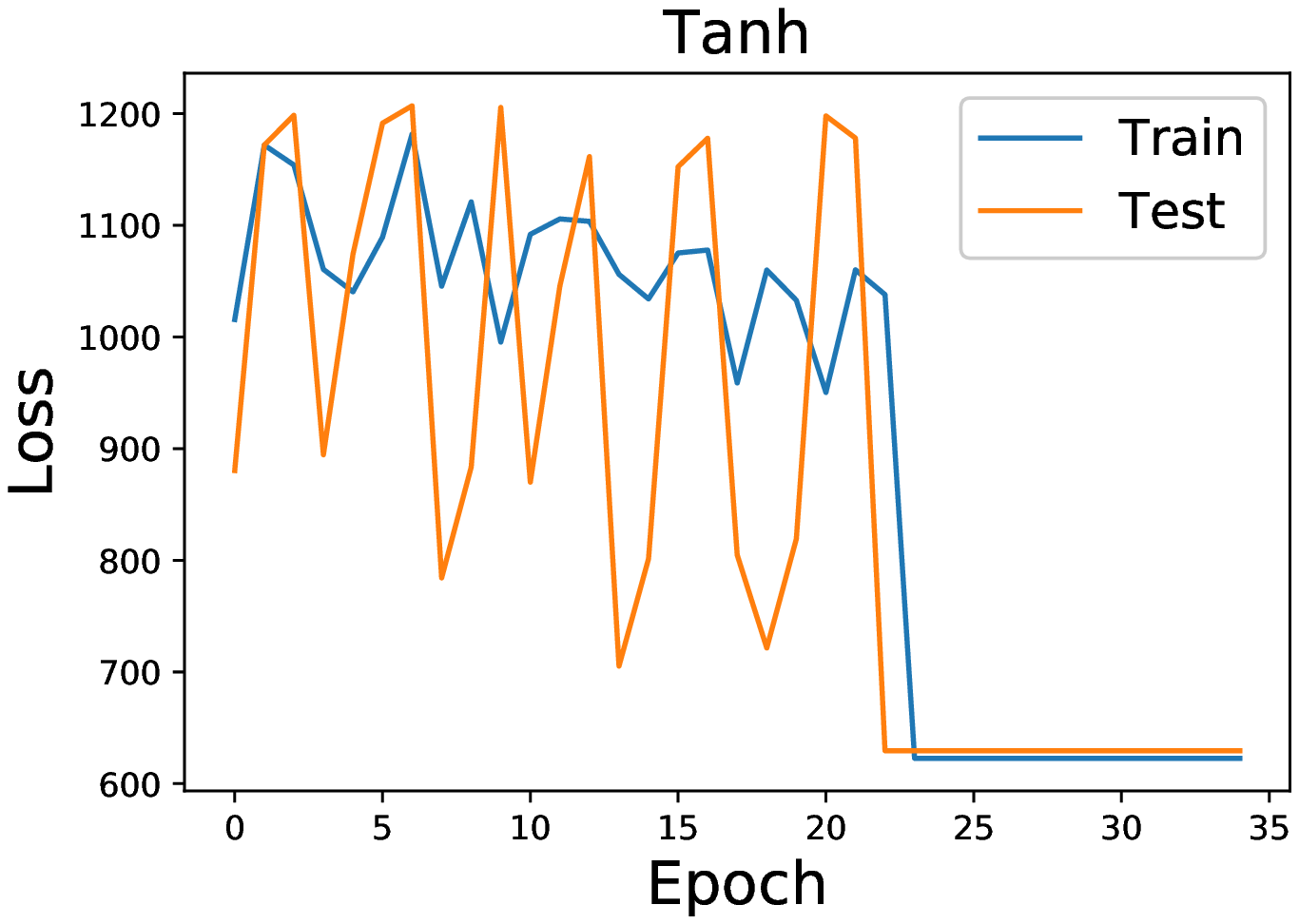}}
	\caption{Loss histories of different activation functions}
	\label{fig:activasion_loss}
\end{figure}

\begin{figure}[hbt]
	\centering
	\includegraphics[width=0.5\linewidth]{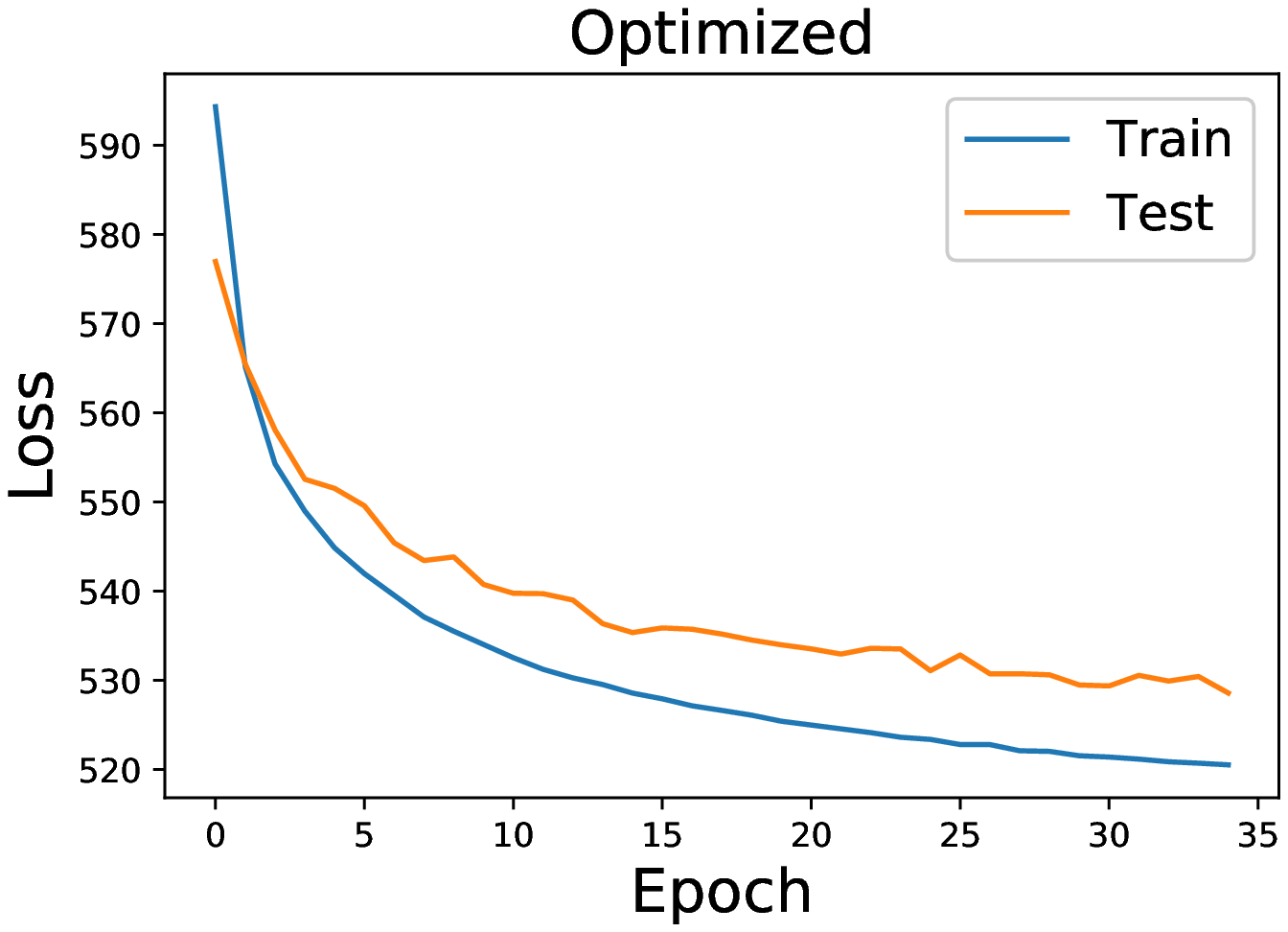}
	\caption{Optimized Relu Loss History}
	\label{fig:optimizedRelu}
\end{figure}

In machine learning and deep learning algorithms, the activation function is used for the computations between hidden and output layers\cite{paper34}. The loss values are compared with different activation functions. Figure \ref{fig:activasion_loss} indicates the comparison results of loss value. \texttt{Sigmoid} and \texttt{ReLU} have the best performance among these values and gave the best results. \texttt{Sigmoid} has more losses at lower epochs than \texttt{ReLU}, but it has better results. Therefore, it is aimed to reach the best result of activation function in both layers. The model with the least loss value is to make the coding parts with the \texttt{ReLU} function and to use the \texttt{exponential} and \texttt{softplus} functions in the analysis part respectively. These functions are used in our study. Figure \ref{fig:optimizedRelu} illustrates the result of the loss function, and Figure \ref{fig:optimizedAE} presents the structure of the model with the activation functions.

\subsection{Tuning Parameters}
The tuning parameters for autoencoders depend on the dataset we use and what we try to apply. As previously mentioned, \texttt{ReLU} and sigmoid function are selected to be activation function for our model \cite{paper33,paper34}. \texttt{ReLU} is the activation function through the whole autoencoder while exponential is the softplus being the output layer’s activation function which yields the minimal loss. Figure \ref{fig:optimizedAE} presents the input size as 784 due to our dataset and MNIST dataset contains 28x28 pixel images\cite{paper35}. Encoding part for our autoencoder size is $784 \times 504 \times 28$ and decoding size is $28 \times 504 \times 784$. 

This structure is selected by the various neural network structures that take the square of the size of the matrix, lower it, and give it to its dimension size lastly. The last hidden layer of the decoding part with the size of 504 uses \texttt{exponential} activation function, and an output layer with the size of 784 uses \texttt{softplus} activation function \cite{paper38,paper39}. We used \texttt{adam} optimizer with categorical crossentropy\cite{paper37,paper36}. We see that a small number is enough for training, so we select epoch number for autoencoder as 35. This is the best epoch value to get meaningful results for both models with autoencoder and without autoencoder to see accuracy. In lower values, models get their accuracy scores too low for us to see the difference between them, even though some models are structurally stronger than others.

\section{Experiments with MNIST Dataset} \label{sec:experimentswithmnist}
\subsection{Introduction}

We examine the robustness of autoencoder for adversarial machine learning with different machine learning algorithms and models to see that autoencoding can be a generalized solution and an easy to use defense mechanism for most adversarial attacks. We use various linear machine learning model algorithms and neural network model algorithms against adversarial attacks. 

\subsection{Autoencoding}

In this section, we look at the robustness provided with auto-encoding. We select a linear model and a neural network model to demonstrate this effectiveness. In these models, we also observe the robustness of different attack methods. We also use the MNIST dataset for these examples.

\subsubsection{Multi-Class Logistic Regression}

\begin{figure}[htbp!]
	\centering
	\includegraphics[width=0.8\linewidth]{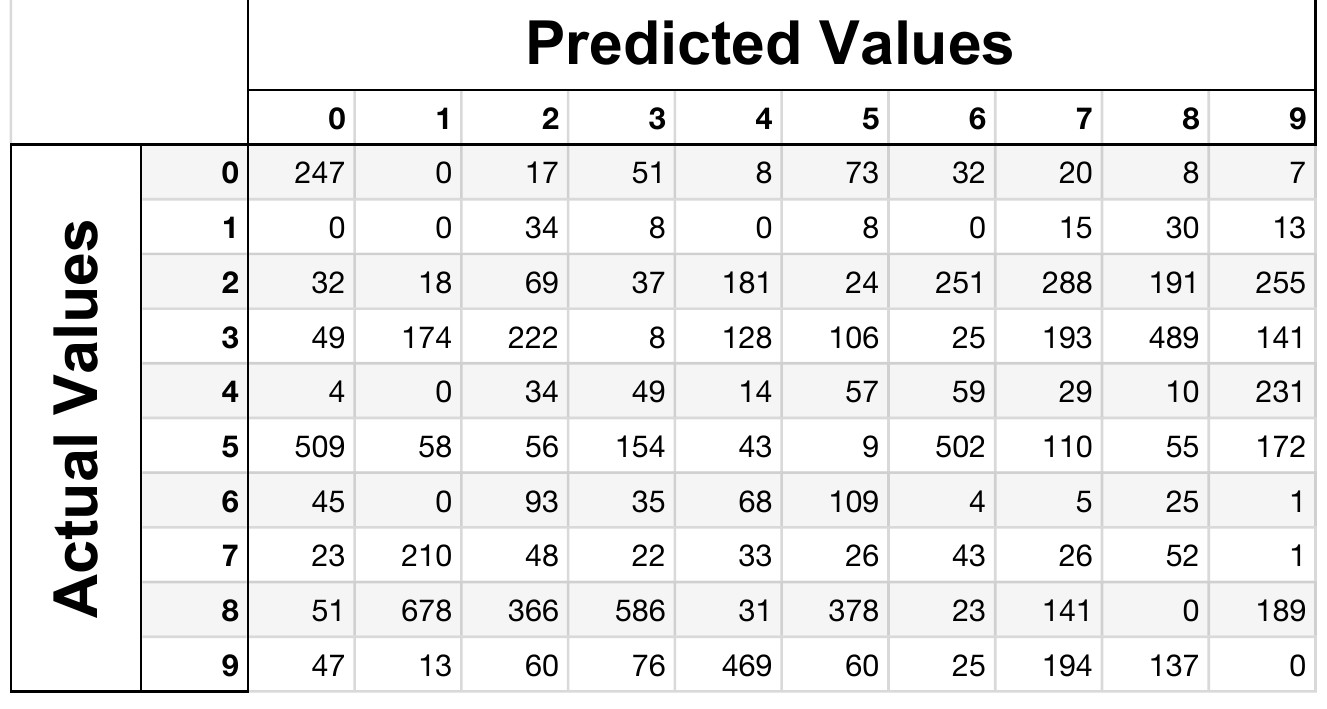}
	\caption{Confusion matrix of non-targeted attack to model without autoencoder}
	\label{fig:CMWoAE}
\end{figure}

In linear machine learning model algorithms, we use mainly two attack methods: Non-Targeted and Targeted Attacks. The non-targeted attack does not concern with how the machine learning model makes its predictions and tries to force the machine learning model into misprediction. On the other hand, targeted attacks focus on leading some correct predictions into mispredictions. 
We have three methods for targeted attacks: Natural, Non-Natural, and one selected target. Firstly, natural targets are derived from the most common mispredictions made by the machine learning model. For example, guessing number 5 as 8, and number 7 as 1 are common mispredictions. Natural targets take these non-targeted attack results into account and attack directly to these most common mispredictions. So, when number 5 is seen, an attack would try to make it guessed as number 8. Secondly, non-natural targeted attacks are the opposite of natural targeted attacks. It takes the minimum number of mispredictions made by the machine learning model with the feedback provided by non-natural attacks. For example, if number 1 is least mispredicted as 0, the non-natural target for number 1 is 0. Therefore, we can see that how much the attack affects the machine learning model beyond its common mispredictions. Lastly, one targeted attack focuses on some random numbers. The aim is to make the machine learning model mispredict the same number for all numbers. For linear classifications, we select multi-class logistic regression to analyze the attacks. Because we do not interact with these linear classification algorithms aside from calling their defined functions from scikit-learn library, we use a black-box environment for these attacks. In our study, the attack method against multi-class classification models developed in NIPS 2017 is used \cite{paper11}. An epsilon value is used to determine the severity of the attack, which we select 50 in this study to demonstrate the results better. We apply a non-targeted attack to a multi-class logistic regression trained model which is trained with MNIST dataset without an autoencoder. The confusion matrix of this attack is presented in \ref{fig:CMWoAE}.

\begin{figure}[htbp!]
	\centering
	\includegraphics[width=0.8\linewidth]{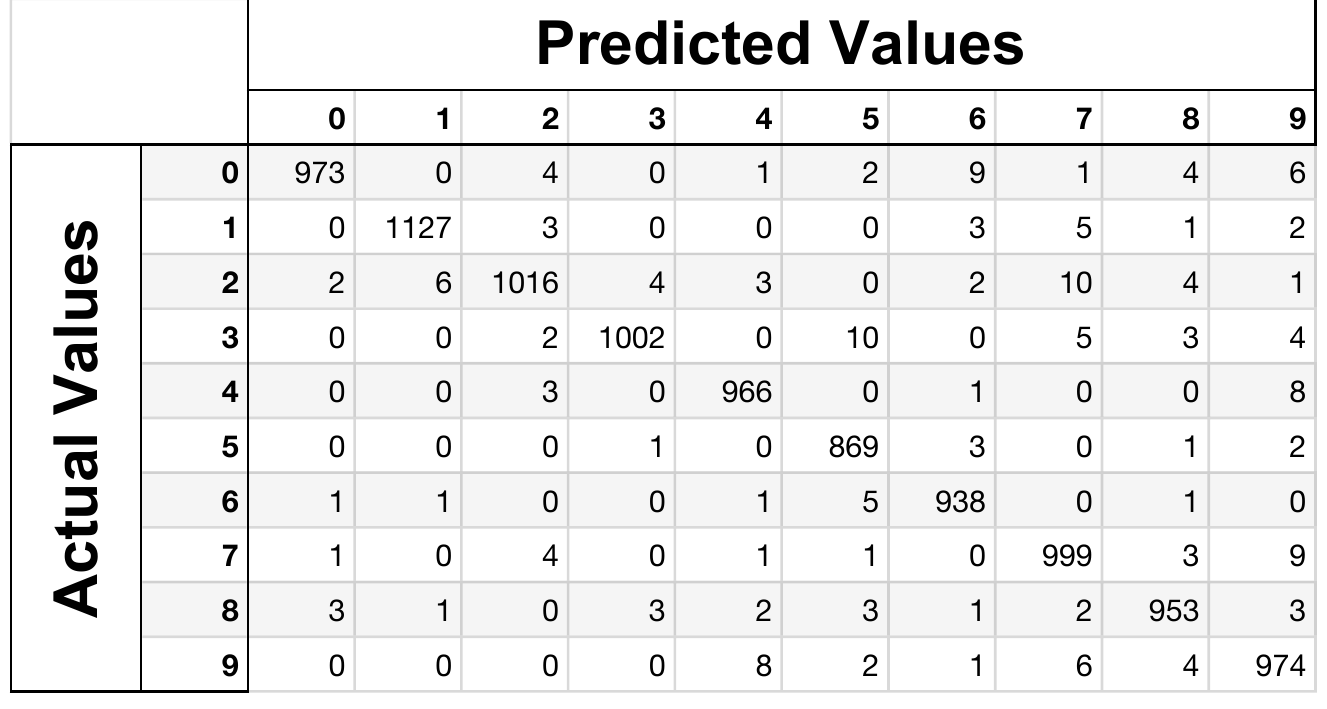}
	\caption{Confusion matrix of the model without any attack and without autoencoder}
	\label{fig:WithoutAttack_WOAE_MCLR}
\end{figure}

\begin{figure}[htbp!]
	\centering
	\includegraphics[width=0.8\linewidth]{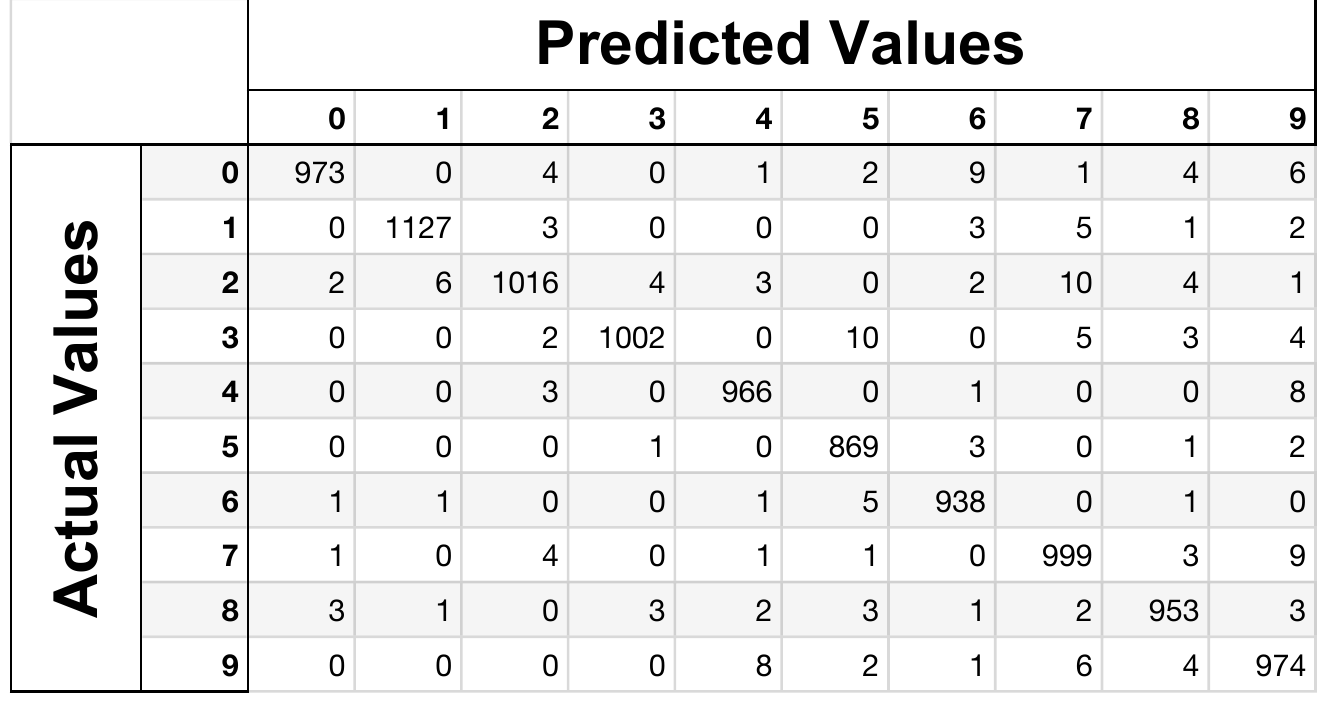}
	\caption{Confusion matrix of the model without any attack and with autoencoder}
	\label{fig:WithoutAttack_AE_MCLR}
\end{figure}

\begin{figure}[htbp!]
	\centering
	\includegraphics[width=0.8\linewidth]{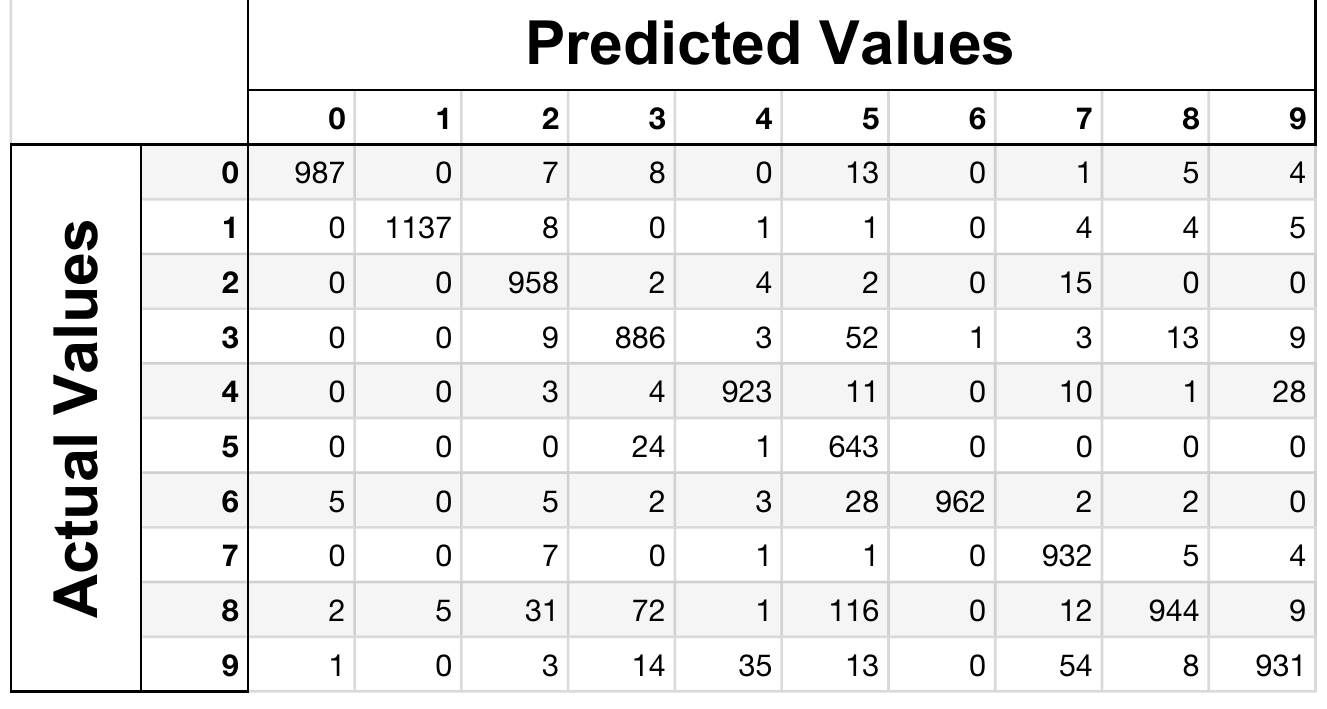}
	\caption{Confusion matrix of non-targeted attack to model with autoencoder}
	\label{fig:CMAE}
\end{figure}

\begin{figure}[htbp!]
	\centering
	\includegraphics[width=0.75\linewidth]{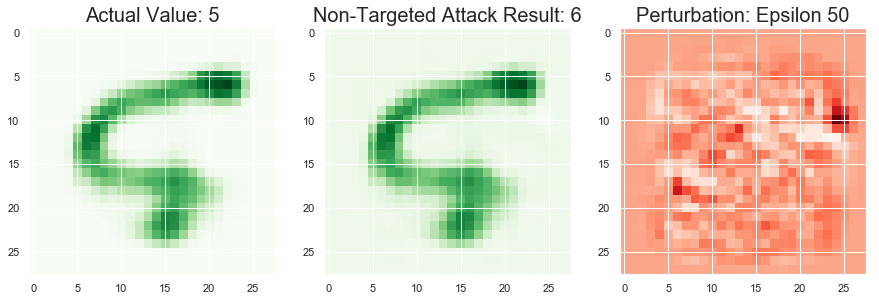}
	\caption{Value change and perturbation of a non-targeted attack on model without autoencoder}
	\label{fig:perturbasyonWoAE}
\end{figure}
\begin{figure}[htbp!]
	\includegraphics[width=1.0\linewidth]{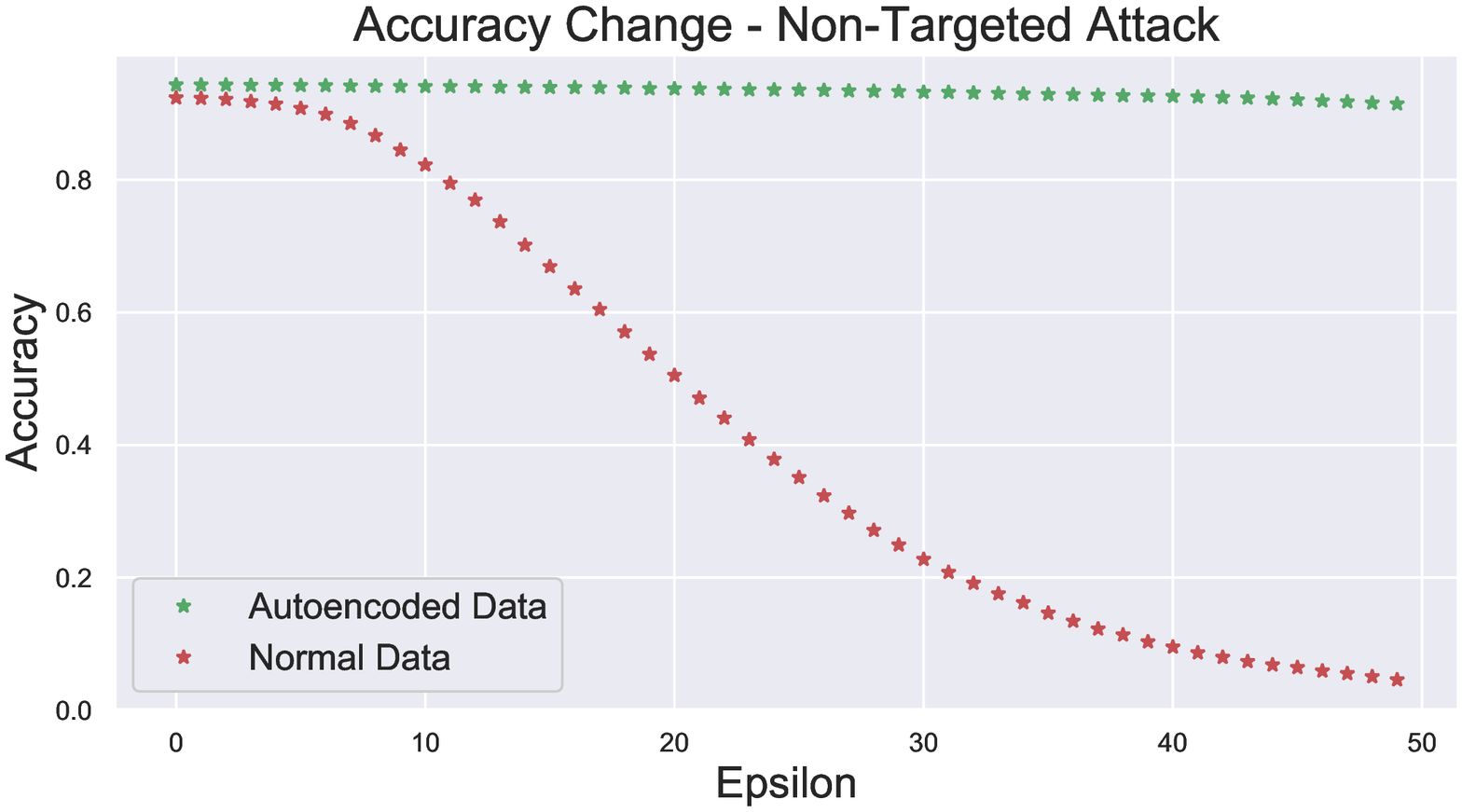}
	\caption{Comparison of accuracy with and without autoencoder for non-targeted attack}
	\label{fig:nontargetedAttack1}
\end{figure}

The findings from Figure \ref{fig:CMWoAE} and \ref{fig:CMAE} show that an autoencoder model provides robustness against non-targeted attacks. The accuracy value change with epsilon is presented in Figure \ref{fig:nontargetedAttack1}. Figure \ref{fig:perturbasyonWoAE} illustrates the change and perturbation of the selected attack with epsilon value as 50.

\begin{figure}[htbp!]
	\centering
	\includegraphics[width=0.75\linewidth]{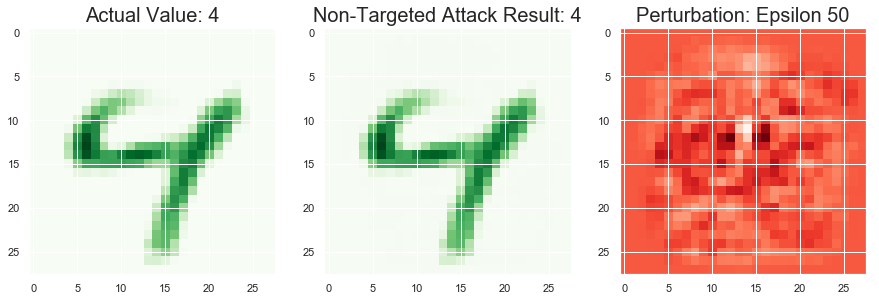}
	\caption{Value change and perturbation of a non-targeted attack on model with autoencoder}
	\label{fig:perturbasyonAE}
\end{figure}

\begin{figure}[htbp!]
	\includegraphics[width=1.0\linewidth]{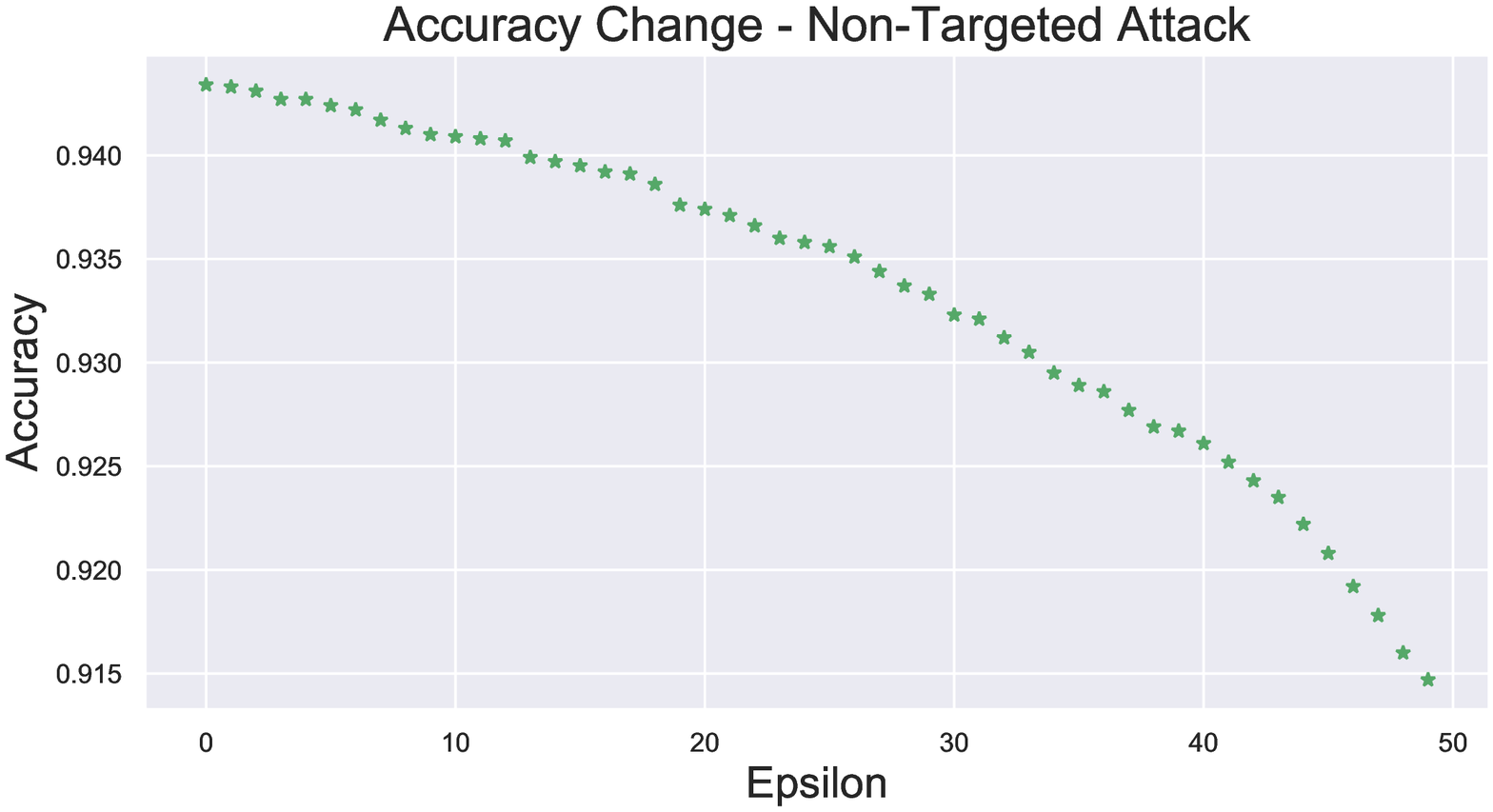}
	\caption{Details of accuracy with autoencoder for non-targeted attack}
	\label{fig:nontargetedAttack2}
\end{figure}

We apply a non-targeted attack on the multi-class logistic regression model with autoencoder and without autoencoder. Figure \ref{fig:nontargetedAttack1} provides a difference in accuracy metric. The detailed graph of the non-targeted attack on the model with autoencoder is presented in Figure \ref{fig:nontargetedAttack2}. The changes in the MNIST dataset after autoencoder is provided in Figure \ref{fig:autoencodedMnist}. The value change and perturbation of an epsilon 50 value on data are indicated in Figure \ref{fig:perturbasyonAE}.

\begin{figure}[htbp!]
	\centering
	\includegraphics[width=0.7\linewidth]{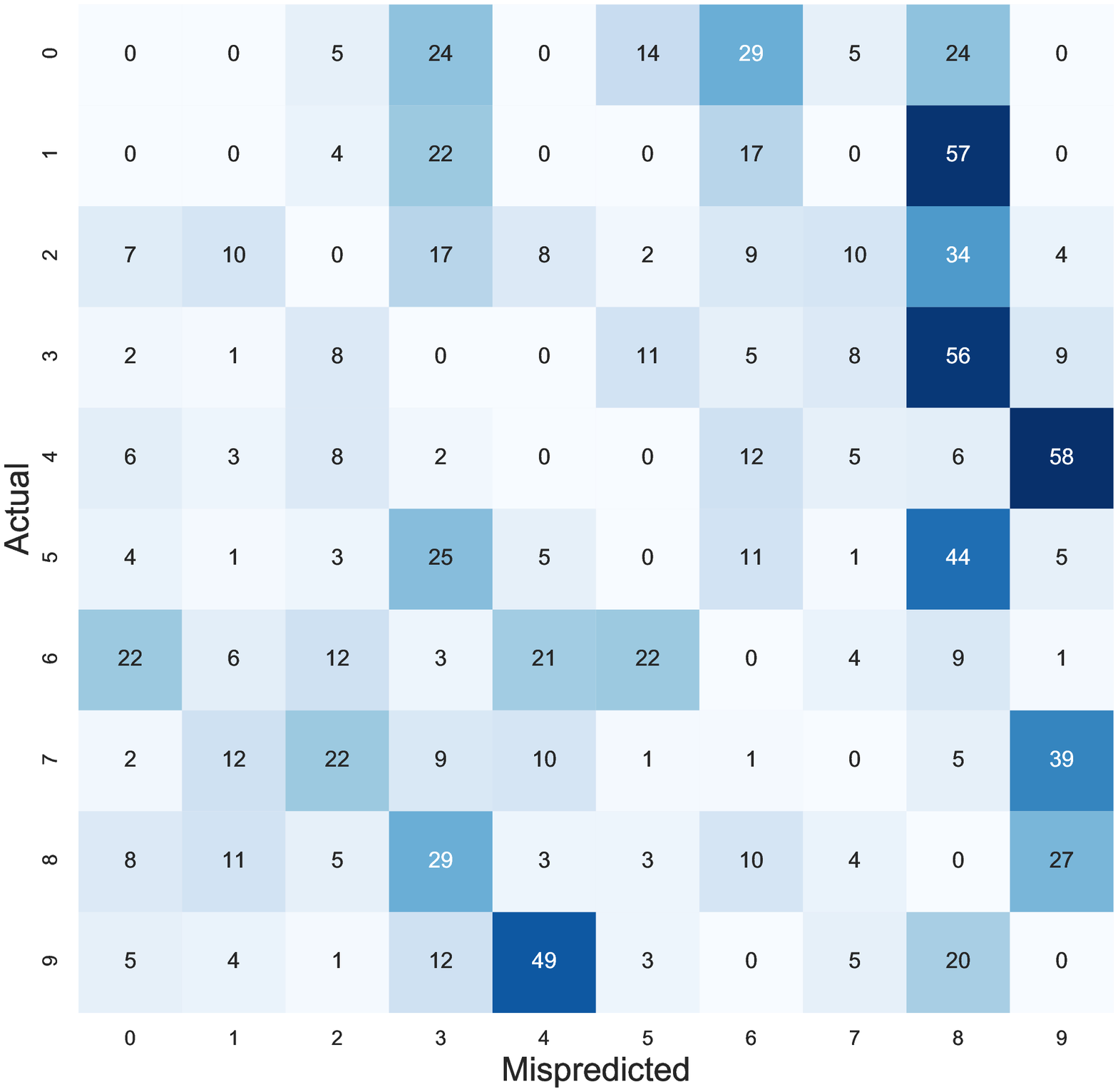}
	\caption{Heatmap of actual numbers and mispredictions}
	\label{fig:Heatmap}
\end{figure}

The following process is presented in Figure \ref{fig:diyagram}. In the examples with the autoencoder, data is passed through the autoencoder and then given to the training model, in our current case a classification model with multi-class logistic regression. Multi-class logistic regression uses the encoded dataset for training. Figure \ref{fig:CMAE} provides to see improvement as a confusion matrix. For the targeted attacks, we select three methods to use. The first one is natural targets for MNIST dataset, which is also defined in NIPS 2017 \cite{paper11}. Natural targets take the non-targeted attack results into account and attack directly to these most common mispredictions. For example, the natural target for number 3 is 8. When we apply the non-targeted attack, we obtain these results. Heat map for these numbers is indicated in Figure \ref{fig:Heatmap}. 

The second method of targeted attacks is non-natural targets which is the opposite of natural targets. We select the least mis predicted numbers as the target. These numbers is indicated as the heat map in Figure \ref{fig:Heatmap}. The third method is the selection one number and making all numbers predict it. We randomly choose 7 as that target number. Targets for these methods are presented in Figure \ref{fig:targetValues}. The confusion matrixes for these methods are presented below.

\begin{figure}[htbp!]
	\centering
	\includegraphics[scale=0.45]{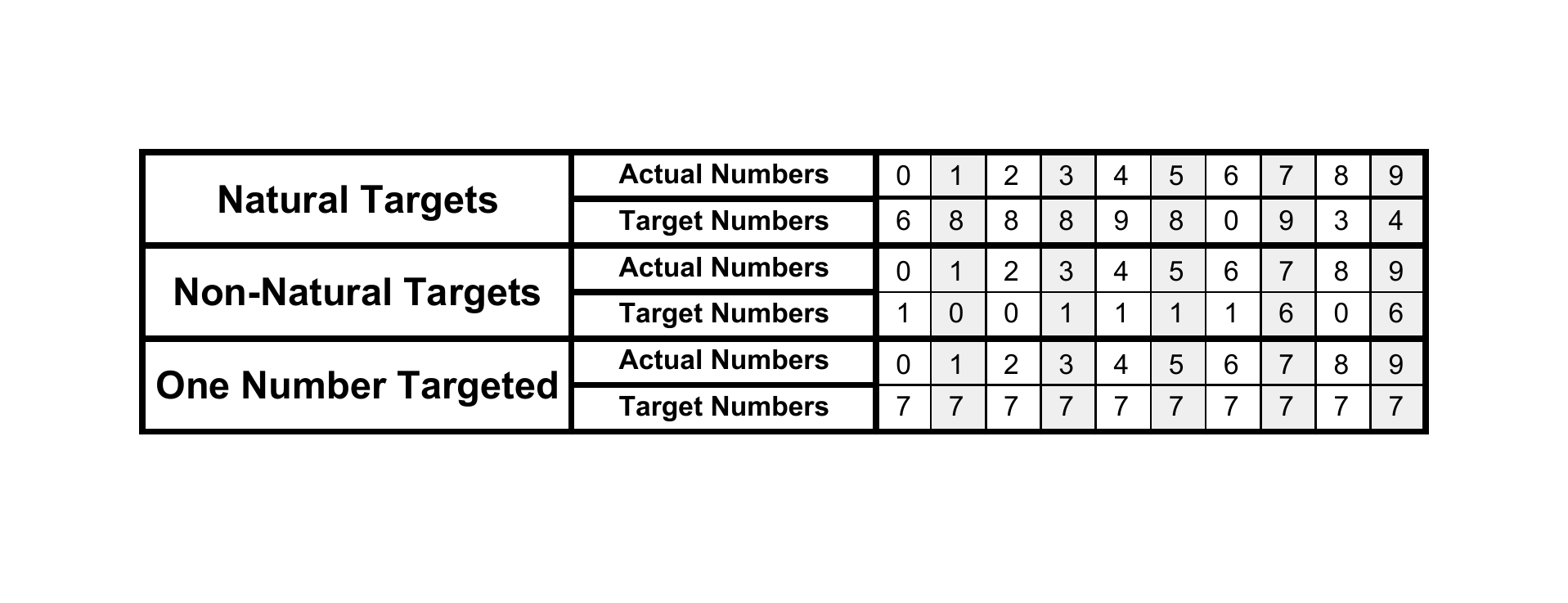}
	\caption{Actual numbers and their target values for each targeted attack method}
	\label{fig:targetValues}
\end{figure}

\begin{figure}[htbp!]
	\centering
	\includegraphics[scale=0.5]{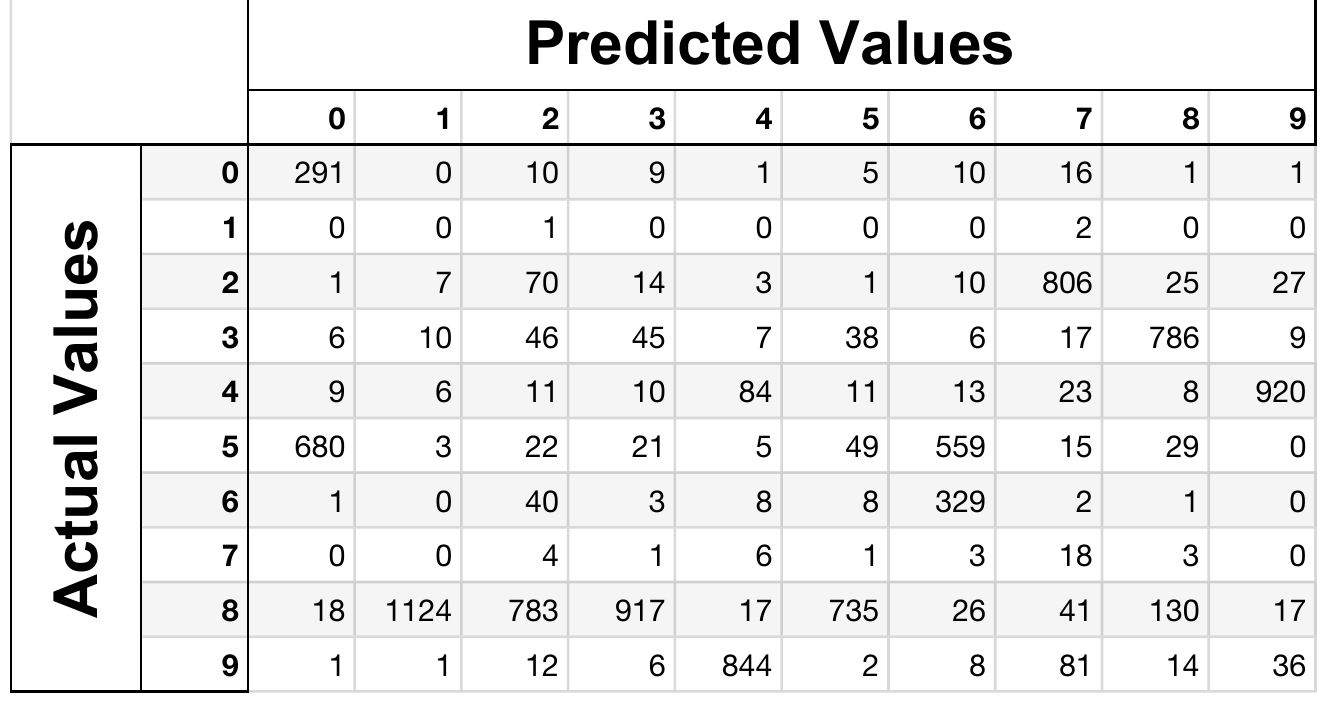}
	\caption{Confusion matrix of \textbf{natural targeted attack} to model without autoencoder}
	\label{fig:NWO}
\end{figure}

\begin{figure}[htbp!]
	\centering
	\includegraphics[scale=0.5]{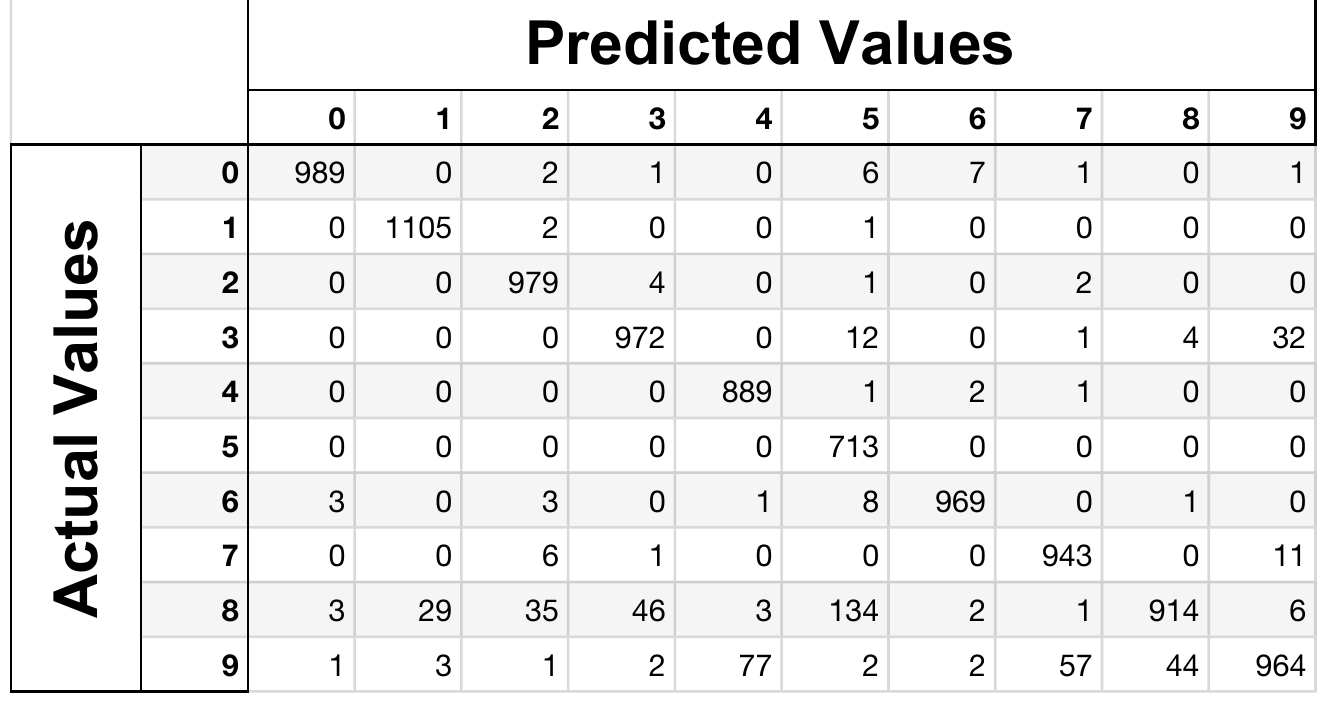}
	\caption{Confusion matrix of \textbf{natural targeted attack} to model with autoencoder}
	\label{fig:NAE}
\end{figure}

\begin{figure}[htbp!]
	\centering
	\includegraphics[scale=0.5]{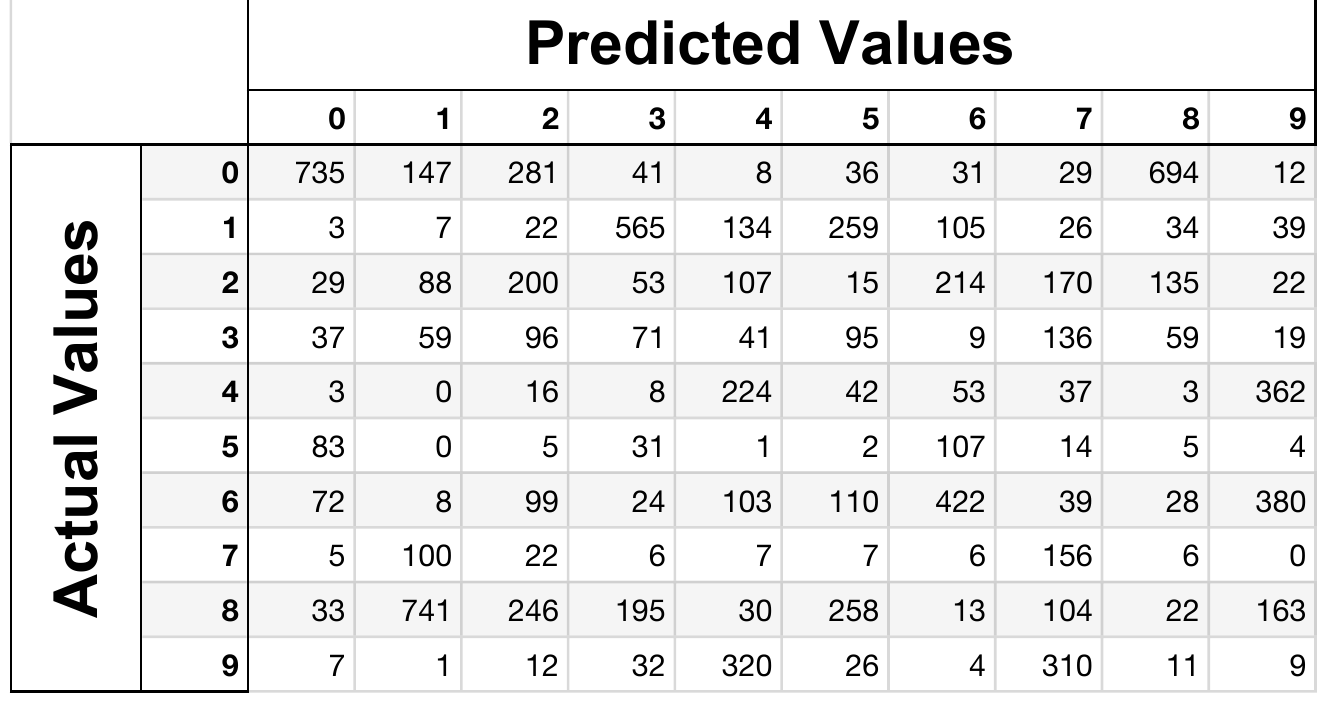}
	\caption{Confusion matrix of \textbf{non-natural targeted attack} to model without autoencoder}
	\label{fig:NNWO}
\end{figure}

\begin{figure}[htbp!]
	\centering
	\includegraphics[scale=0.5]{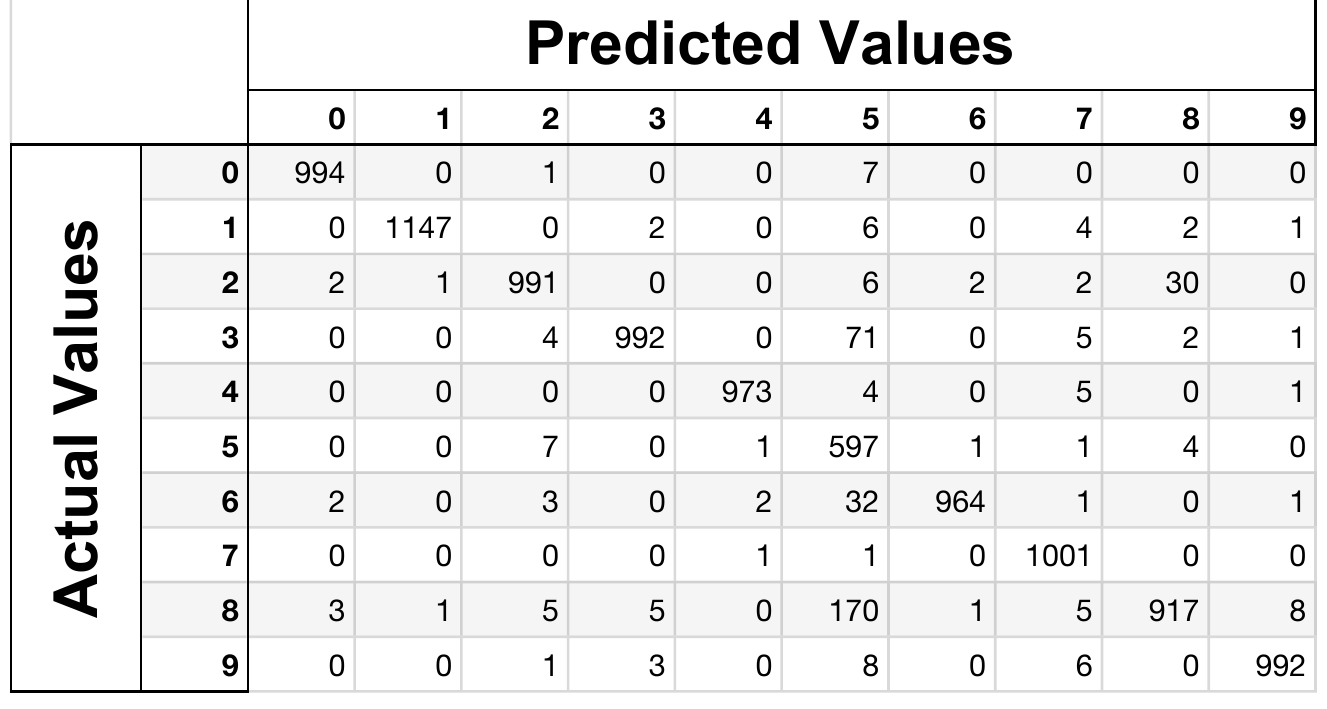}
	\caption{Confusion matrix of \textbf{non-natural targeted attack} to model with autoencoder}
	\label{fig:NNAE}
\end{figure}

\begin{figure}[htbp!]
	\centering
	\includegraphics[scale=0.5]{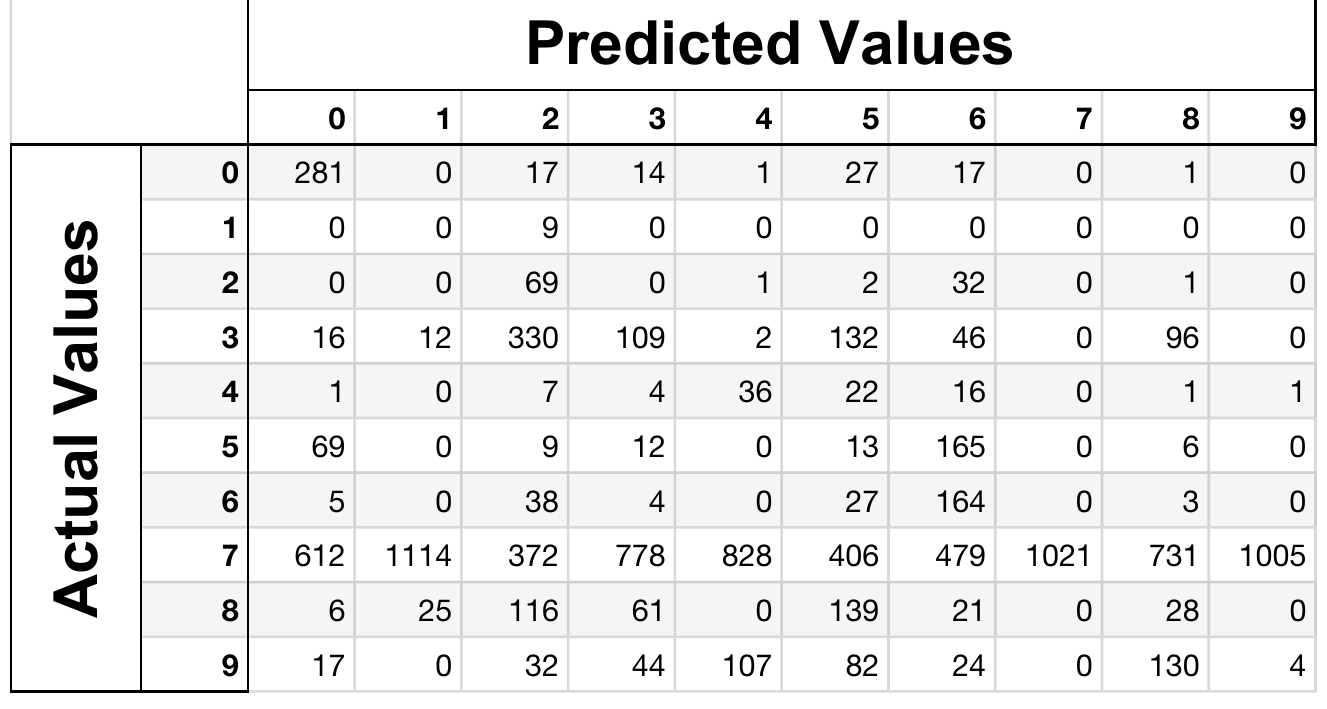}
	\caption{Confusion matrix of \textbf{one number targeted attack} to model without autoencoder}
	\label{fig:ONWO}
\end{figure}

\begin{figure}[htbp!]
	\centering
	\includegraphics[scale=0.5]{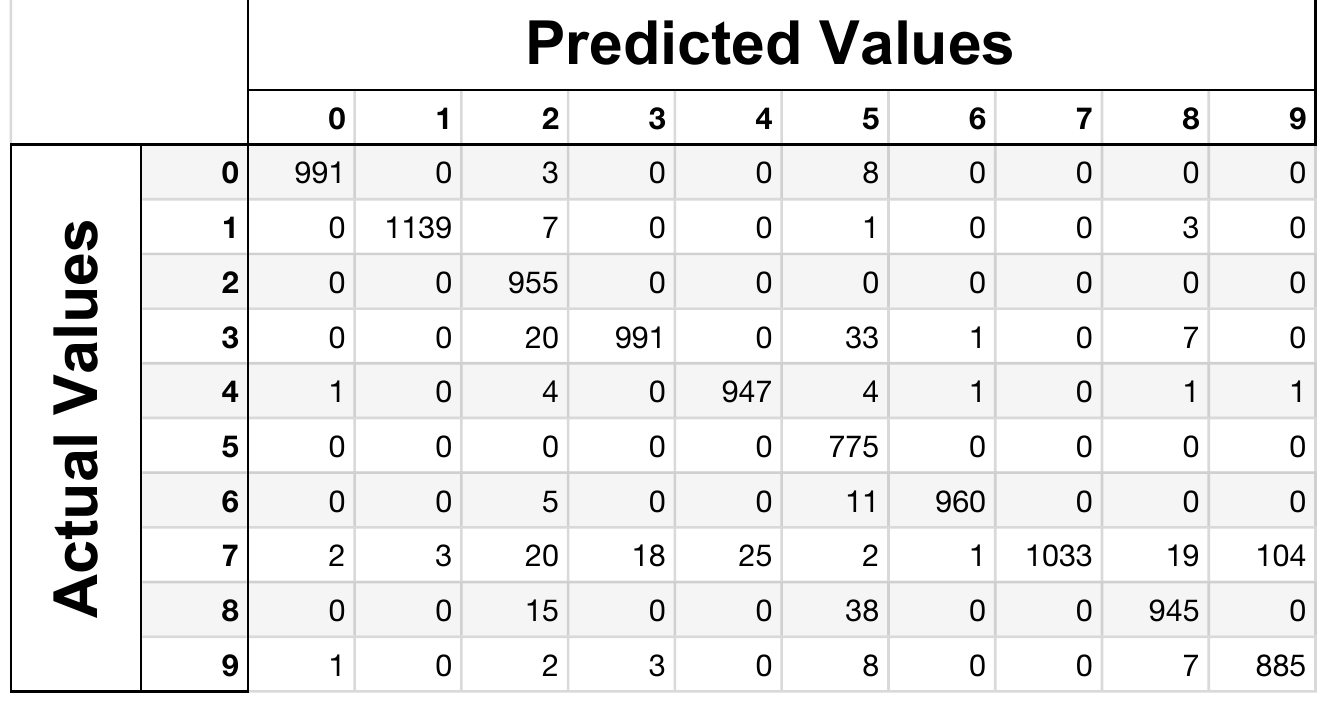}
	\caption{Confusion matrix of \textbf{one number targeted attack} to model with autoencoder}
	\label{fig:ONAE}
\end{figure}

\begin{figure}[htbp!]
	\includegraphics[width=1.0\linewidth]{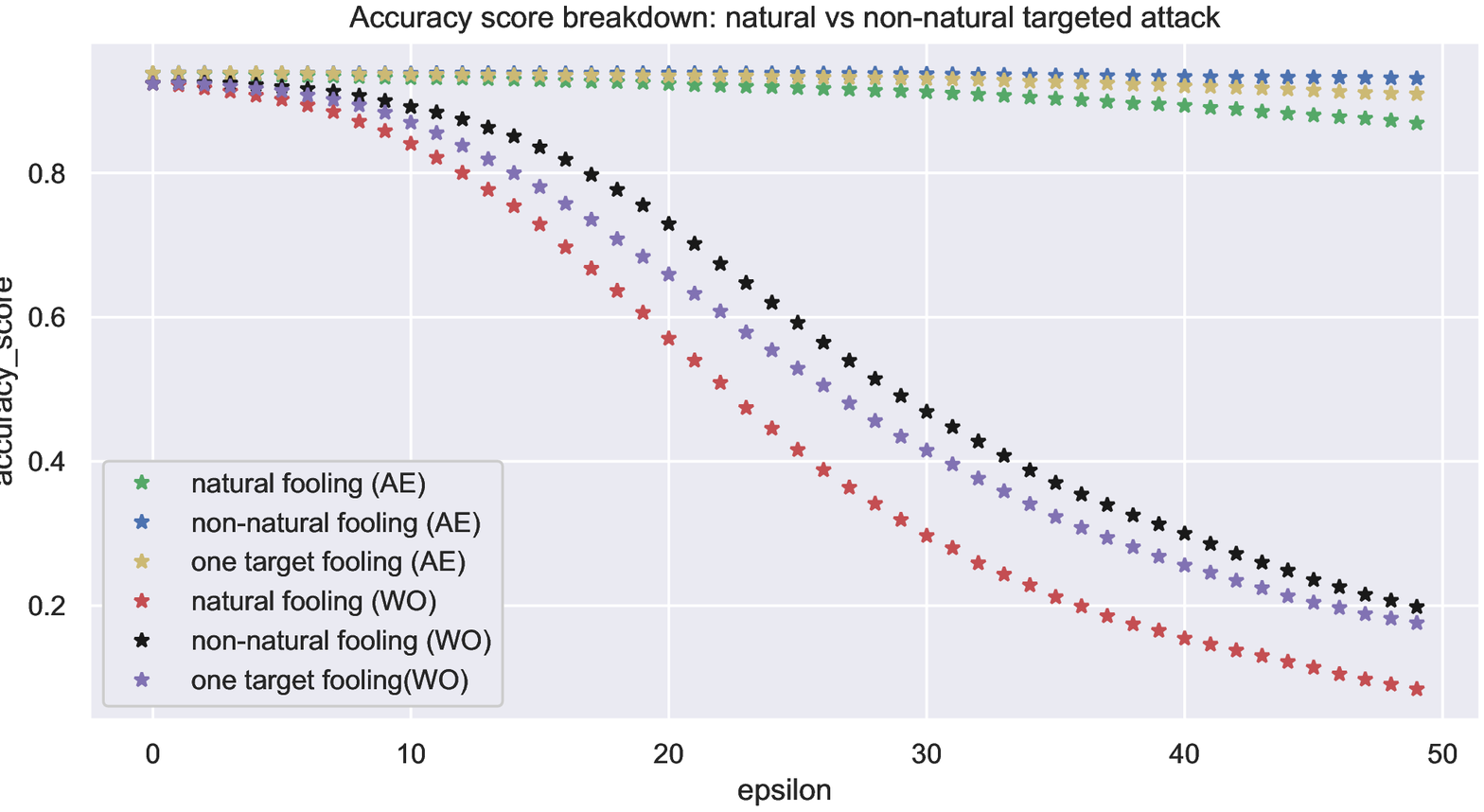}
	\caption{Comparison of accuracy with and without autoencoder for targeted attacks. \textit{AE stands for the models with autoencoder, WO stands for models without autoencoder}}
	\label{fig:targetedAttacks}
\end{figure}

\begin{figure}[htbp!]
	\includegraphics[width=1.0\linewidth]{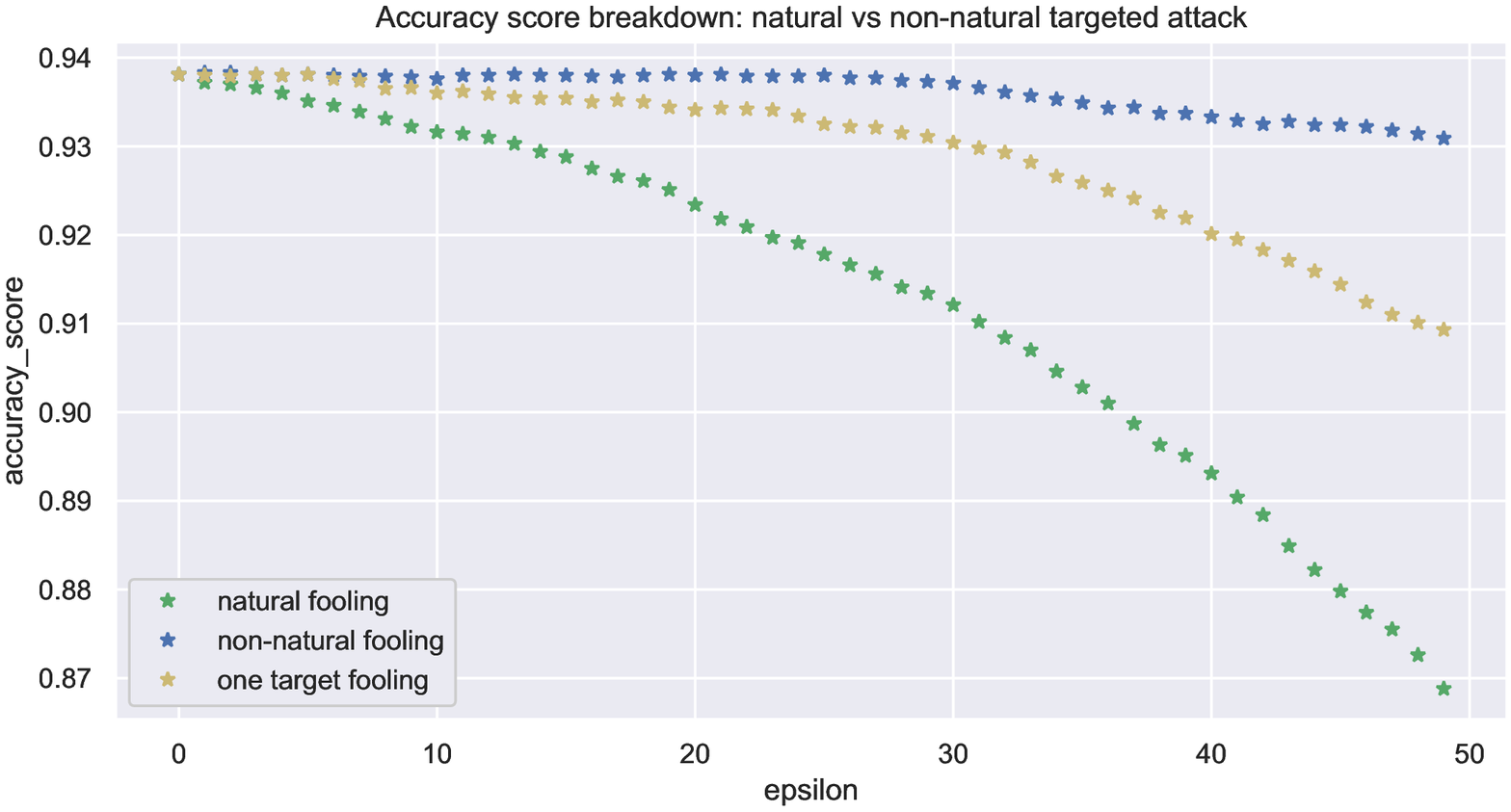}
	\caption{Details of accuracy with autoencoder for targeted attacks}
	\label{fig:targetedAttacks2}
\end{figure}




\subsubsection{Neural Networks} 

We use neural networks with the same principles as multi-class logistic regressions and make attacks to the machine learning model. We use the same structure, layer, activation functions and epochs for these neural networks as we use in our autoencoder for simplicity. Although this robustness will work with other neural network structures, we will not demonstrate them in this study due to structure designs that can vary for all developers. We also compare the results of these attacks with both the data from the MNIST dataset and the encoded data results of the MNIST dataset. As for attack methods, we select three methods:  FGSM, T-FGSM and BIM. Cleverhans library is used for providing these attack methods to the neural network, which is from the Keras library.

\begin{figure}[htbp!]
	\centering
	\includegraphics[scale=0.56]{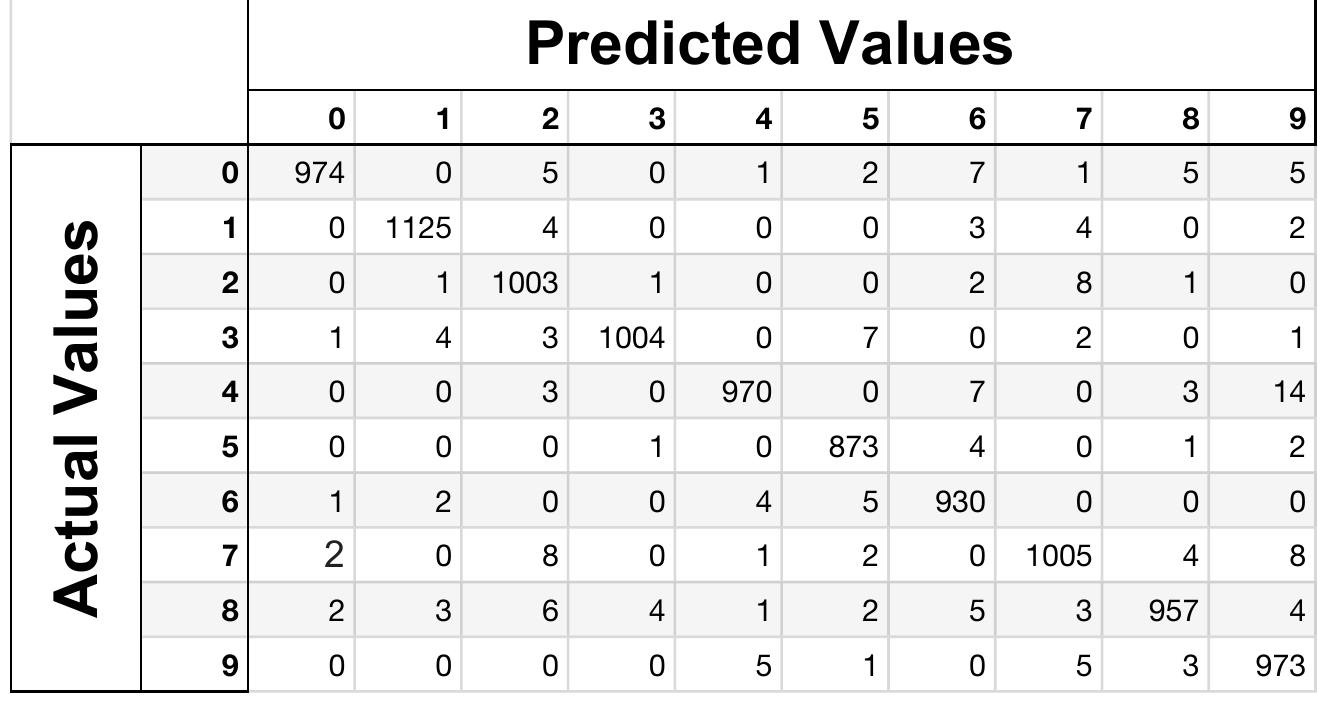}
	\includegraphics[scale=0.5]{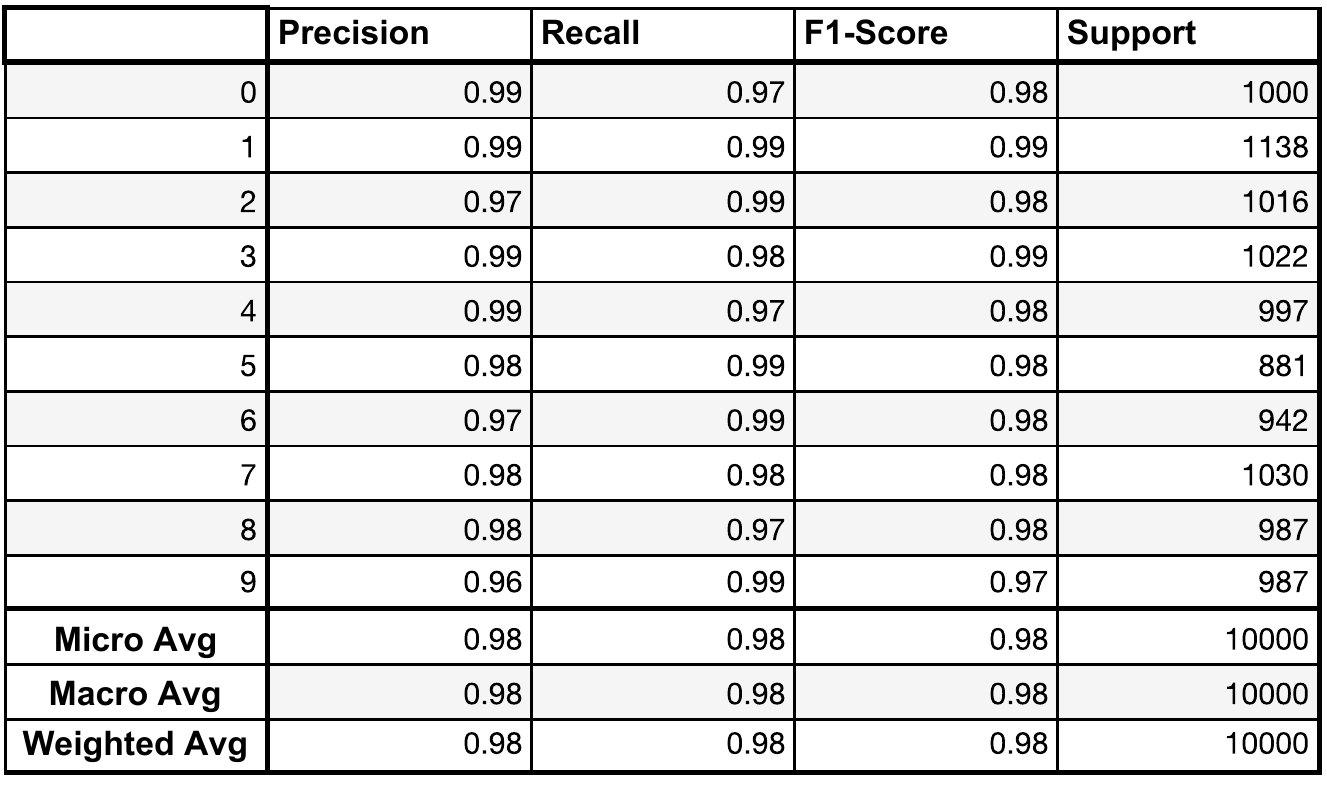}
	\caption{Confusion matrix and classification report of the neural network model without autoencoder}
	\label{fig:NN_AE_WOAE}
\end{figure}

\begin{figure}[htbp!]
	\centering
	\includegraphics[scale=0.56]{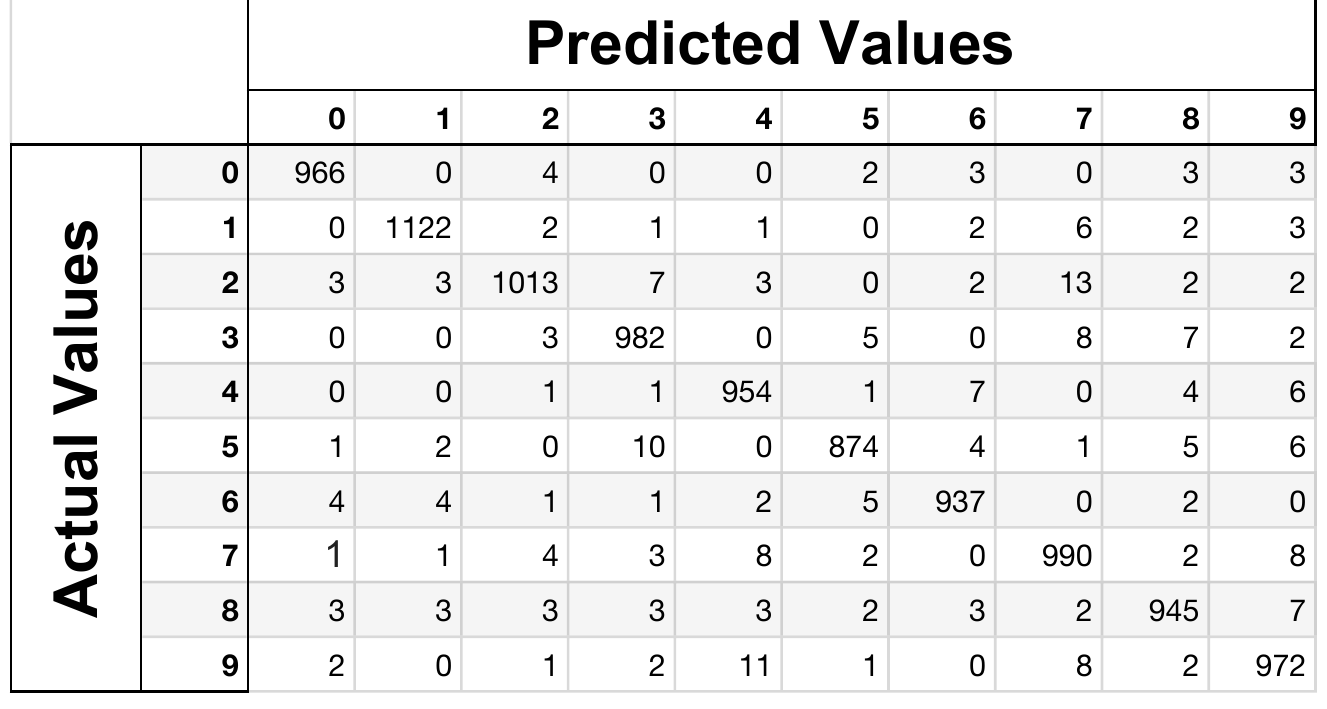}
	\includegraphics[scale=0.5]{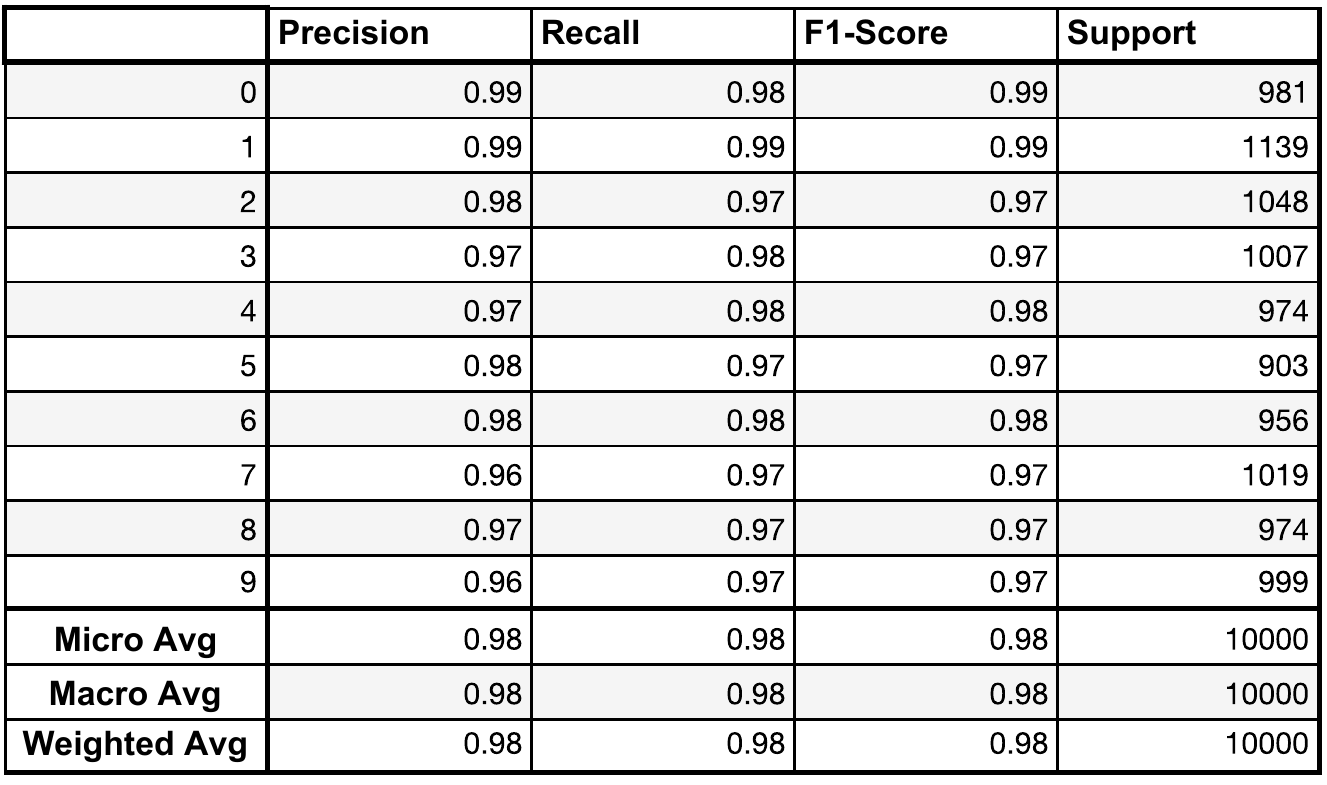}
	\caption{Confusion matrix and classification report of the neural network model with autoencoder}
	\label{fig:NN_AE_AE}
\end{figure}

We examine the differences between the neural network model that has autoencoder and the neural network model that takes data directly from the MNIST dataset with confusion matrixes and classification reports. Firstly, our model without autoencoder gives the following results, as seen in Figure \ref{fig:NN_AE_WOAE} for the confusion matrix and the classification report. The results with the autoencoder are presented in Figure \ref{fig:NN_AE_AE}. Note that these confusion matrixes and classification reports are indicated before any attack.

\textbf{Fast Gradient Sign Method:}

There is a slight difference between the neural network models with autoencoder and without autoencoder model. We apply the FGSM attack on both methods. The method uses the gradients of the loss accordingly for creating a new image that maximizes the loss. We can say the gradients are generated accordingly to input images. For these reasons, the FGSM causes a wide variety of models to misclassify their input \cite{paper16}.

\begin{figure}[htbp!]
	\centering
	\includegraphics[scale=0.56]{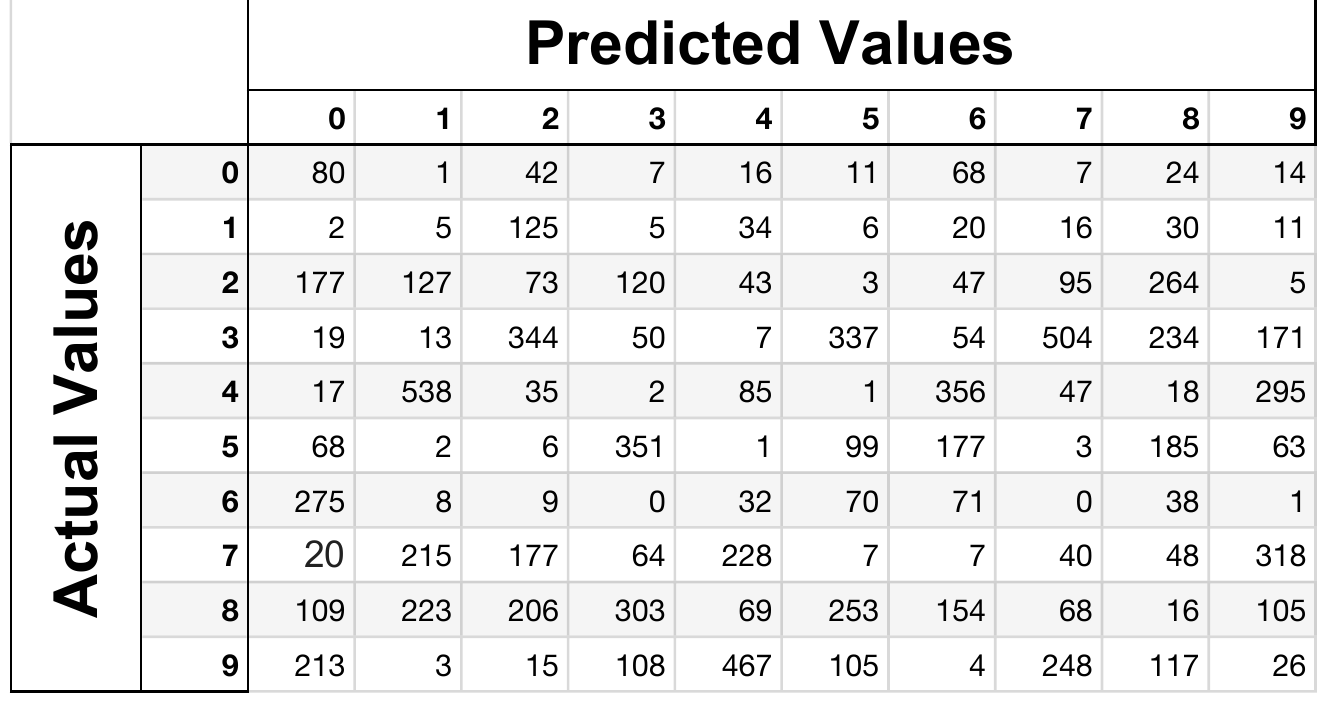}
	\includegraphics[scale=0.5]{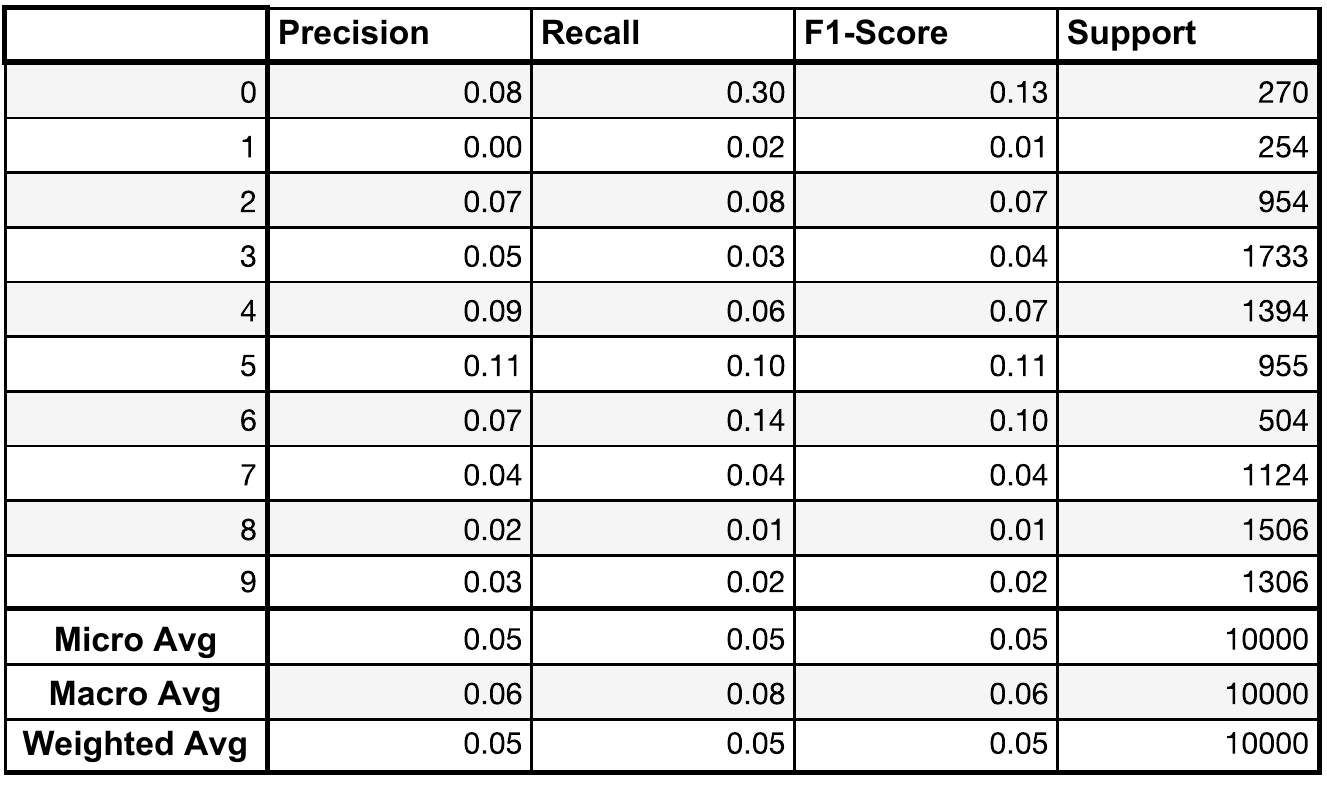}
	\caption{Confusion matrix and classification report of the neural network model without autoencoder after FGSM attack}
	\label{fig:NN_AE_WOAE_FGSM}
\end{figure}

\begin{figure}[htbp!]
	\centering
	\includegraphics[scale=0.56]{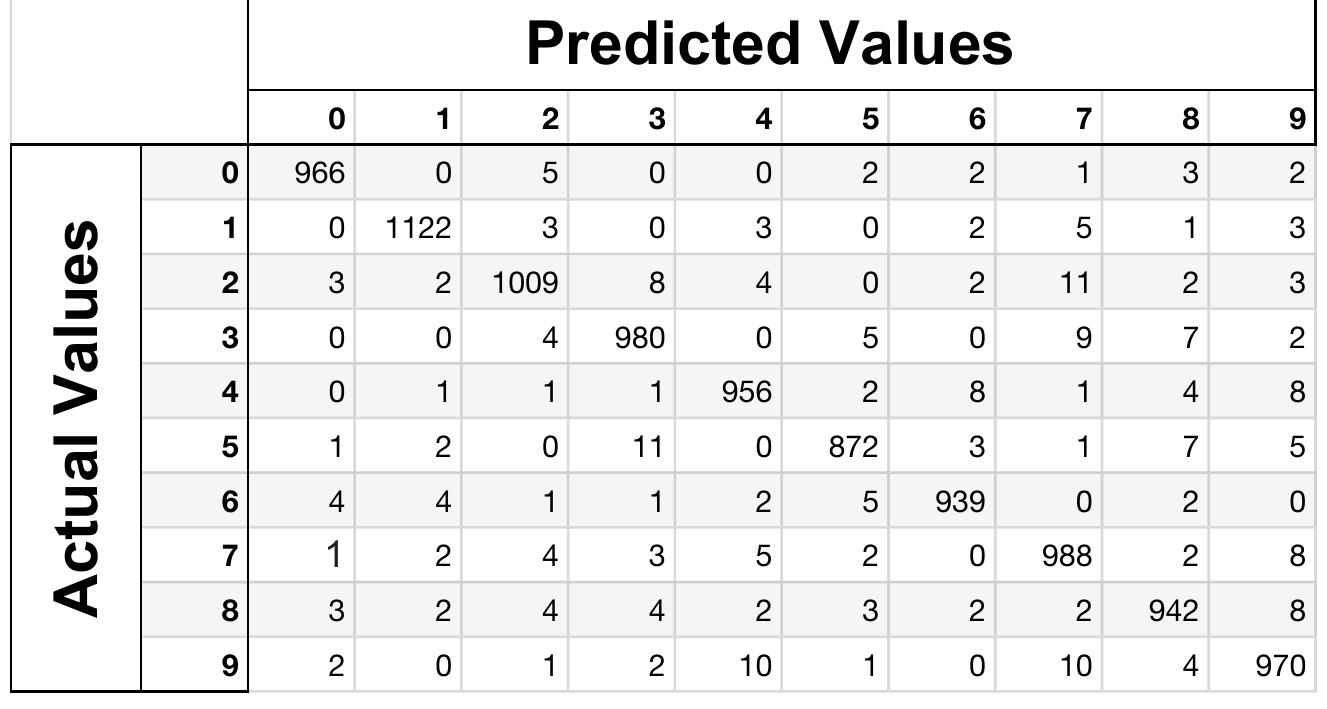}
	\includegraphics[scale=0.5]{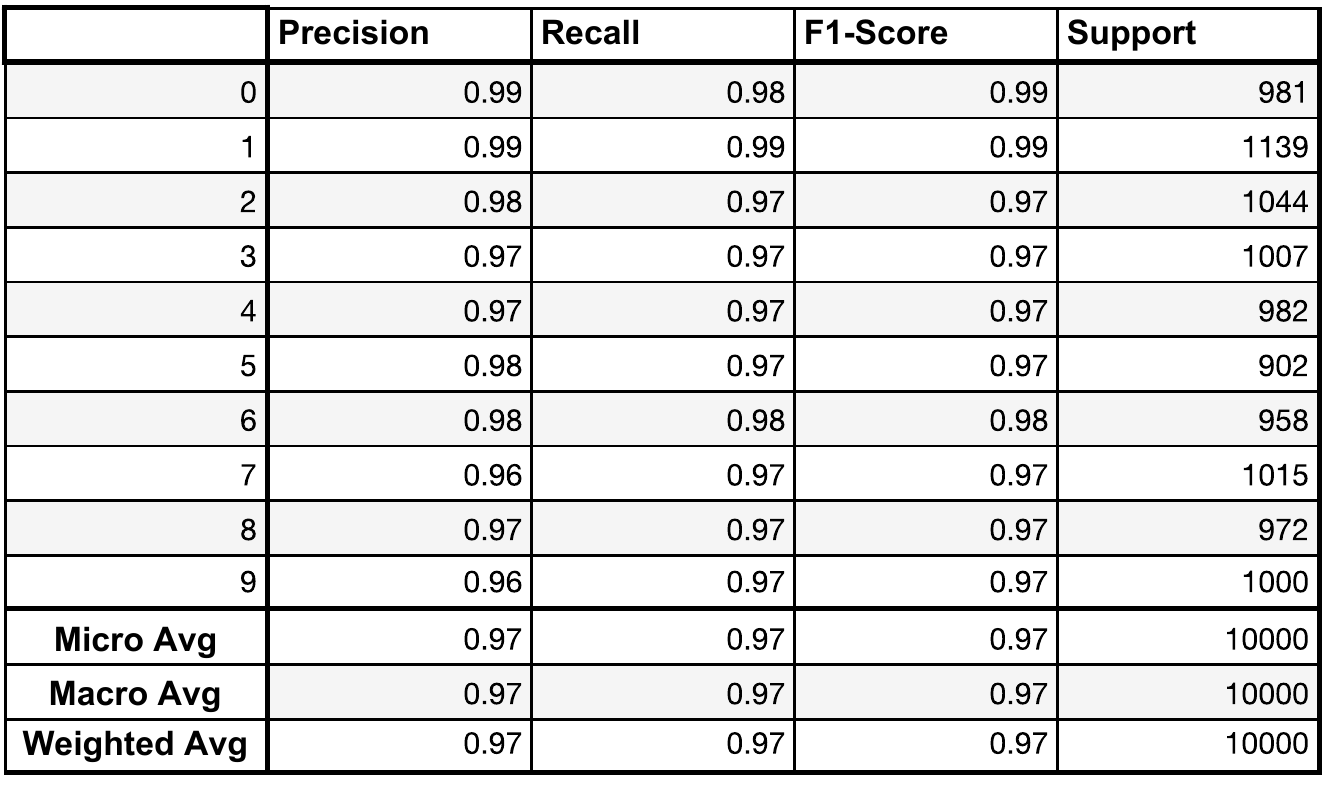}
	\caption{Confusion matrix and classification report of the neural network model with autoencoder after FGSM attack}
	\label{fig:NN_AE_AE_FGSM}
\end{figure}

As we expect due to results from multi-class logistic regression, autoencoder gives robustness to the neural network model too. After the DGSM, the neural network without an autoencoder suffers an immense drop in its accuracy, and the FGSM works as intended. But the neural network model with autoencoder only suffers a 0.01 percent accuracy drop. 

\textbf{Targeted Fast Gradient Sign Method:}
There is a directed type of FGSM, called T-FGSM. It uses the same principles to maximize the loss of the target. In this method, a gradient step is computed for giving the same misprediction for different inputs.

\begin{figure}[htbp!]
	\centering
	\includegraphics[scale=0.56]{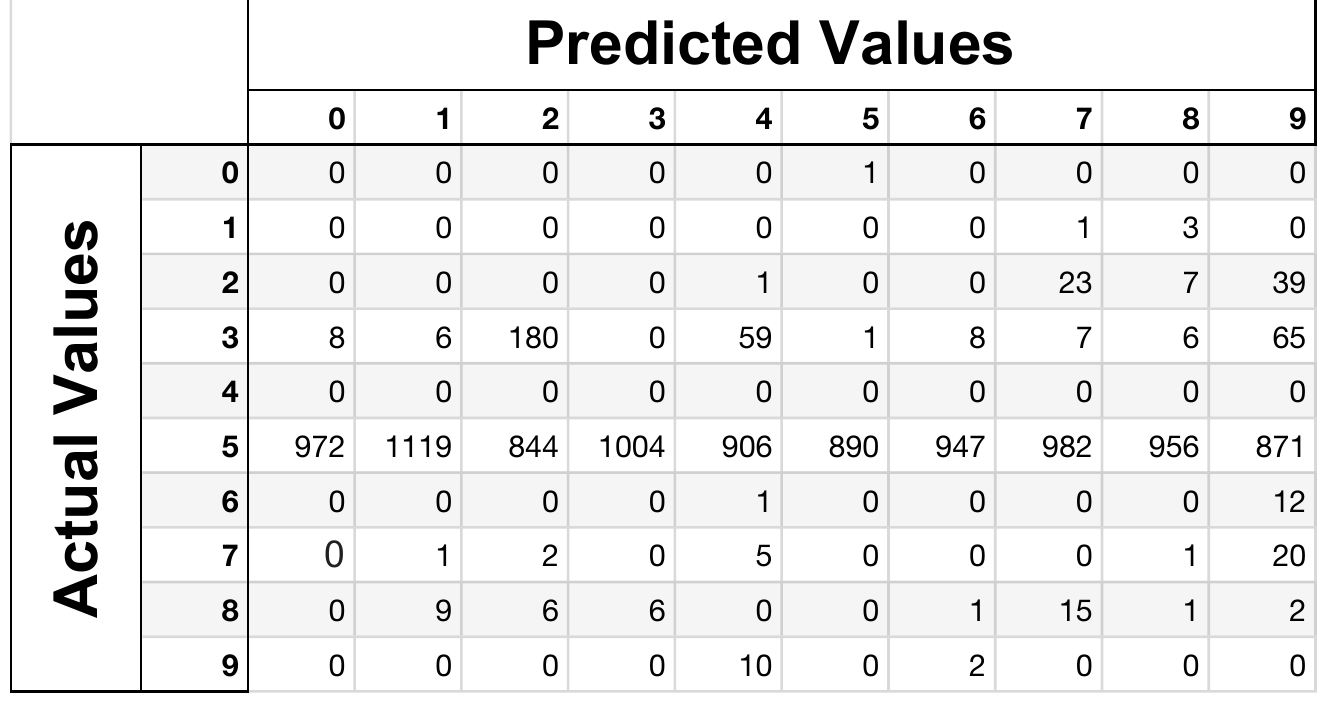}
	\includegraphics[scale=0.5]{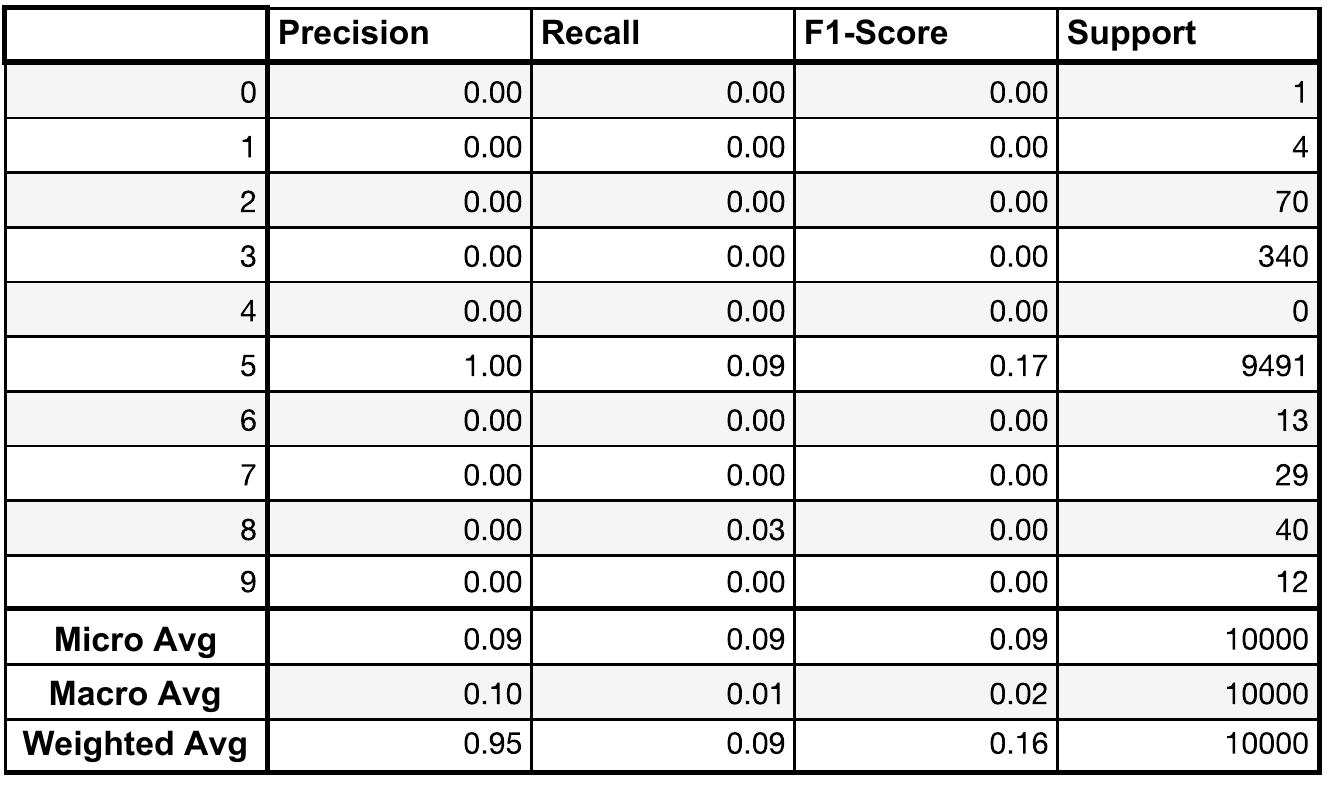}
	\caption{Confusion matrix and classification report of the neural network model without autoencoder after T-FGSM attack}
	\label{fig:NN_AE_WOAE_TFGSM}
\end{figure}

\begin{figure}[htbp!]
	\centering
	\includegraphics[scale=0.56]{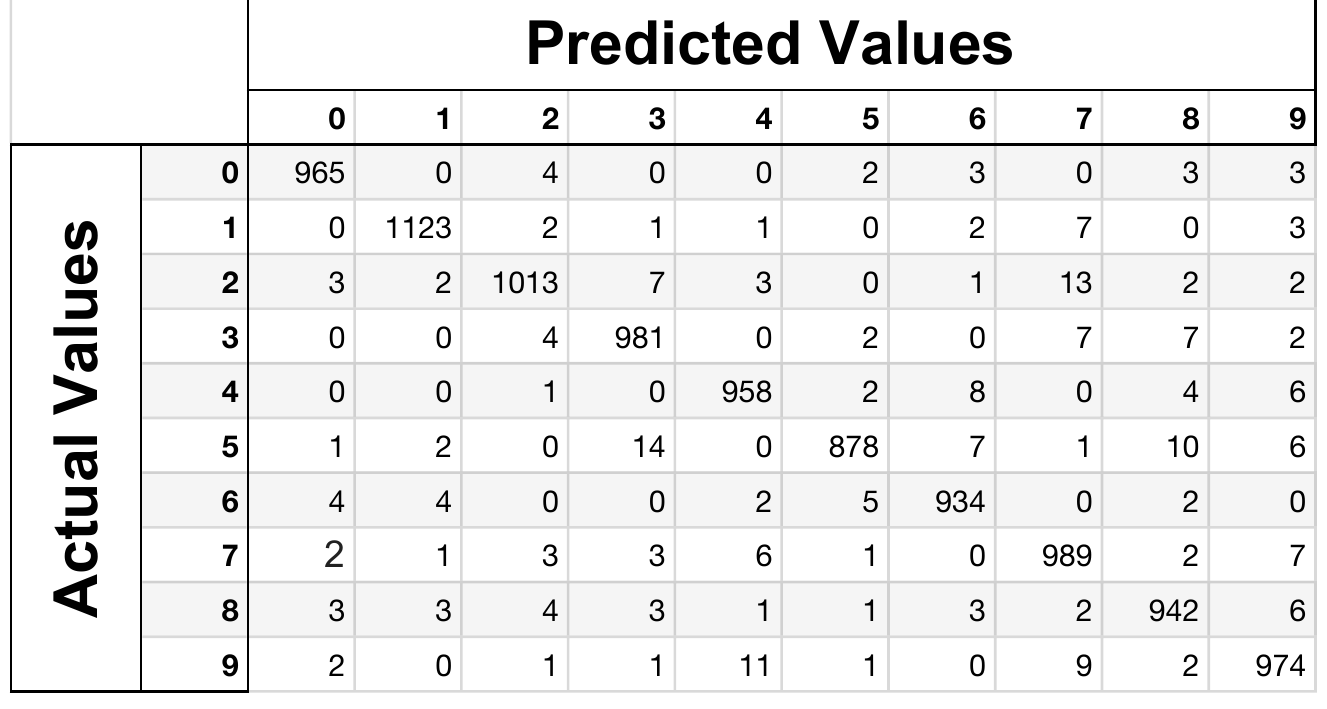}
	\includegraphics[scale=0.5]{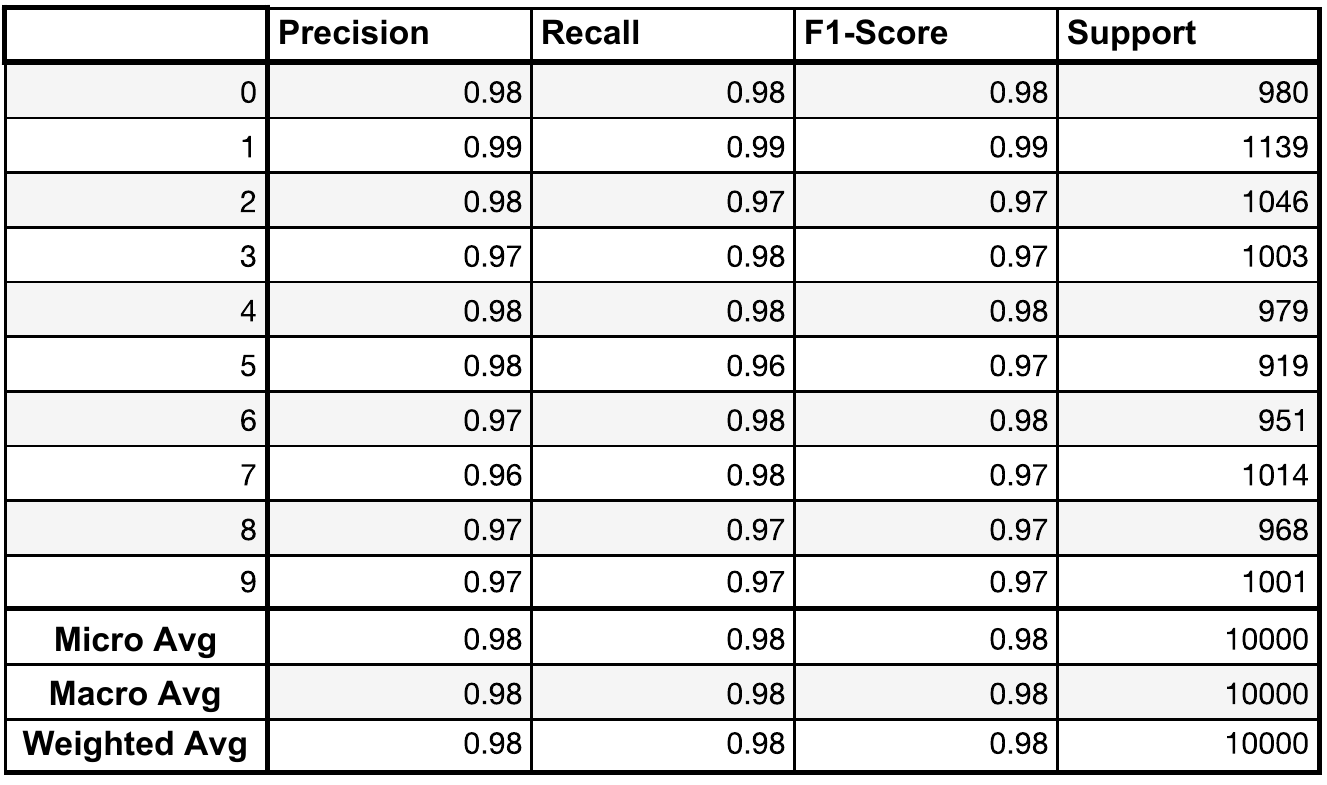}
	\caption{Confusion matrix and classification report of the neural network model with autoencoder after T-FGSM attack}
	\label{fig:NN_AE_AE_TFGSM}
\end{figure}

In the confusion matrix, the target value for this attack is number 5. The neural network model with the autoencoder is still at the accuracy of 0.98. The individual differences are presented when compare with Figure \ref{fig:NN_AE_AE}.

\textbf{Basic Iterative Method:}

BIM is an extension of FGSM to apply it multiple times with iterations. It provides the recalculation of a gradient attack for each iteration.

\begin{figure}[htbp!]
	\centering
	\includegraphics[scale=0.56]{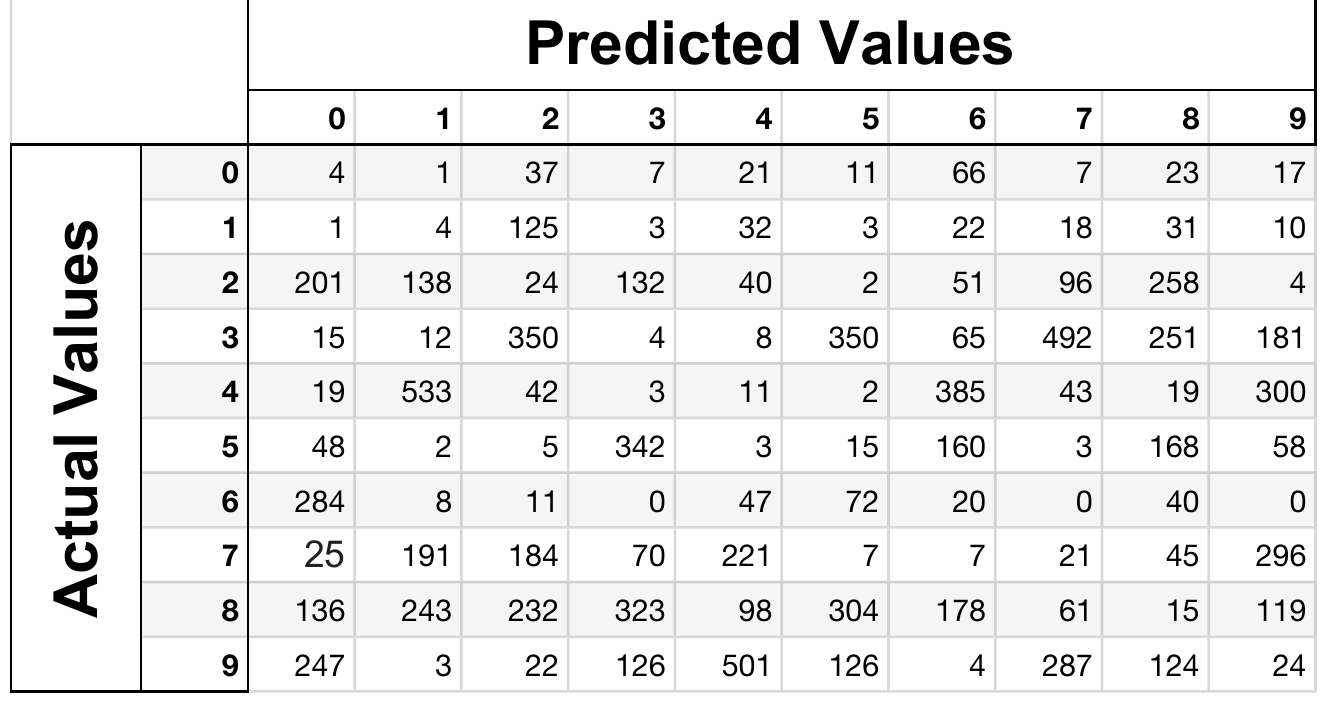}
	\includegraphics[scale=0.5]{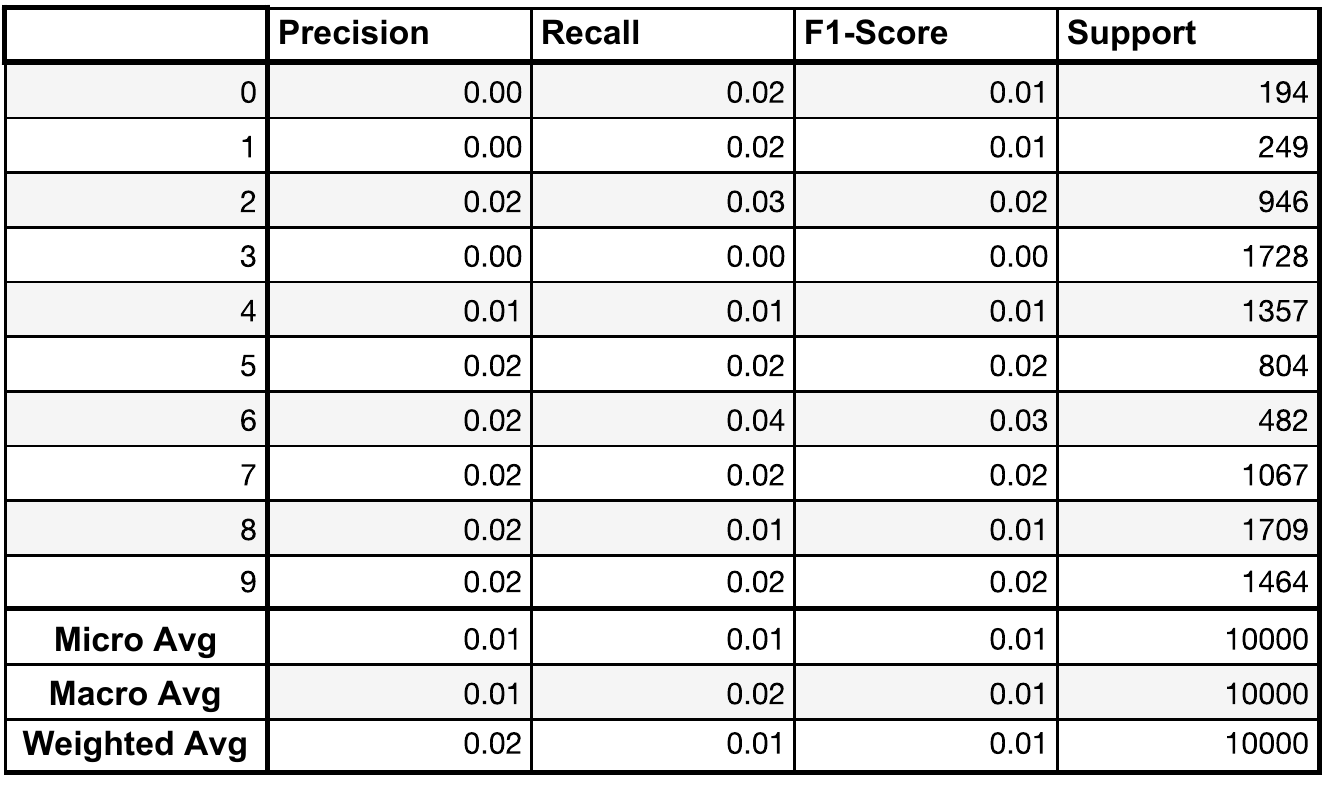}
	\caption{Confusion matrix and classification report of the neural network model without autoencoder after basic iterative method attack}
	\label{fig:NN_AE_WOAE_BIM}
\end{figure}

\begin{figure}[htbp!]
	\centering
	\includegraphics[scale=0.56]{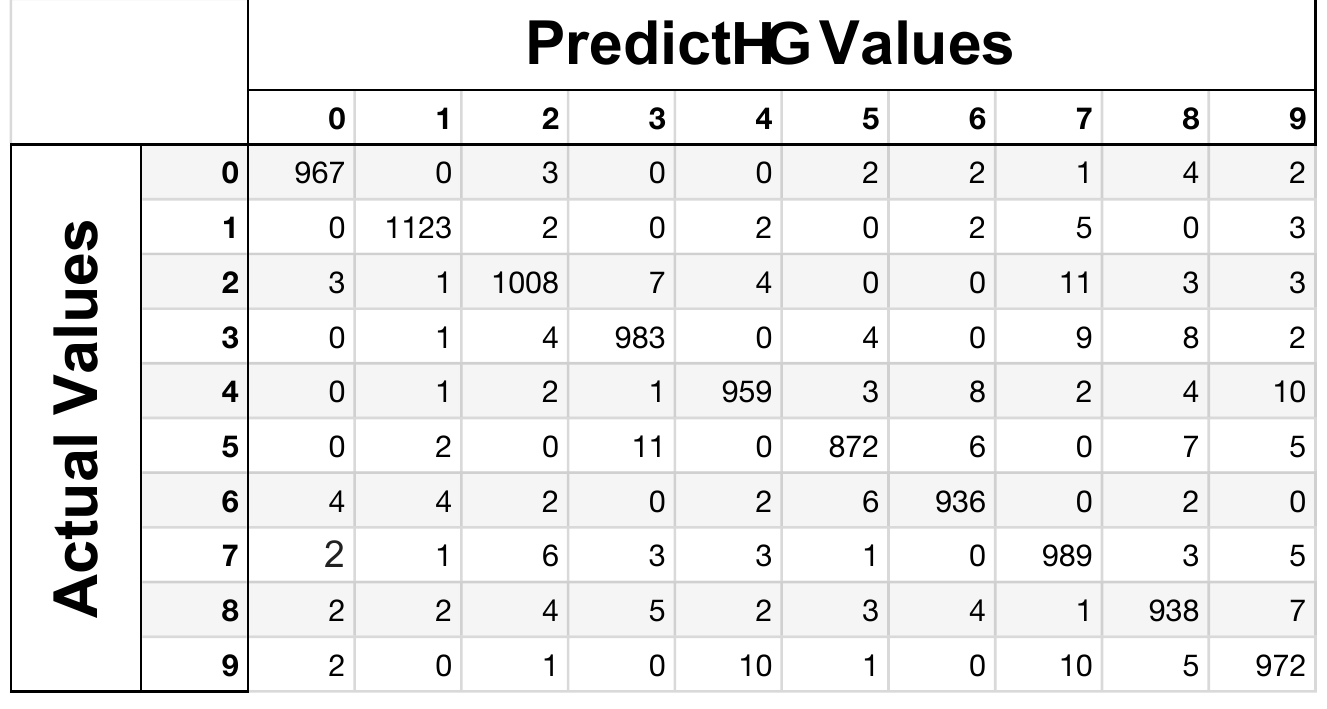}
	\includegraphics[scale=0.5]{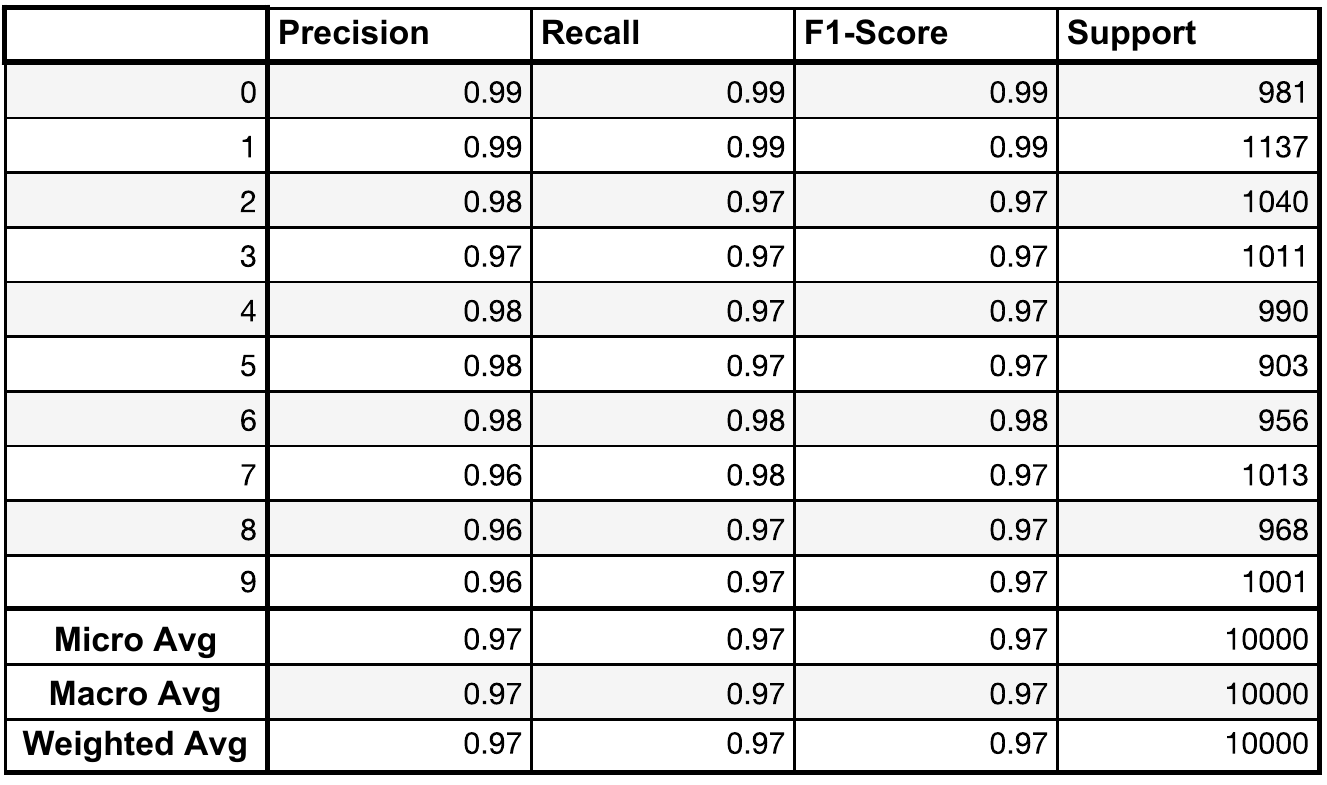}
	\caption{Confusion matrix and classification report of the neural network model with autoencoder after basic iterative method attack}
	\label{fig:NN_AE_AE_BIM}
\end{figure}

This is the most damaging attack for the neural network model that takes its inputs directly from the MNIST Dataset without an autoencoder. The findings from Figure \ref{fig:NN_AE_WOAE_BIM} show that the accuracy drops between 0.01 and 0.02 percent. The neural network model with autoencoder’s accuracy stays as 0.97 percent, losing only 0.1 percent.

Findings indicate that autoencoding before giving dataset as input to linear models and neural network models improve robustness against adversarial attacks significantly. We use vanilla autoencoders. They are the basic autoencoders without modification. In the other sections, we apply the same attacks with the same machine learning models with different autoencoder types.

\subsection{Sparse Autoencoder}

Sparse autoencoders present improved performance on classification tasks. It includes more hidden layers than the input layer. The significant part is defining a small number of hidden layers to be active at once to encourage sparsity. This constraint forces the training model to respond uniquely to the characteristics of translation and uses the statistical features of the input data.

Because of this sparse autoencoders involve sparsity penalty $\Omega(h)$ in their training.
$L(x,x')+\Omega(h)$

This penalty makes the model to activate specific areas of the network depending on the input data while making all other neurons inactive. We can create this sparsity by relative entropy, also known as Kullback-Leibler divergence.

$\widehat{\rho}_j= \frac{1}{m}\sum_{i=1}^m[h_j(x_i)]$
$\widehat{\rho}_j$ is our average activation function of the hidden layer $j$ which is averaged over $m$ training examples. For increasing the sparsity in terms of making the number of active neurons as smaller as it can be, we would want $\rho$ close to zero. The sparsity penalty term $\Omega(h)$ will punish $\widehat{\rho}_j$ for deviating from $\rho$, which will be basically exploiting Kullback-Leibler divergence. $KL(p||\widehat{\rho}_j)$ is our Kullback-Leibler divergence between a random variable $\rho$ and random variable with mean $\widehat{\rho}_j$.

$\sum_{j=1}^sKL(\rho||\widehat{\rho}_j) = \sum_{j=1}^s[\rho log\frac{\rho}{\widehat{\rho}_j}+(1-\rho)log\frac{1-\rho}{1-\widehat{\rho}_j}] $

Sparsity can be achieved with other ways, such as applying L1 and L2 regularization terms on the activation of the hidden layer. $L$ is our loss function and $\lambda$ is our scale parameter.

$L(x,x') + \lambda\sum_i|h_i|$

\subsubsection{Multi-Class Logistic Regression of Sparse Autoencoder}

\begin{figure}[htbp!]
	\centering
	\includegraphics[width=0.5\linewidth]{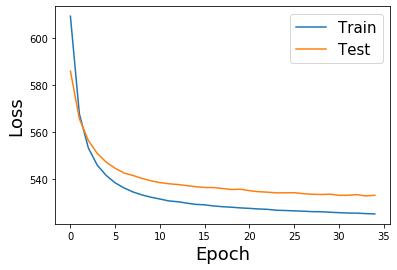}
	\caption{Optimized Relu Loss History for Sparse Autoencoder}
	\label{fig:lossSAE}
\end{figure}

This section presents multi-class logistic regressions with sparse autoencoders. The difference from the autoencoder section is the autoencoder type. The findings from Figure \ref{fig:optimizedRelu} and Figure \ref{fig:lossSAE} show that loss is higher compared to the autoencoders in sparse autoencoder.

\begin{figure}[htbp!]
	\includegraphics[width=1.0\linewidth]{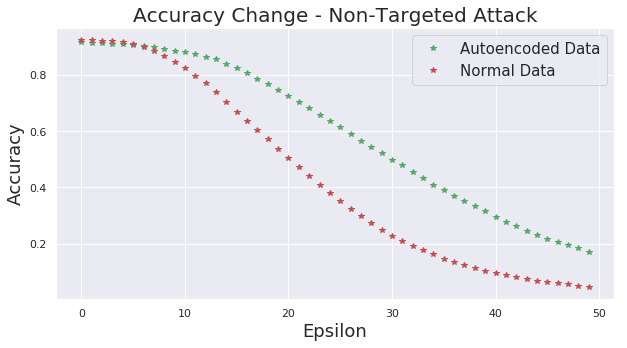}
	\caption{Comparison of accuracy with and without sparse autoencoder for non-targeted attack}
	\label{fig:nontargetedAttackSAE}
\end{figure}

\begin{figure}[htbp!]
	\centering
	\includegraphics[width=0.75\linewidth]{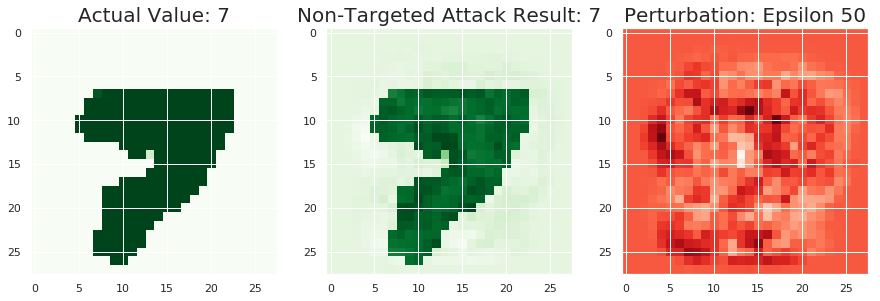}
	\caption{Value change and perturbation of a non-targeted attack on model without sparse autoencoder}
	\label{fig:perturbationUnsuccessSAE}
\end{figure}

\begin{figure}[htbp!]
	\centering
	\includegraphics[width=0.75\linewidth]{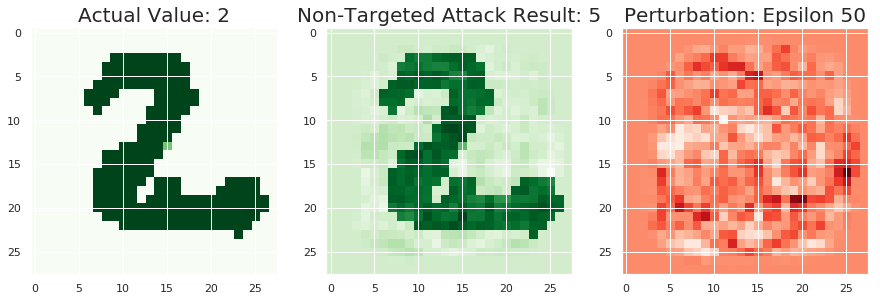}
	\caption{Value change and perturbation of a non-targeted attack on model with sparse autoencoder}
	\label{fig:perturbationSuccessSAE}
\end{figure}

The difference between perturbation is presented in Figure \ref{fig:perturbationUnsuccessSAE} and Figure \ref{fig:perturbationSuccessSAE} compared to the perturbation in Figure \ref{fig:perturbasyonWoAE} and Figure \ref{fig:perturbasyonAE}. The perturbation is sharper in sparse autoencoder.

\begin{figure}[htbp!]
	\includegraphics[width=1.0\linewidth]{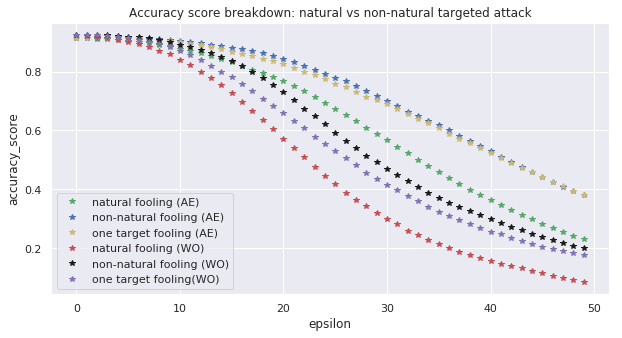}
	\caption{Comparison of accuracy with and without sparse autoencoder for targeted attacks. \textit{AE stands for the models with sparse autoencoder, WO stands for models without autoencoder}}
	\label{fig:targetedAttacksSAE}
\end{figure}

Figure \ref{fig:targetedAttacksSAE} indicates that sparse autoencoders performs poorly compared to autoencoders in multi-class logistic regression.

\subsubsection{Neural Network of Sparse Autoencoder}

Sparse autoencoder results for neural networks indicate that vanilla autoencoder seems to be slightly better than sparse autoencoders for neural networks. Sparse autoencoders do not perform as well in linear machine learning models, in our case, multi-class logistic regression.

\begin{figure}[htbp!]
	\centering
	\includegraphics[scale=0.56]{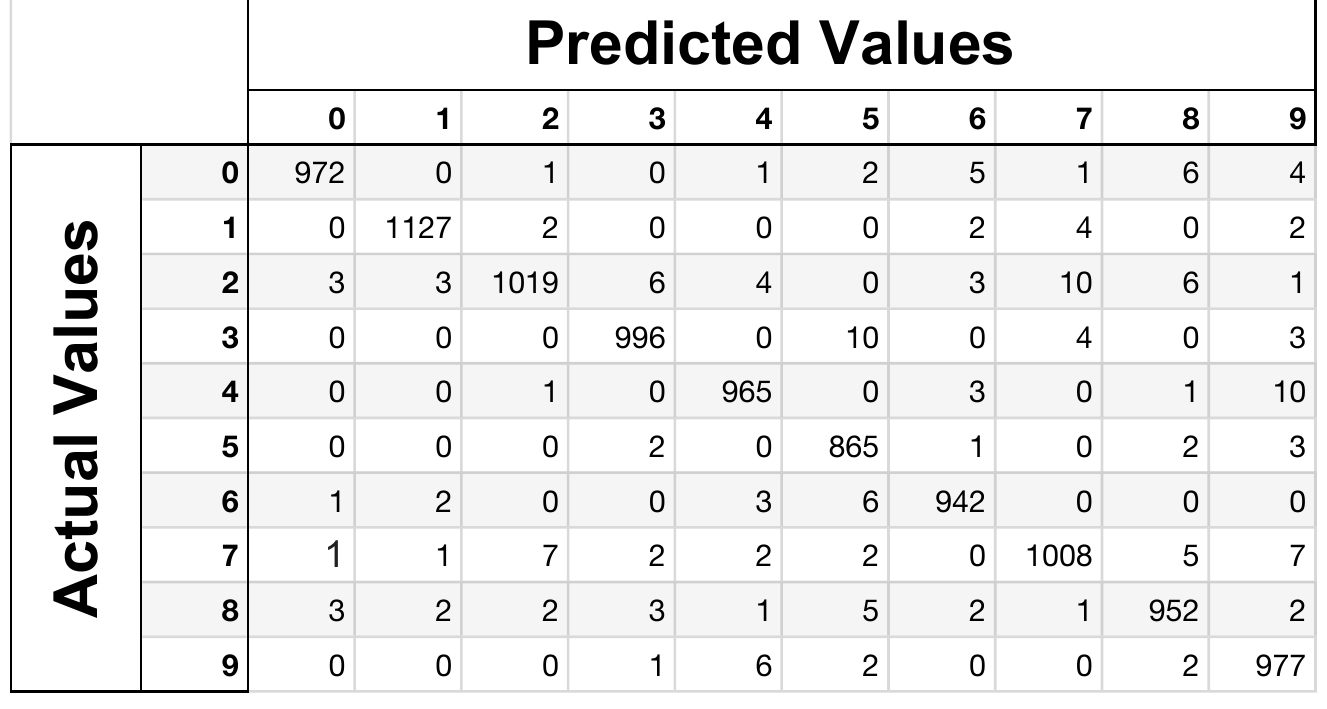}
	\includegraphics[scale=0.5]{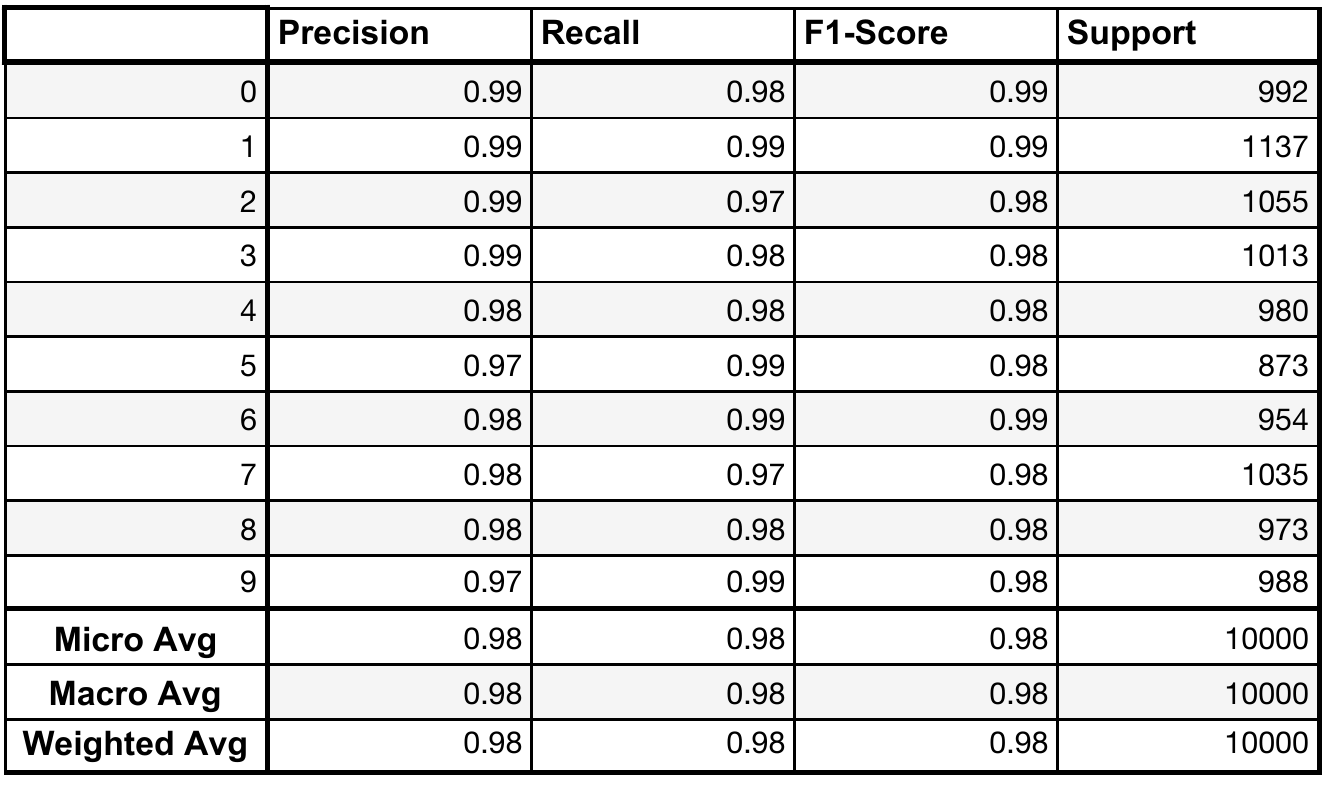}
	\caption{Confusion matrix and classification report of the neural network model without sparse autoencoder}
	\label{fig:NN_SAE_WOAE}
\end{figure}

\begin{figure}[htbp!]
	\centering
	\includegraphics[scale=0.56]{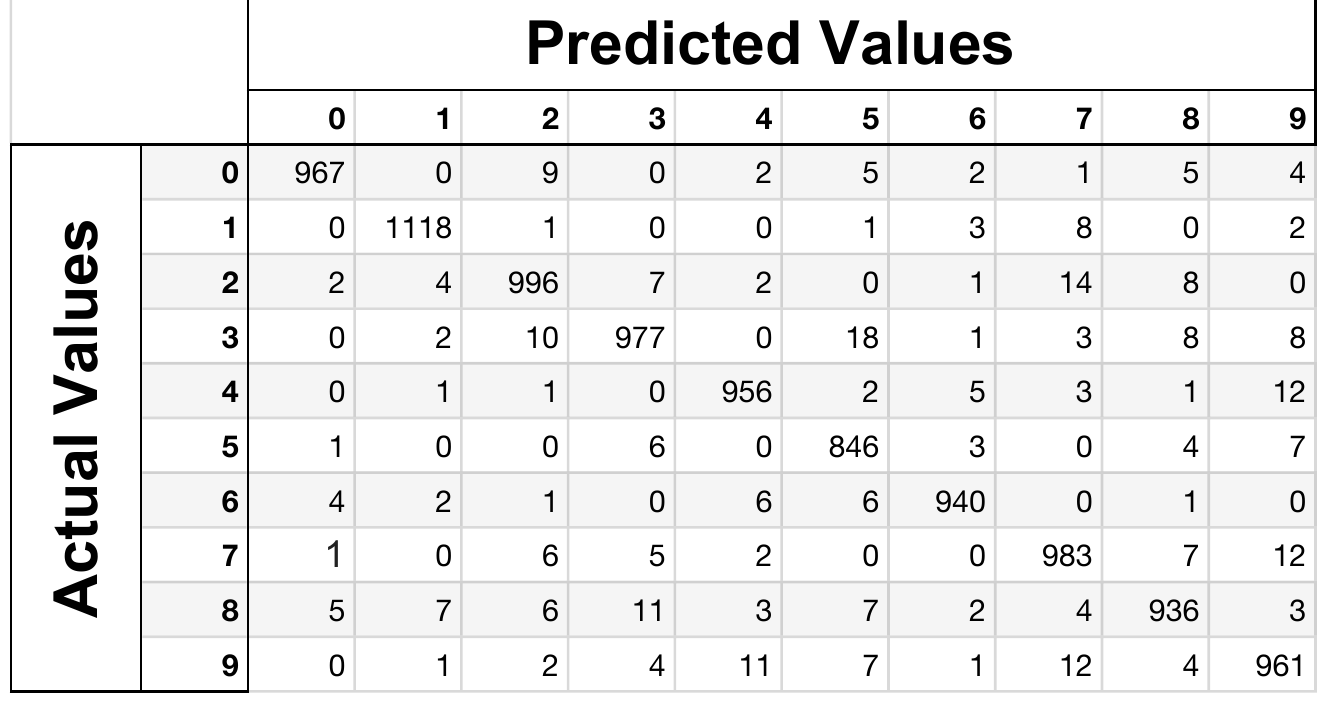}
	\includegraphics[scale=0.5]{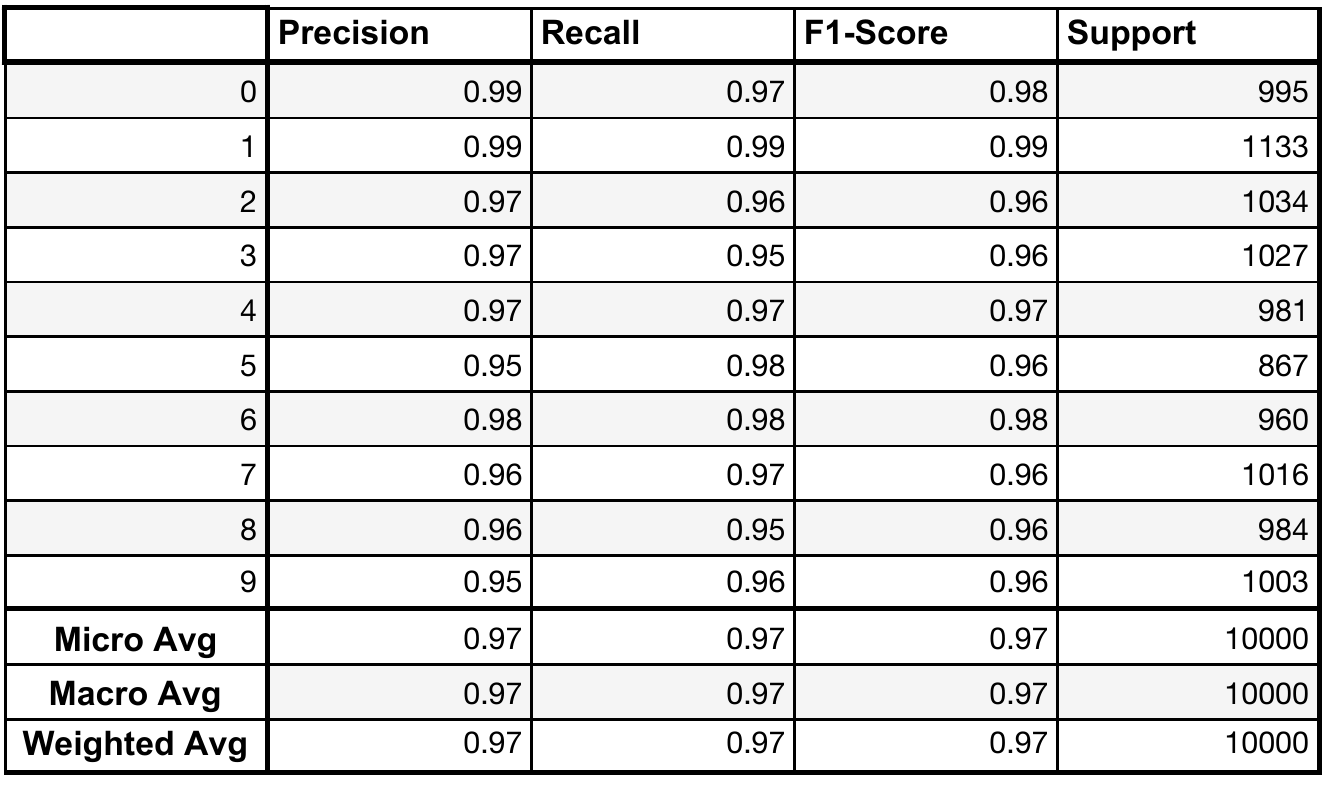}
	\caption{Confusion matrix and classification report of the neural network model with sparse autoencoder}
	\label{fig:NN_SAE_AE}
\end{figure}

\begin{figure}[htbp!]
	\centering
	\includegraphics[scale=0.56]{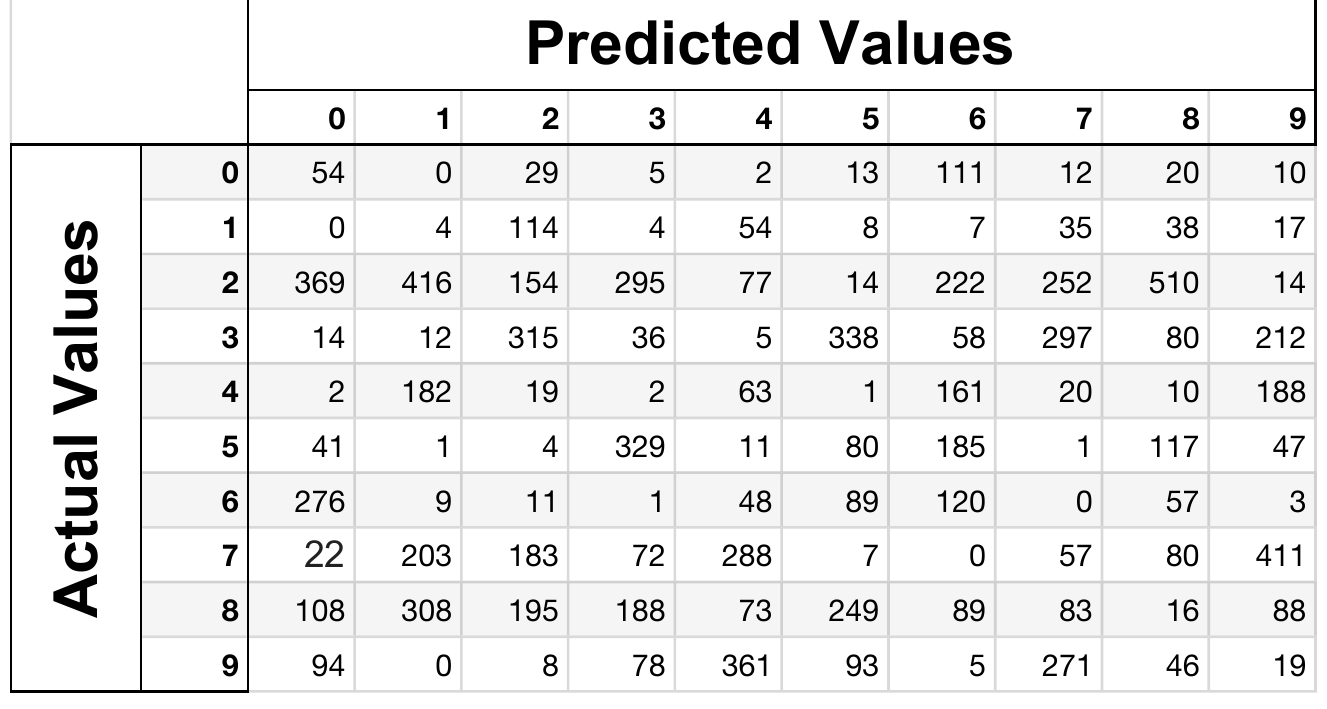}
	\includegraphics[scale=0.5]{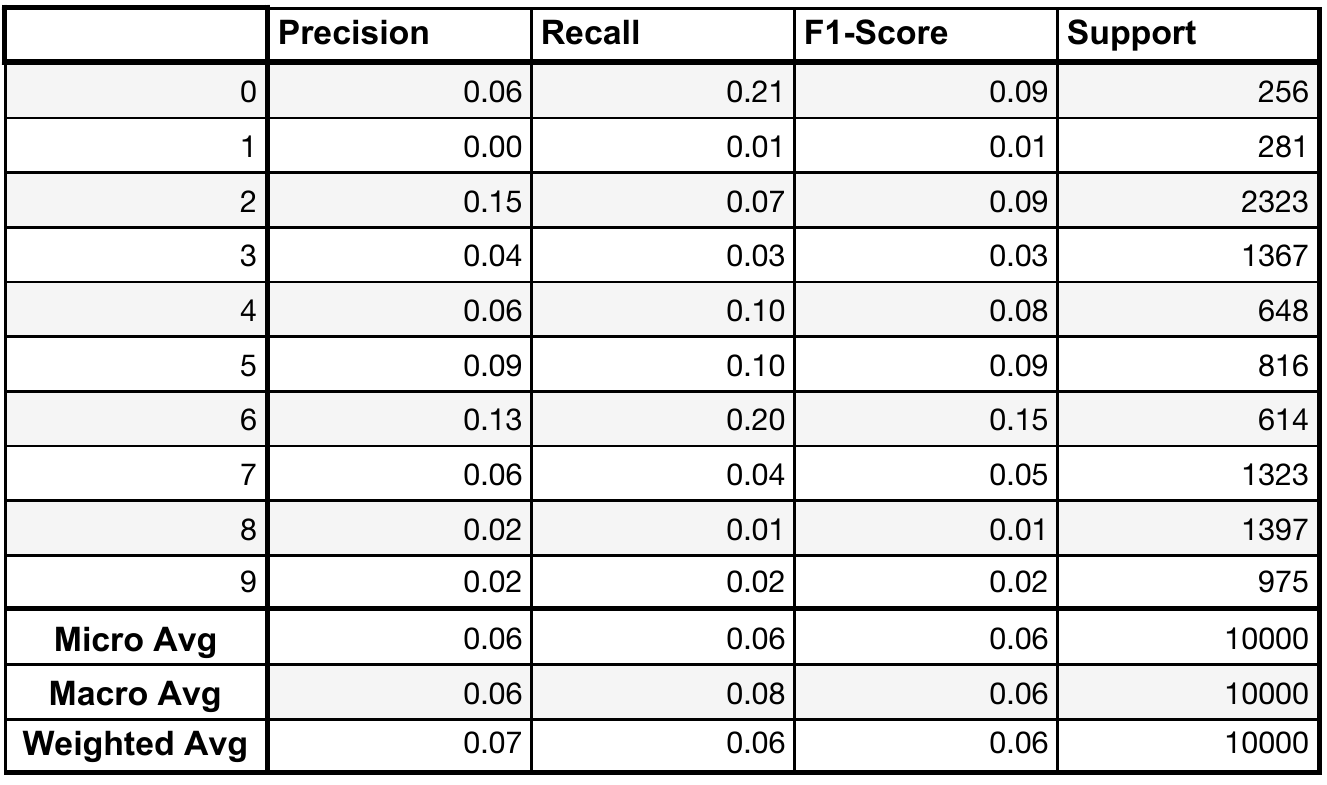}
	\caption{Confusion matrix and classification report of the neural network model without sparse autoencoder after FGSM attack}
	\label{fig:NN_SAE_WOAE_FGSM}
\end{figure}

\begin{figure}[htbp!]
	\centering
	\includegraphics[scale=0.56]{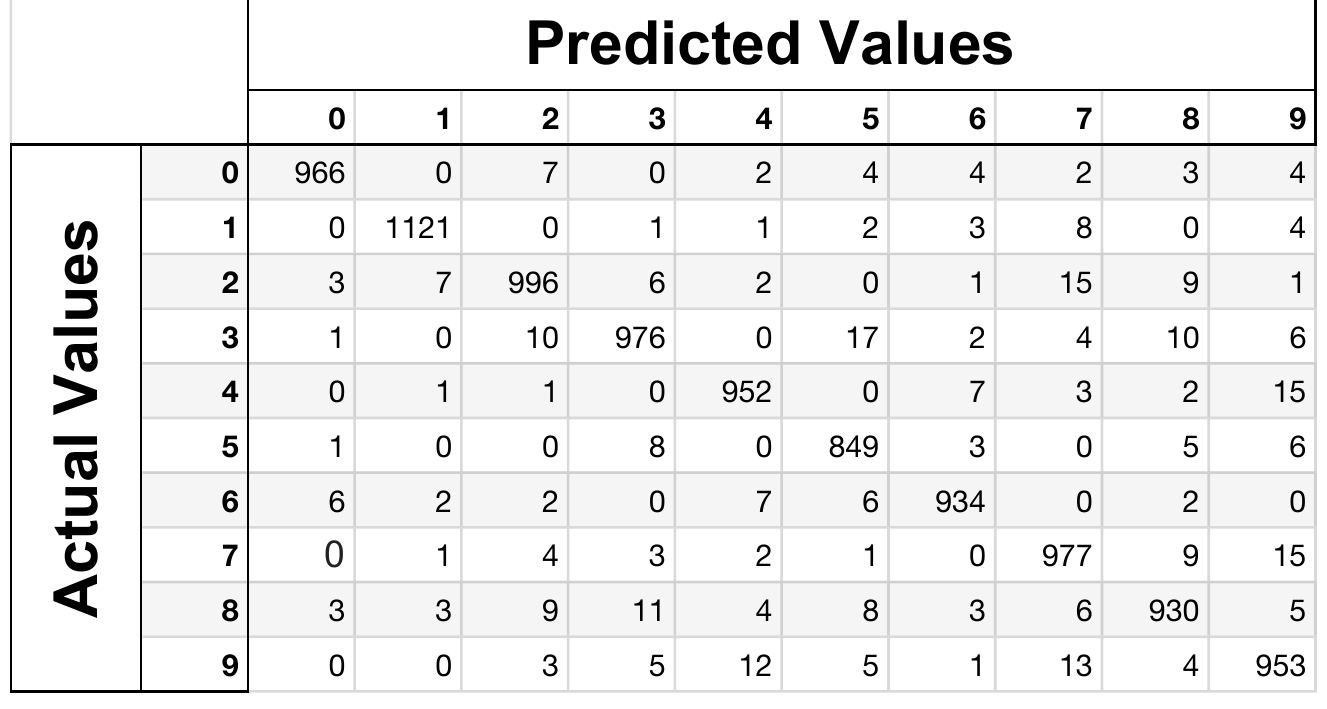}
	\includegraphics[scale=0.5]{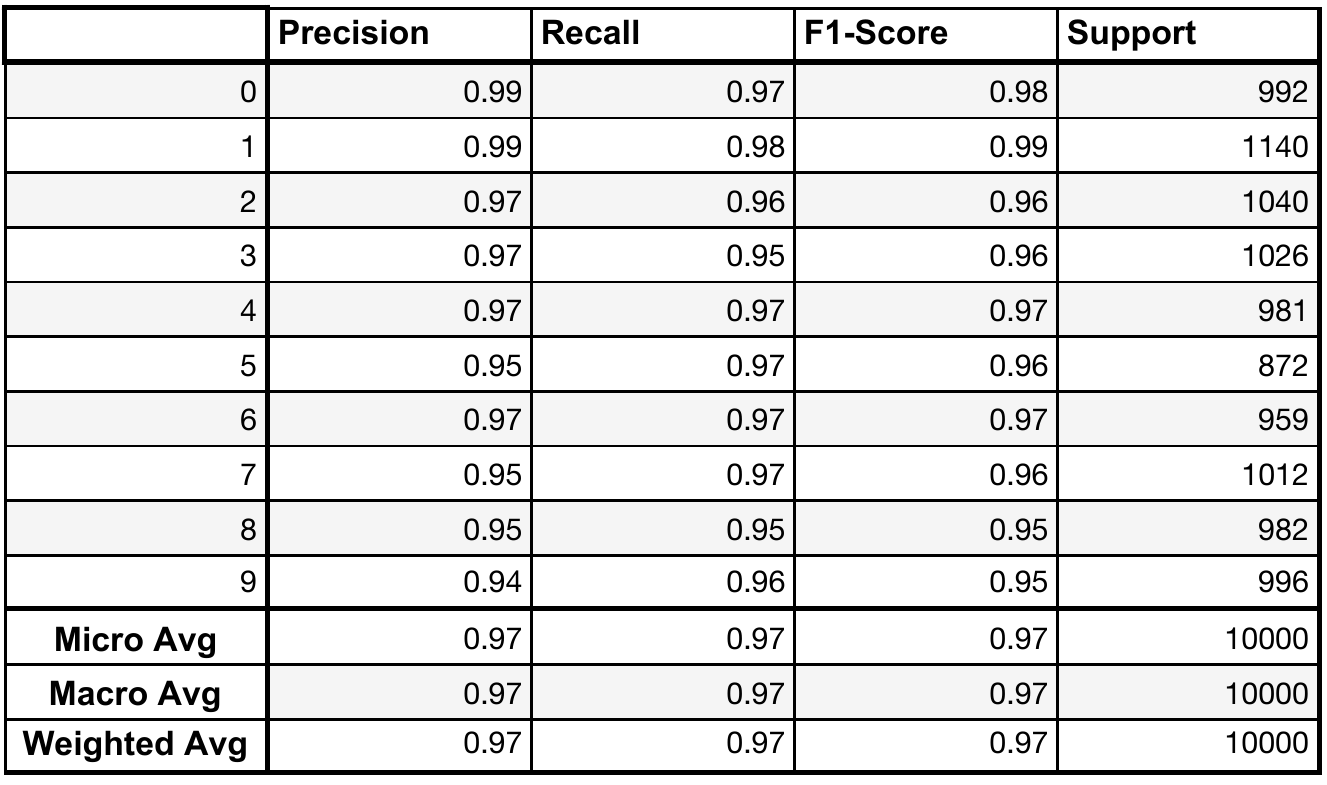}
	\caption{Confusion matrix and classification report of the neural network model with sparse autoencoder after FGSM attack}
	\label{fig:NN_SAE_AE_FGSM}
\end{figure}

\begin{figure}[htbp!]
	\centering
	\includegraphics[scale=0.56]{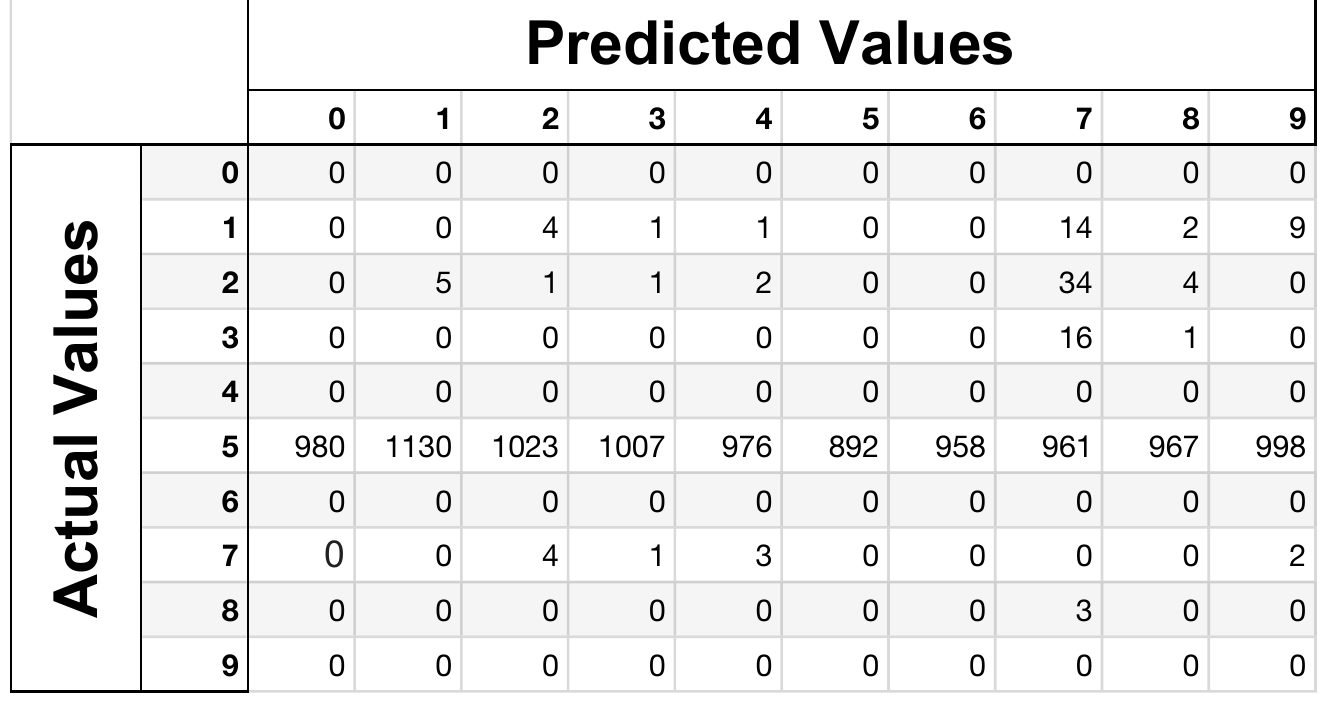}
	\includegraphics[scale=0.5]{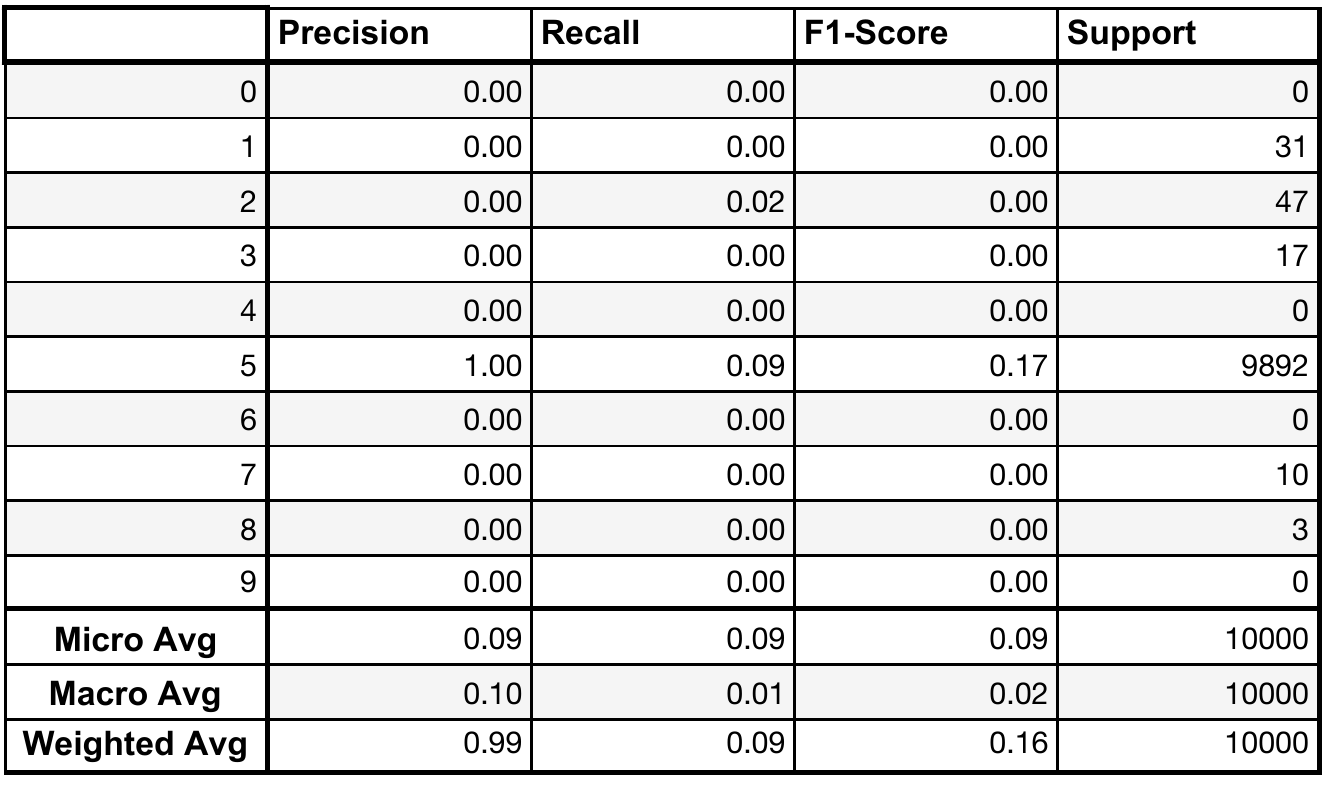}
	\caption{Confusion matrix and classification report of the neural network model without sparse autoencoder after T-FGSM attack}
	\label{fig:NN_SAE_WOAE_TFGSM}
\end{figure}

\begin{figure}[htbp!]
	\centering
	\includegraphics[scale=0.56]{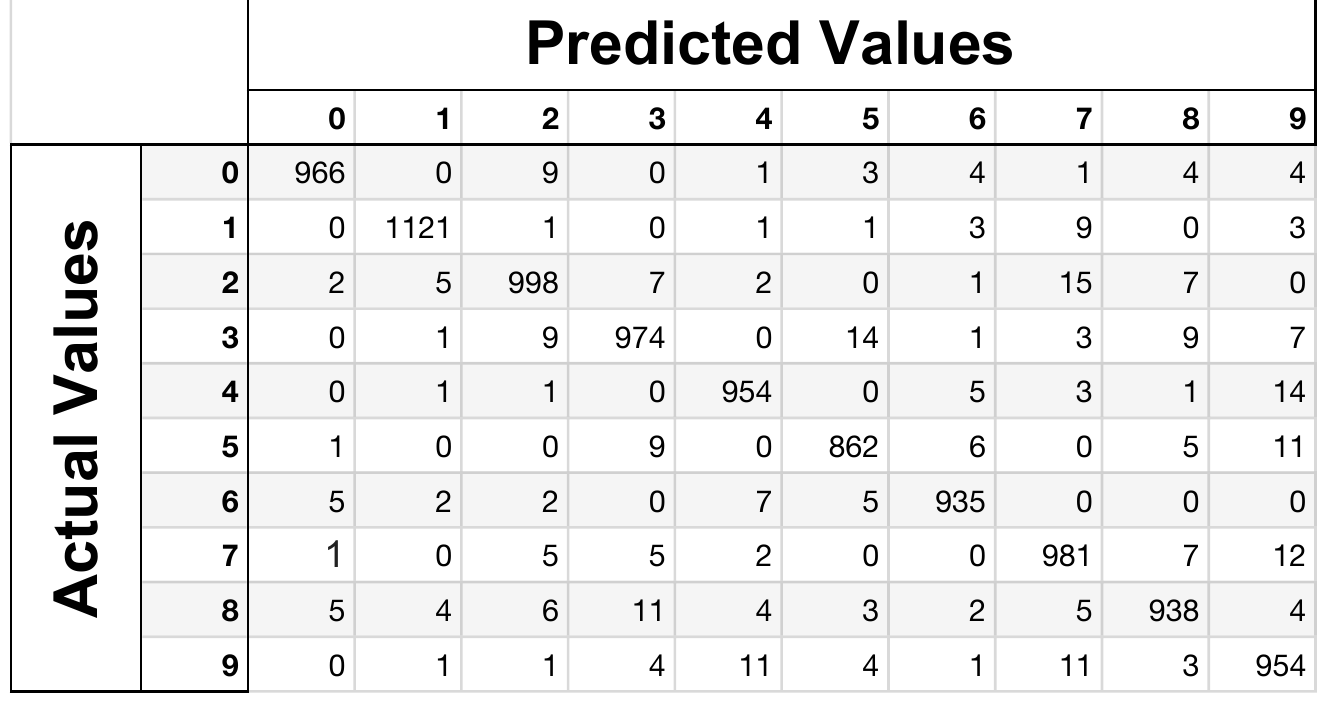}
	\includegraphics[scale=0.5]{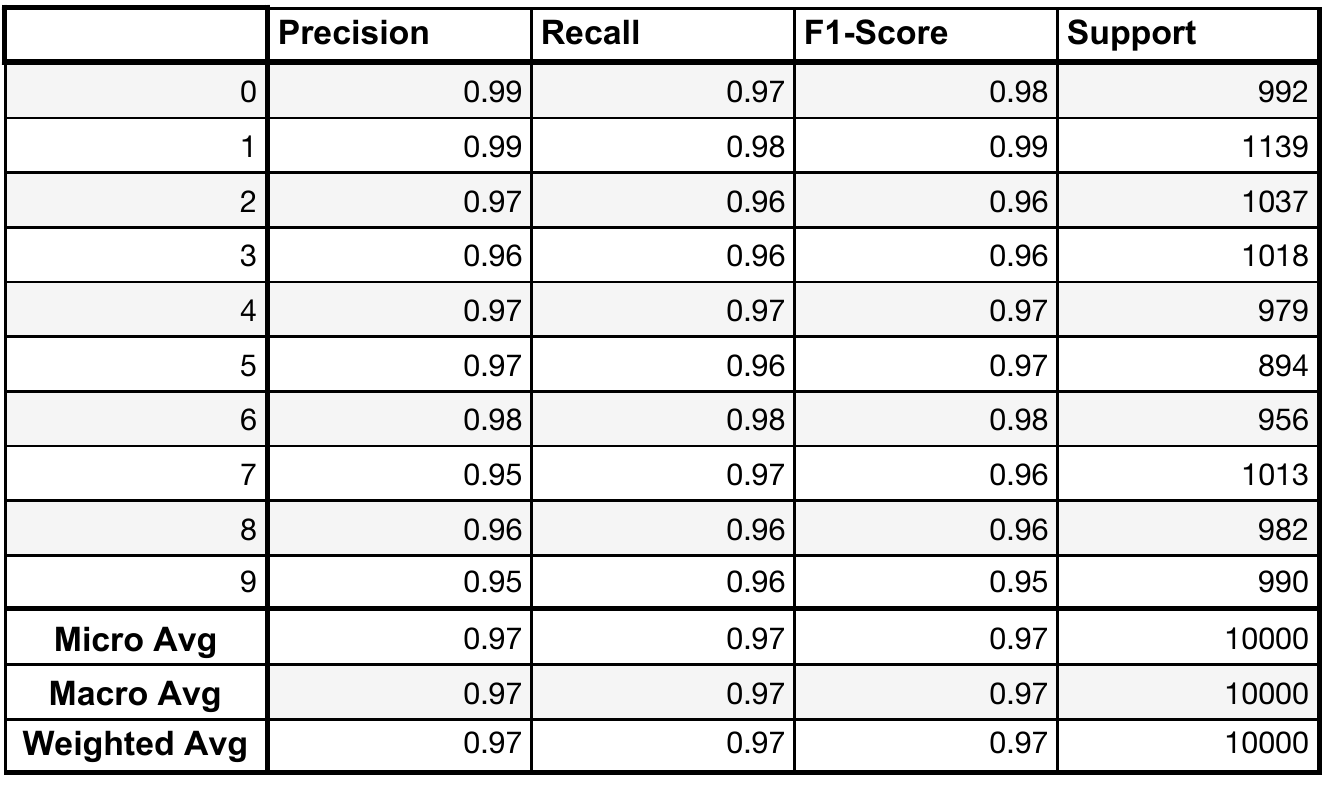}
	\caption{Confusion matrix and classification report of the neural network model with sparse autoencoder after T-FGSM attack}
	\label{fig:NN_SAE_AE_TFGSM}
\end{figure}

\begin{figure}[htbp!]
	\centering
	\includegraphics[scale=0.56]{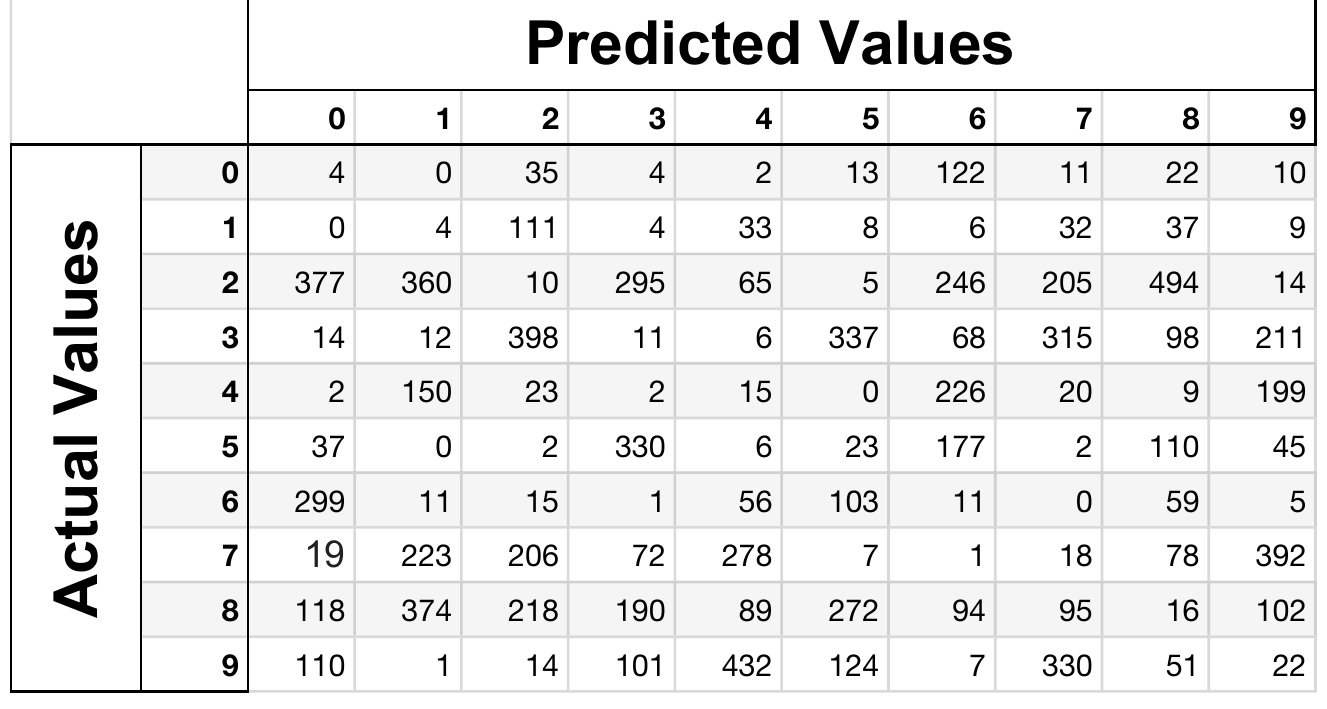}
	\includegraphics[scale=0.5]{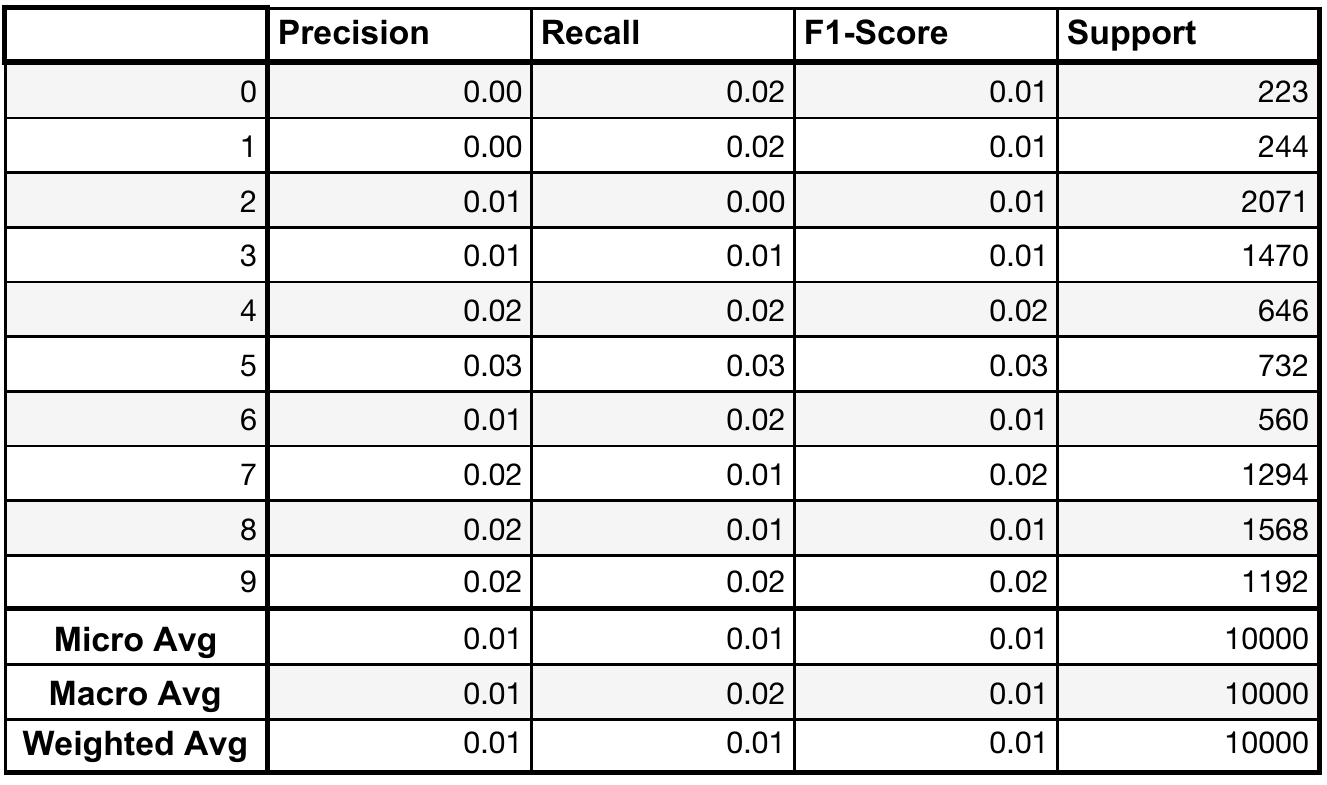}
	\caption{Confusion matrix and classification report of the neural network model without sparse autoencoder after basic iterative method attack}
	\label{fig:NN_SAE_WOAE_BIM}
\end{figure}

\begin{figure}[htbp!]
	\centering
	\includegraphics[scale=0.56]{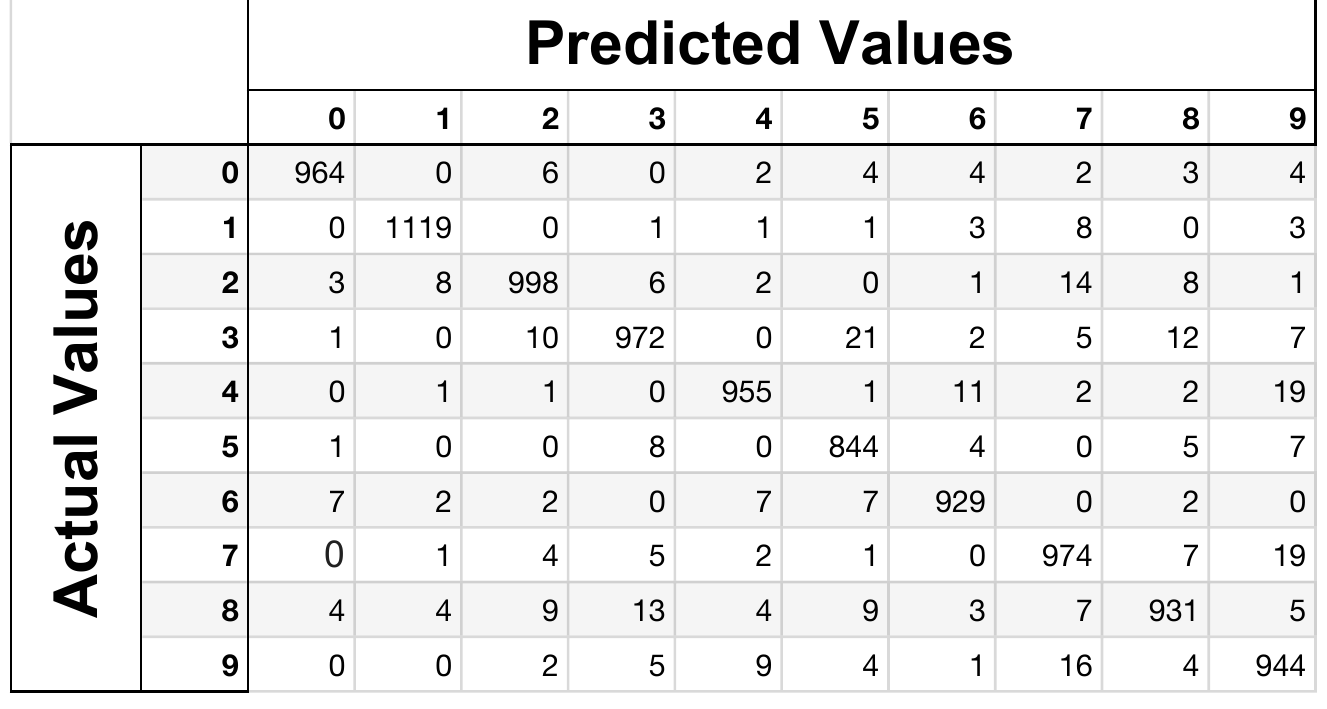}
	\includegraphics[scale=0.5]{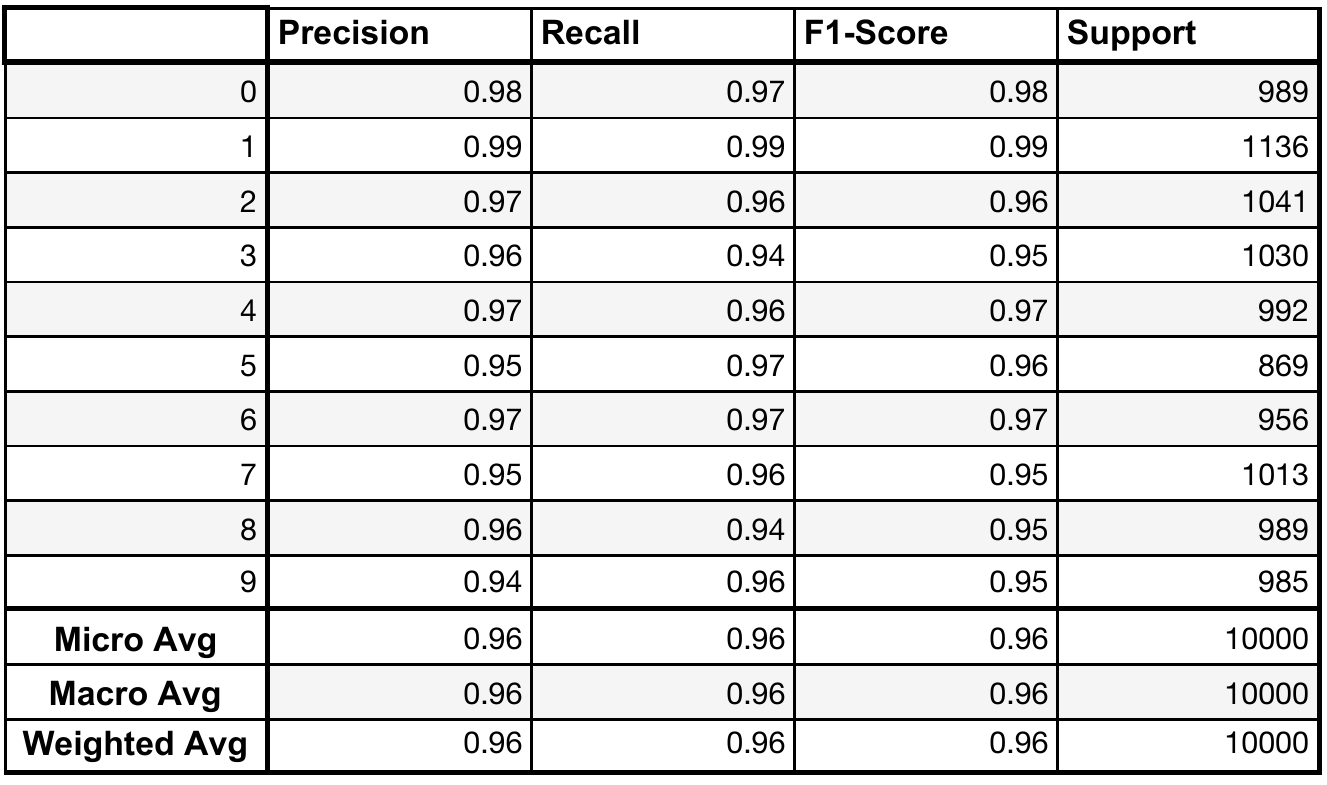}
	\caption{Confusion matrix and classification report of the neural network model with sparse autoencoder after basic iterative method attack}
	\label{fig:NN_SAE_AE_BIM}
\end{figure}

\subsection{Denoising Autoencoder}

Denoising autoencoders are used for partially corrupted input and train it to recover the original undistorted input. In this study, the corrupted input is not used. The aim is to achieve a good design by changing the reconstruction principle for using denoising autoencoders. For achieving this denoising properly, the model requires to extract features that capture useful structure in the distribution of the input. Denoising autoencoders apply corrupted data through stochastic mapping. Our input is $x$ and corrupted data is $\widetilde{x}$ and stochastic mapping is $\widetilde{x}  \sim  q_D(\widetilde{x}|x).$

As its a standard autoencoder, corrupted data $\widetilde{x}$ is mapped to a hidden layer. 

$h=f_\theta(\widetilde{x})= s(W\widetilde{x}+b).$

And from this the model reconstructs $z=g_\theta'(h)$.

\subsubsection{Multi-Class Logistic Regression of Denoising Autoencoder}

In denoising autoencoder for multi-class logistic regression, the loss does not improve for each epoch. Although it starts better at lower epoch values, in the end, vanilla autoencoder seems to be better. Sparse autoencoder's loss is slightly worse.

\begin{figure}[htbp!]
	\centering
	\includegraphics[width=0.5\linewidth]{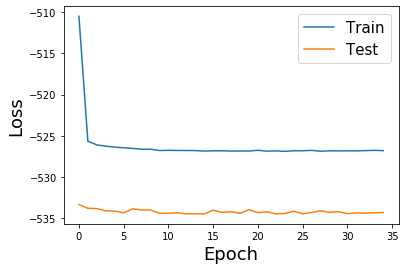}
	\caption{Optimized Relu Loss History for Denoising Autoencoder}
	\label{fig:lossDAE}
\end{figure}

\begin{figure}[htbp!]
	\includegraphics[width=1.0\linewidth]{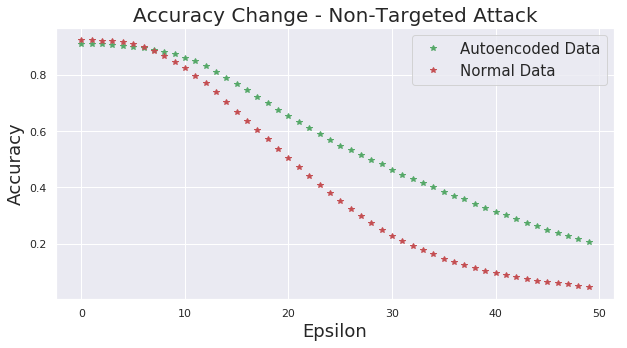}
	\caption{Comparison of accuracy with and without denoising autoencoder for non-targeted attack}
	\label{fig:nontargetedAttackDAE}
\end{figure}

\begin{figure}[htbp!]
	\centering
	\includegraphics[width=0.75\linewidth]{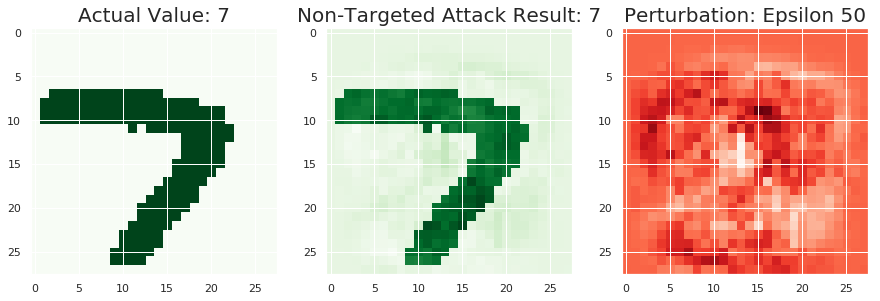}
	\caption{Value change and perturbation of a non-targeted attack on model without denoising autoencoder}
	\label{fig:perturbationUnsuccessDAE}
\end{figure}

\begin{figure}[htbp!]
	\centering
	\includegraphics[width=0.75\linewidth]{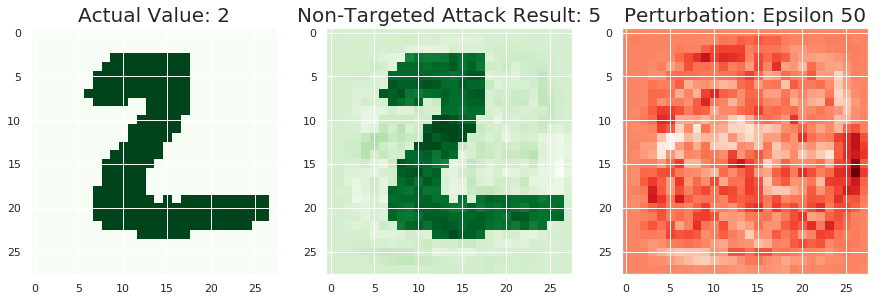}
	\caption{Value change and perturbation of a non-targeted attack on model with denoising autoencoder}
	\label{fig:perturbationSuccessDAE}
\end{figure}

And just like sparse autoencoder, denoising autoencoder also applies a sharp perturbation, which is presented in Figure \ref{fig:perturbationUnsuccessDAE} and Figure \ref{fig:perturbationSuccessDAE}.

We observe that there is a similarity between accuracy results for denoising autoencoder with multi-class logistic regression and sparse autoencoder results. Natural fooling accuracy drops drastically in denoising autoencoder, but non-targeted and one targeted attack seem to be somewhat like sparse autoencoder, one targeted attack having less accuracy in denoising autoencoder. 

\begin{figure}[htbp!]
	\includegraphics[width=1.0\linewidth]{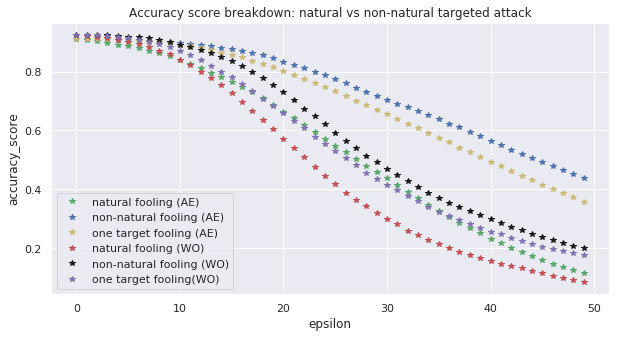}
	\caption{Comparison of accuracy with and without denoising autoencoder for targeted attacks. \textit{AE stands for the models with denoising autoencoder, WO stands for models without autoencoder}}
	\label{fig:targetedAttacksDAE}
\end{figure}


\subsubsection{Neural Network of Denoising Autoencoder}

We investigate that neural network accuracy for denoising autoencoder is worse than sparse autoencoder results and vanilla autoencoder results. It is still a useful autoencoder for denoising corrupted data and other purposes; however, it is not the right choice just for robustness against adversarial examples.

\begin{figure}[htbp!]
	\centering
	\includegraphics[scale=0.56]{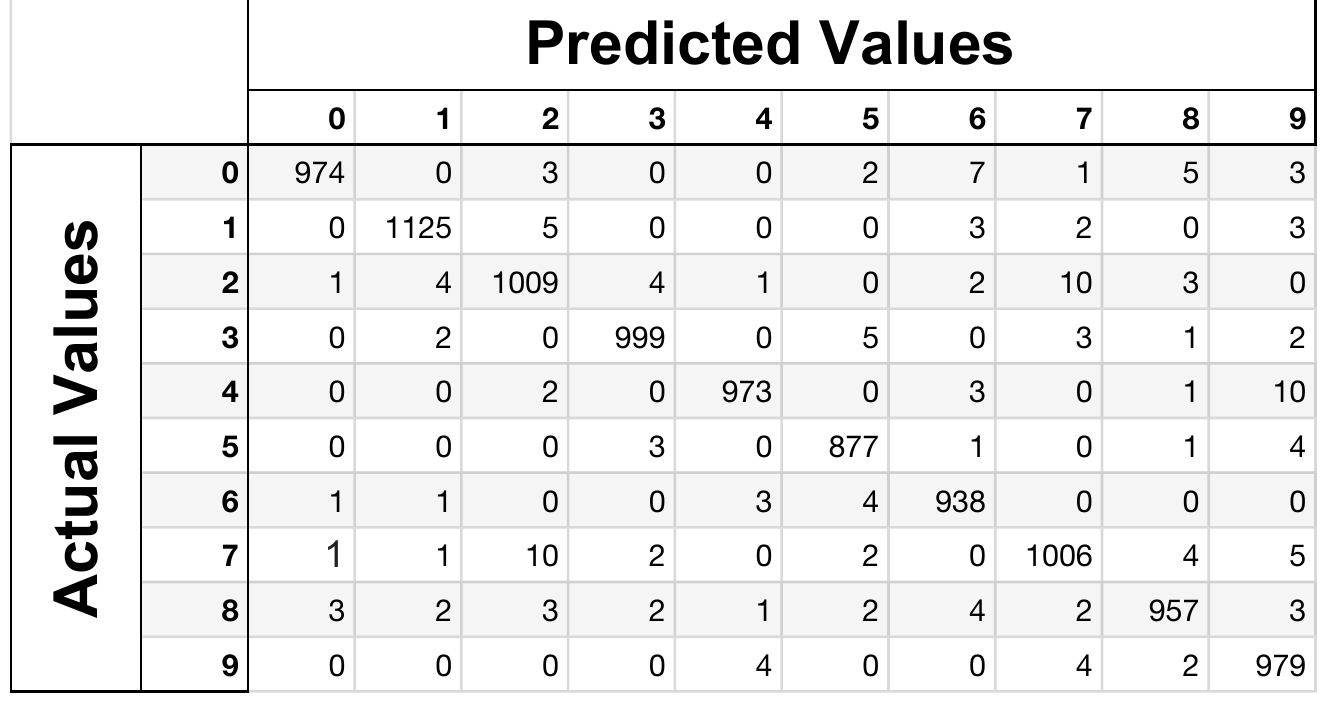}
	\includegraphics[scale=0.5]{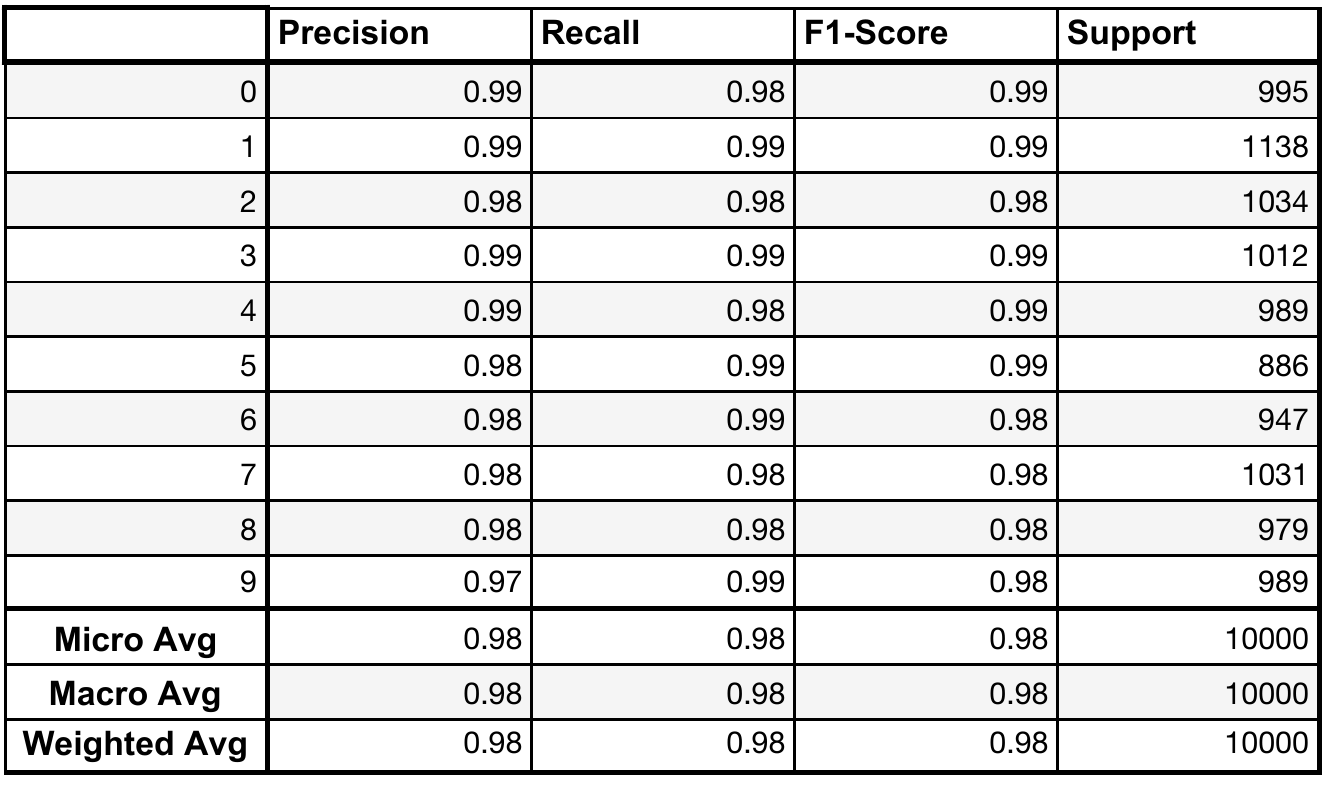}
	\caption{Confusion matrix and classification report of the neural network model without denoising autoencoder}
	\label{fig:NN_DAE_WOAE}
\end{figure}

\begin{figure}[htbp!]
	\centering
	\includegraphics[scale=0.56]{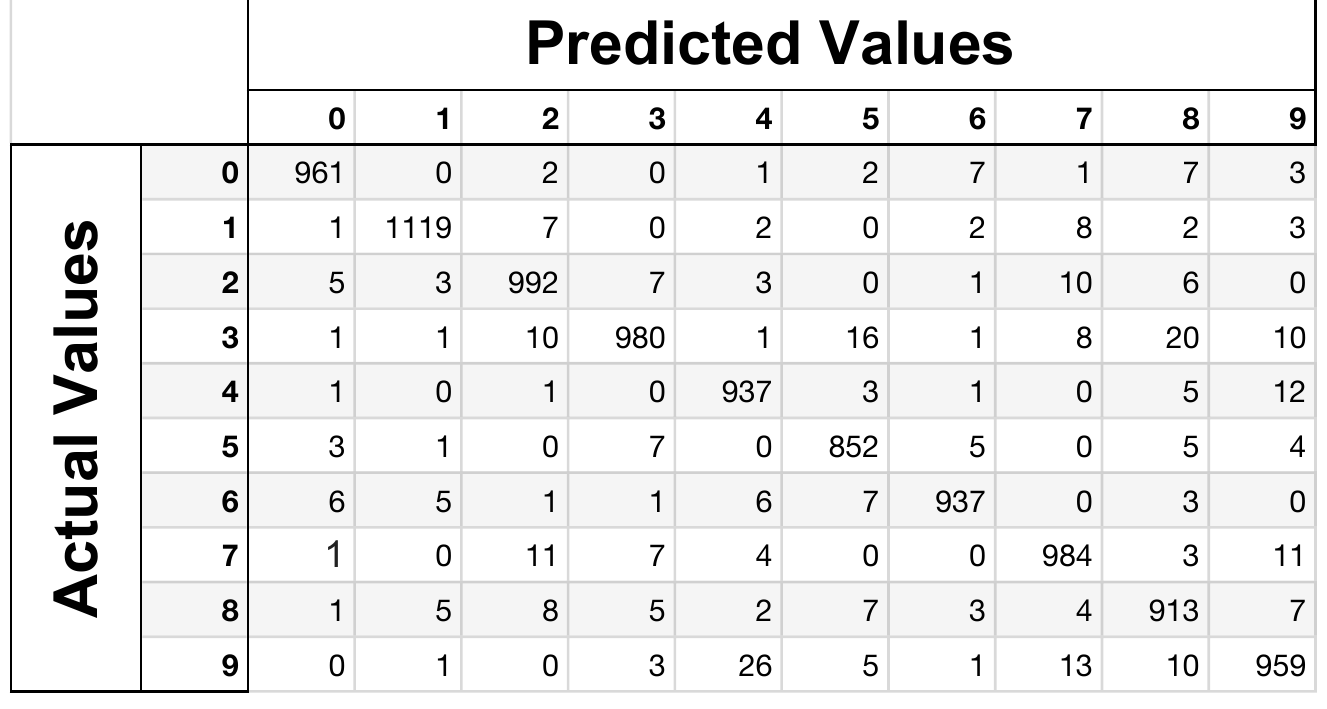}
	\includegraphics[scale=0.5]{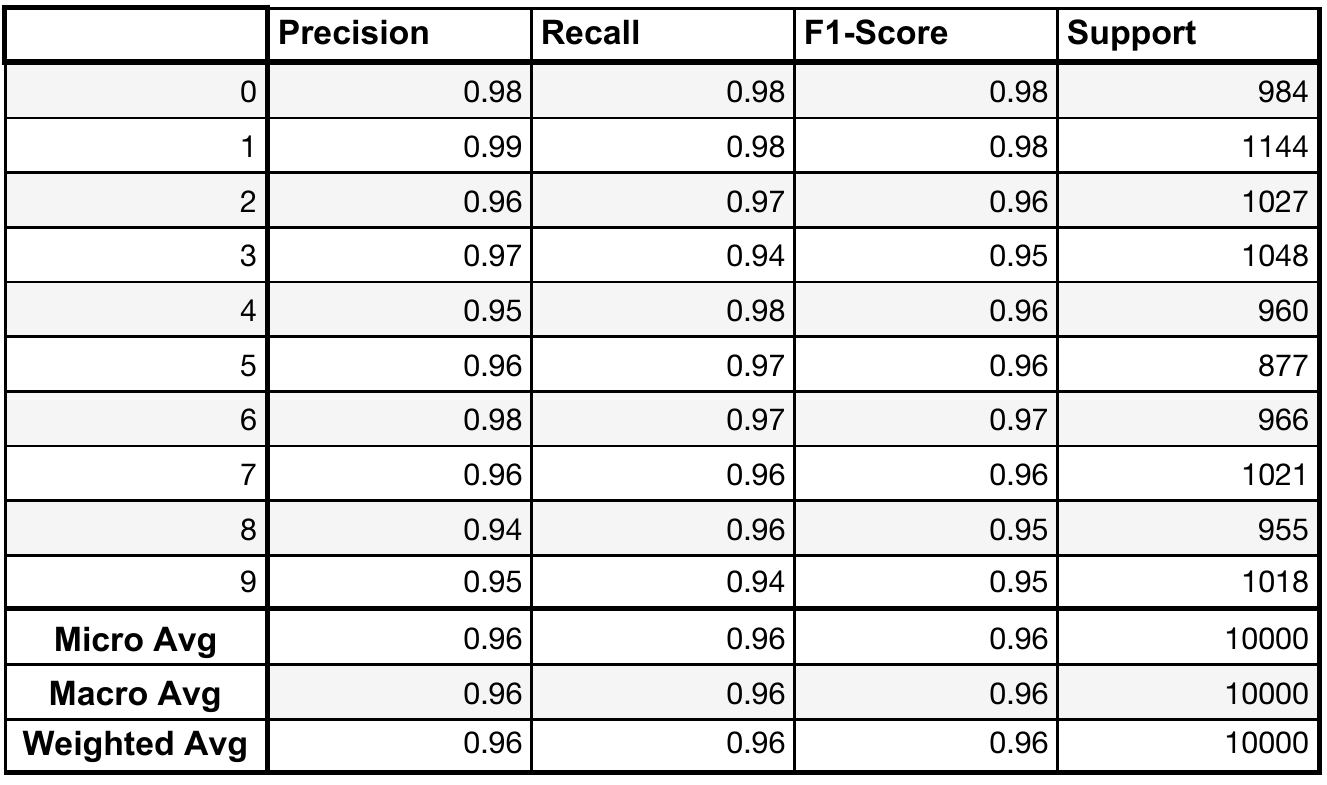}
	\caption{Confusion matrix and classification report of the neural network model with denoising autoencoder}
	\label{fig:NN_DAE_AE}
\end{figure}

\begin{figure}[htbp!]
	\centering
	\includegraphics[scale=0.56]{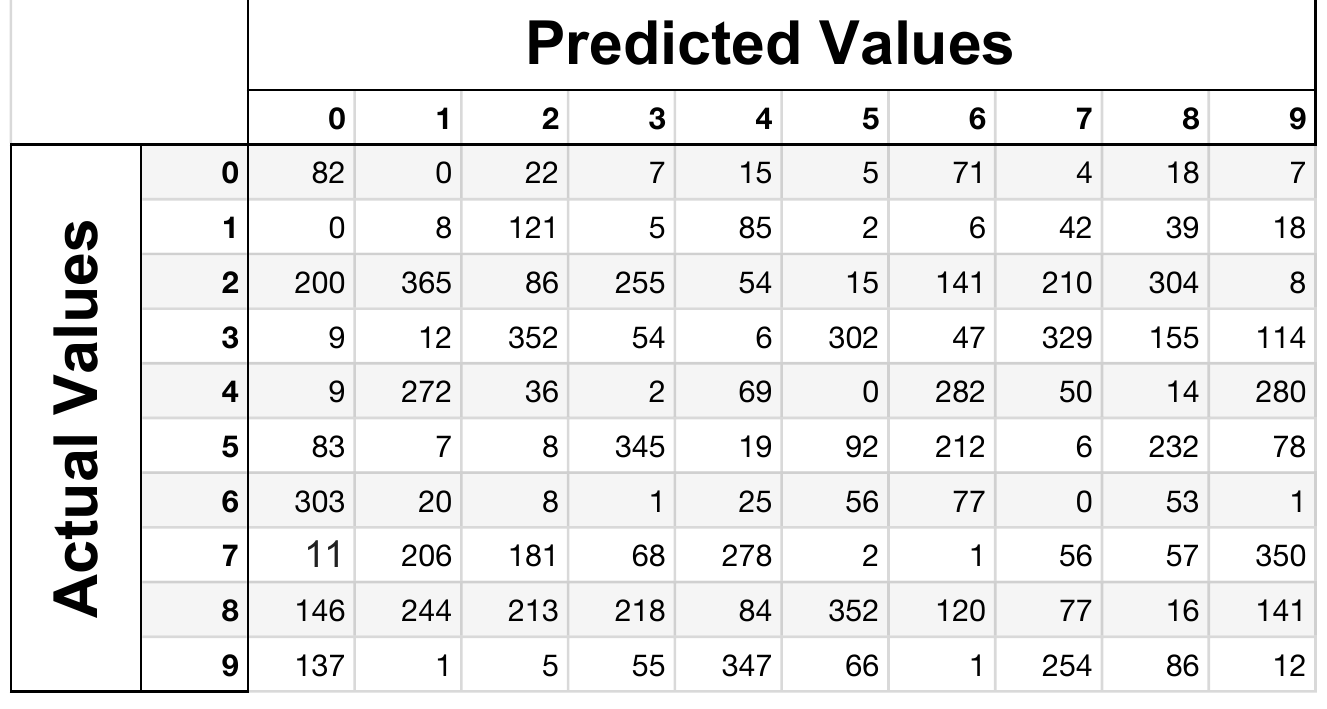}
	\includegraphics[scale=0.5]{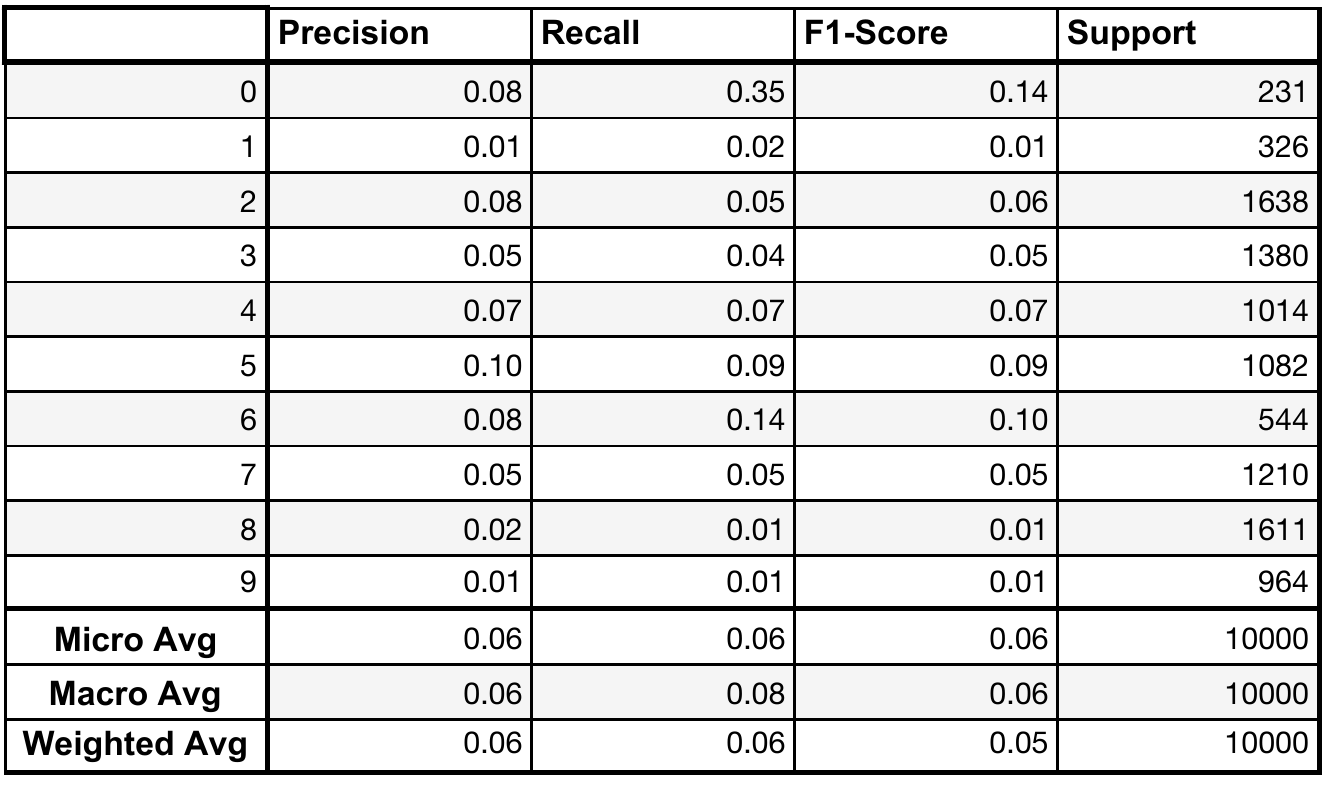}
	\caption{Confusion matrix and classification report of the neural network model without denoising autoencoder after FGSM attack}
	\label{fig:NN_DAE_WOAE_FGSM}
\end{figure}

\begin{figure}[htbp!]
	\centering
	\includegraphics[scale=0.56]{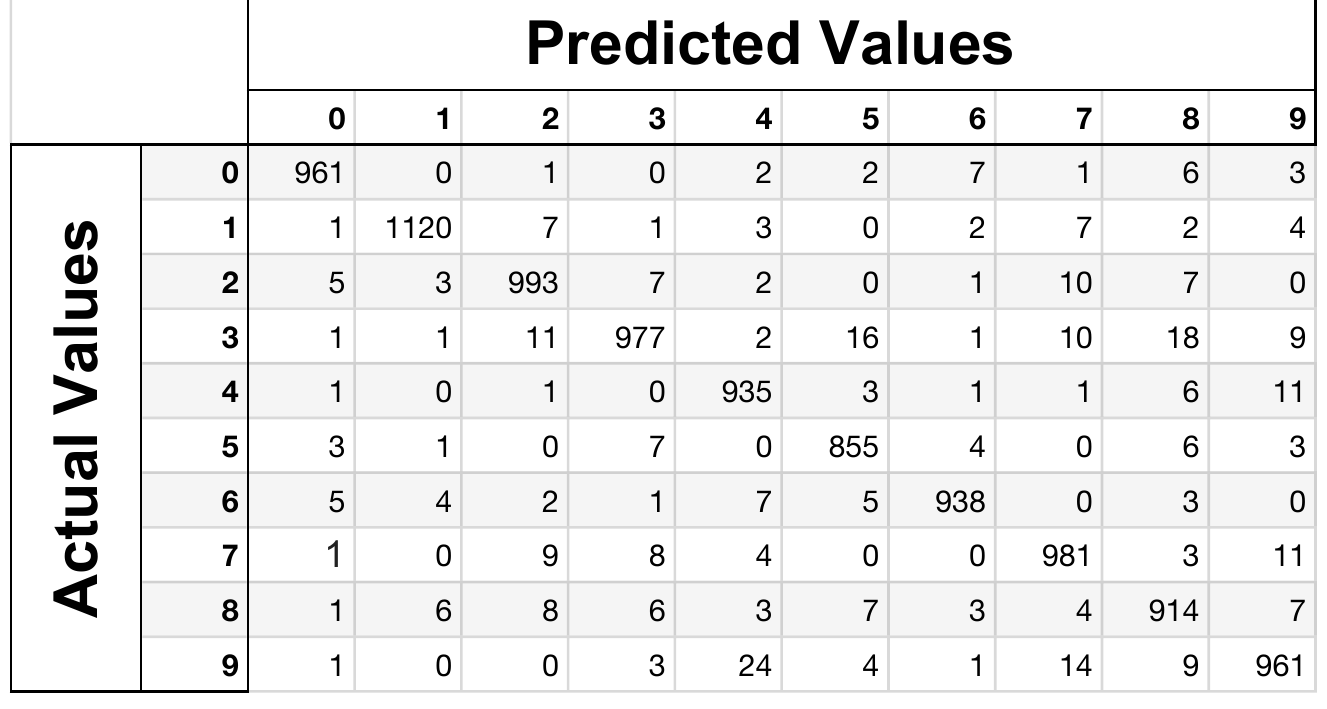}
	\includegraphics[scale=0.5]{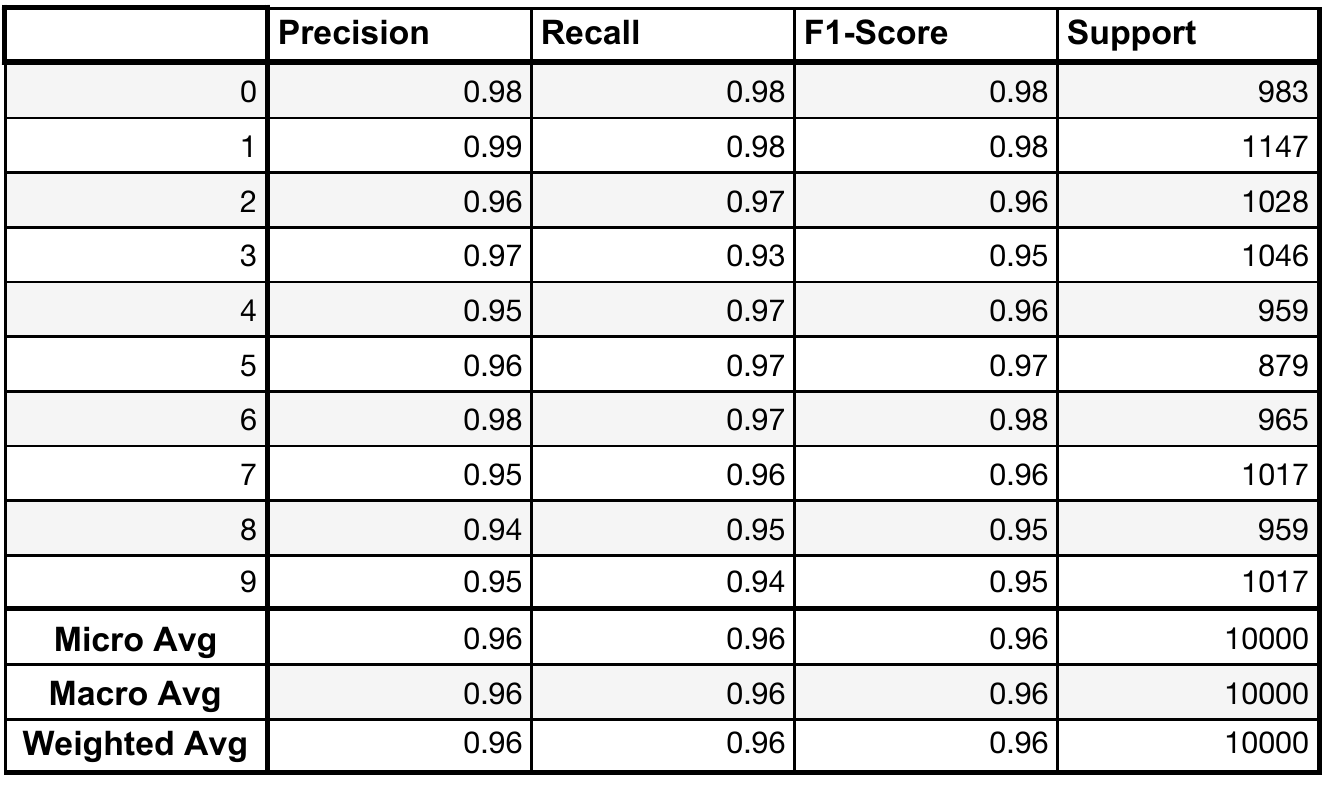}
	\caption{Confusion matrix and classification report of the neural network model with denoising autoencoder after FGSM attack}
	\label{fig:NN_DAE_AE_FGSM}
\end{figure}

\begin{figure}[htbp!]
	\centering
	\includegraphics[scale=0.56]{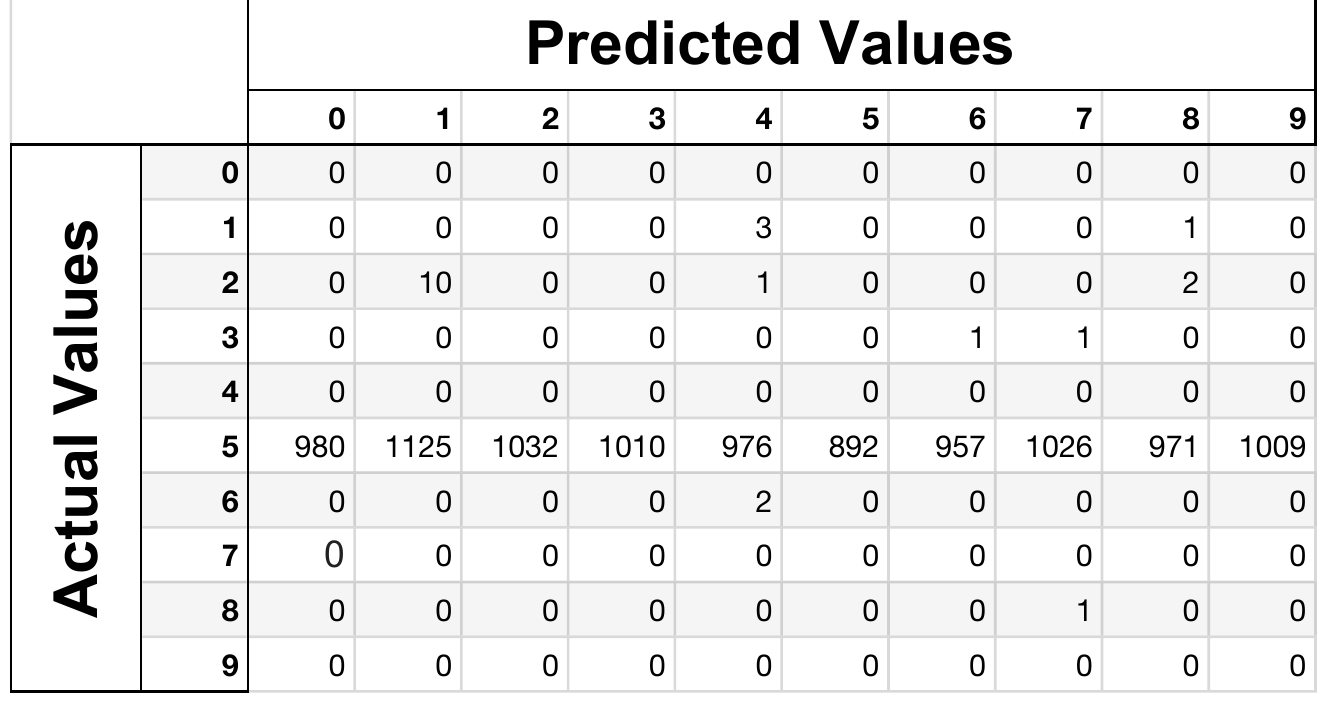}
	\includegraphics[scale=0.5]{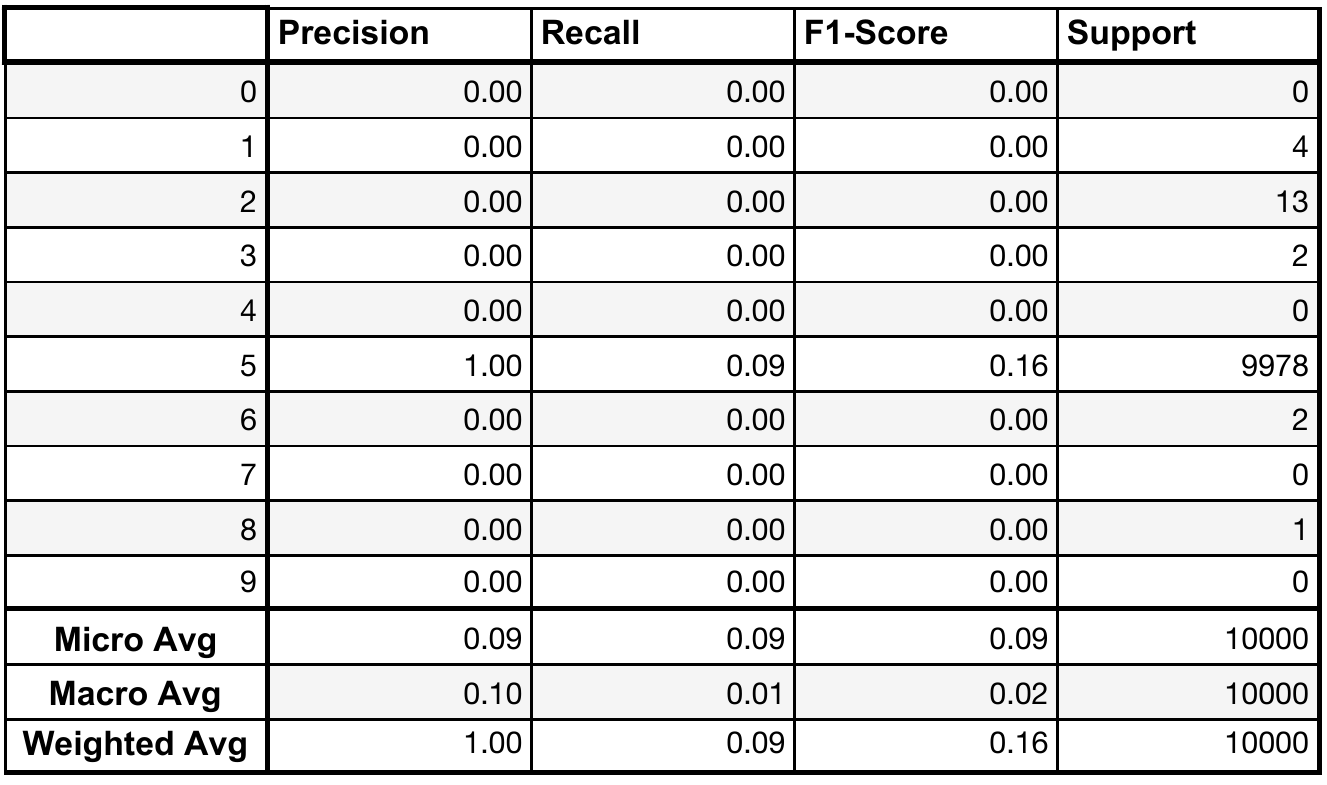}
	\caption{Confusion matrix and classification report of the neural network model without denoising autoencoder after T-FGSM attack}
	\label{fig:NN_DAE_WOAE_TFGSM}
\end{figure}

\begin{figure}[htbp!]
	\centering
	\includegraphics[scale=0.56]{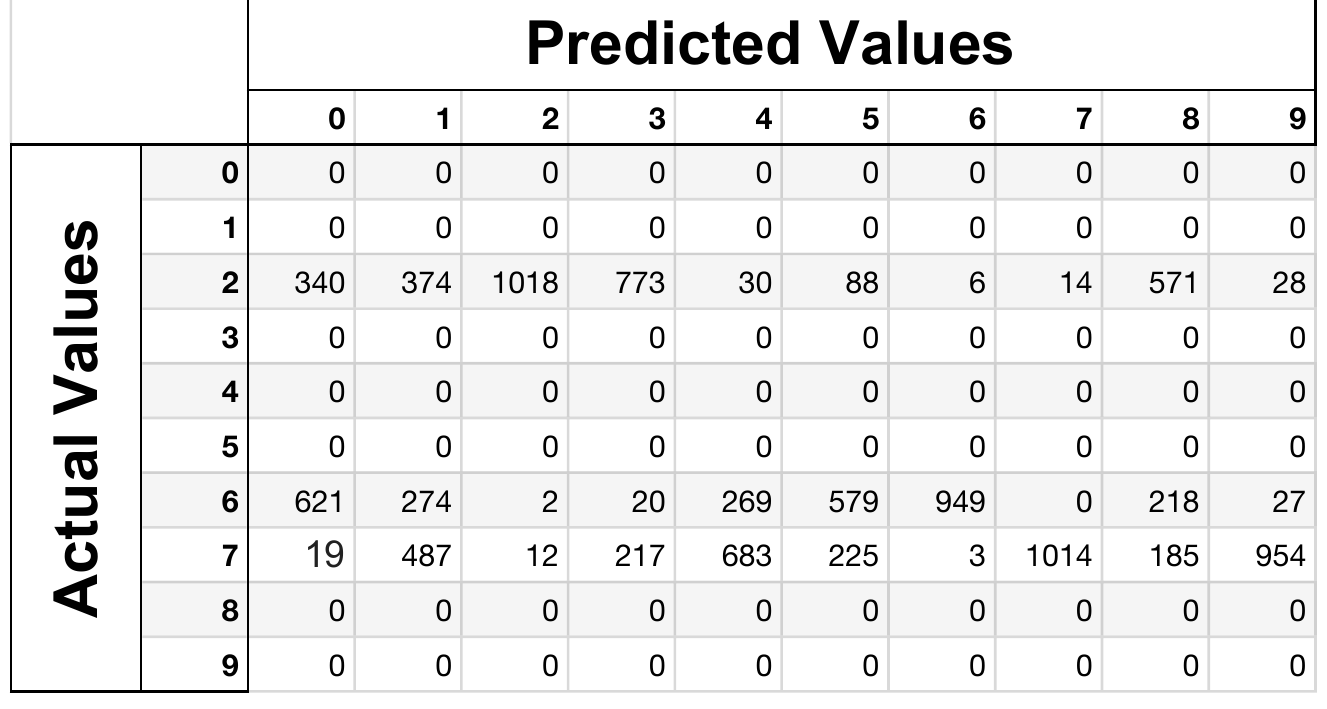}
	\includegraphics[scale=0.5]{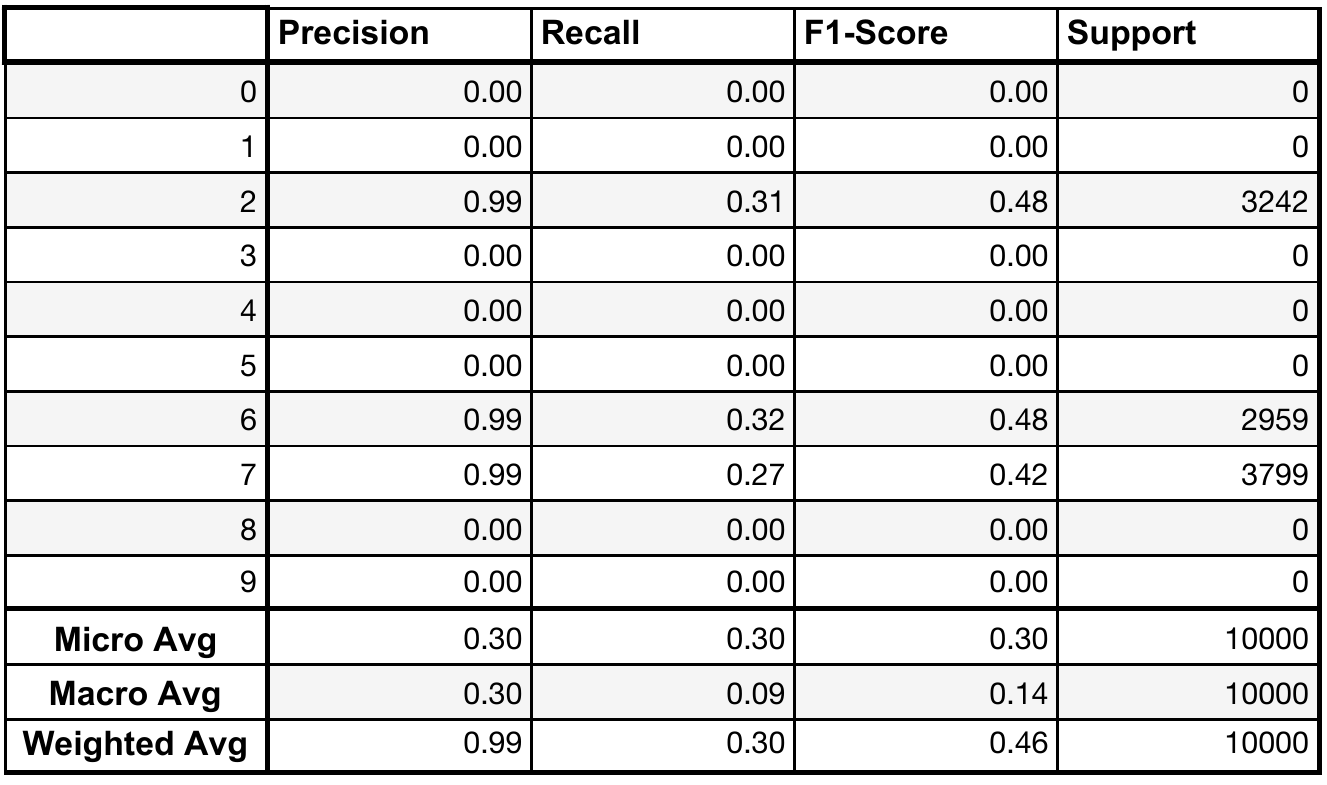}
	\caption{Confusion matrix and classification report of the neural network model with denoising autoencoder after T-FGSM attack}
	\label{fig:NN_DAE_AE_TFGSM}
\end{figure}

\begin{figure}[htbp!]
	\centering
	\includegraphics[scale=0.56]{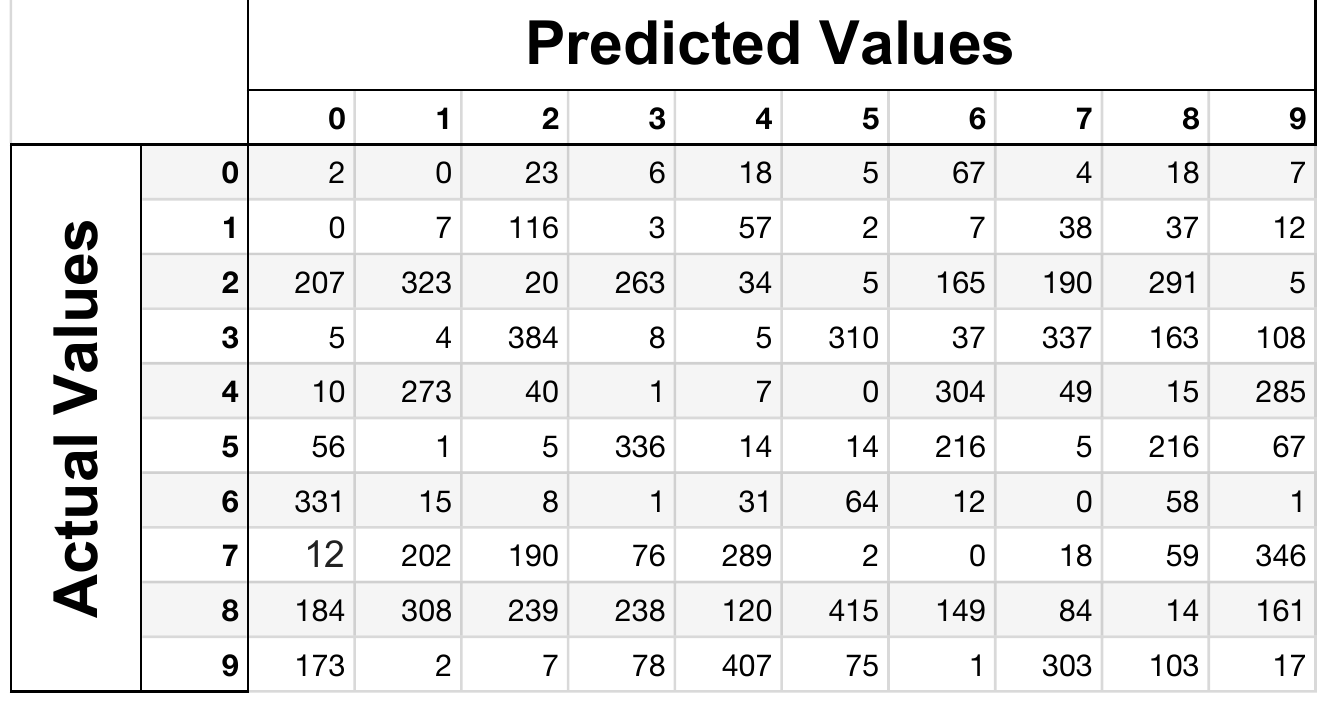}
	\includegraphics[scale=0.5]{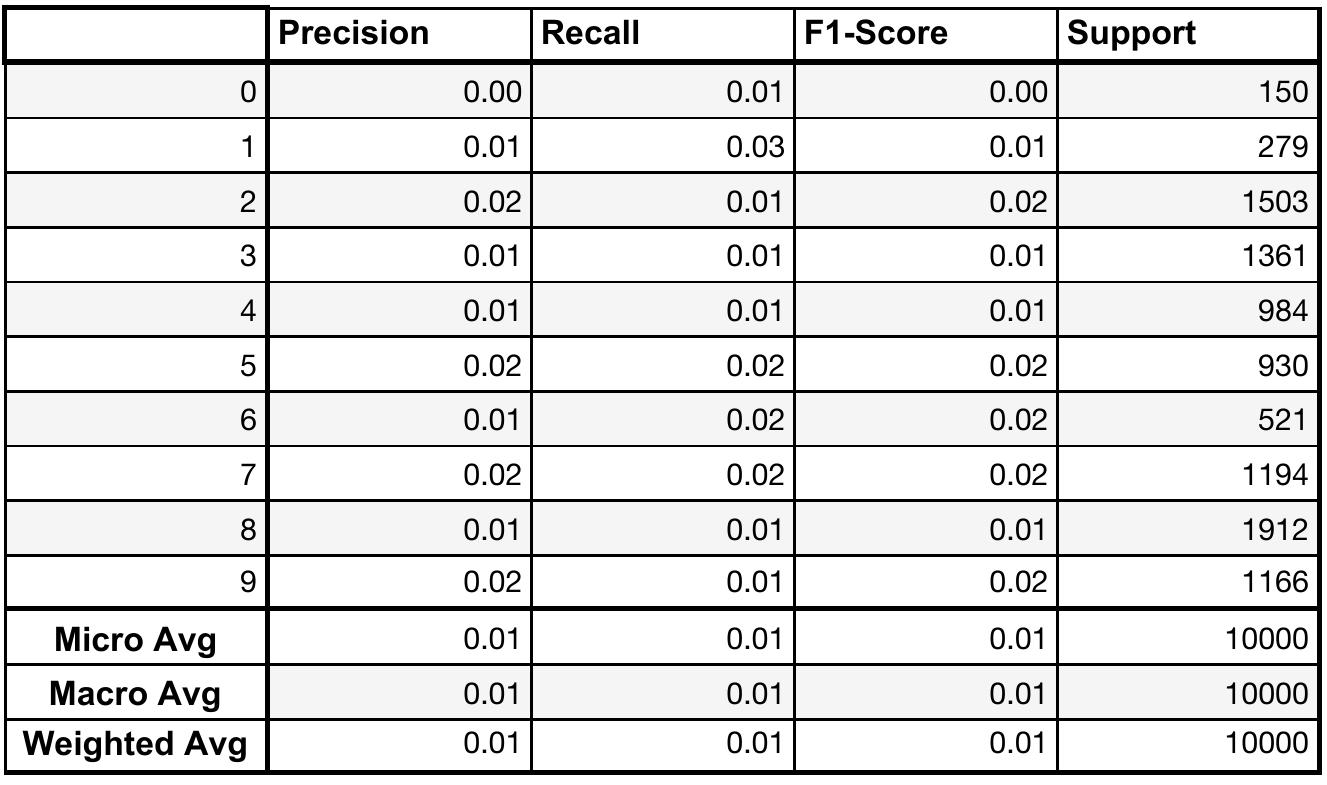}
	\caption{Confusion matrix and classification report of the neural network model without denoising autoencoder after basic iterative method attack}
	\label{fig:NN_DAE_WOAE_BIM}
\end{figure}

\begin{figure}[htbp!]
	\centering
	\includegraphics[scale=0.56]{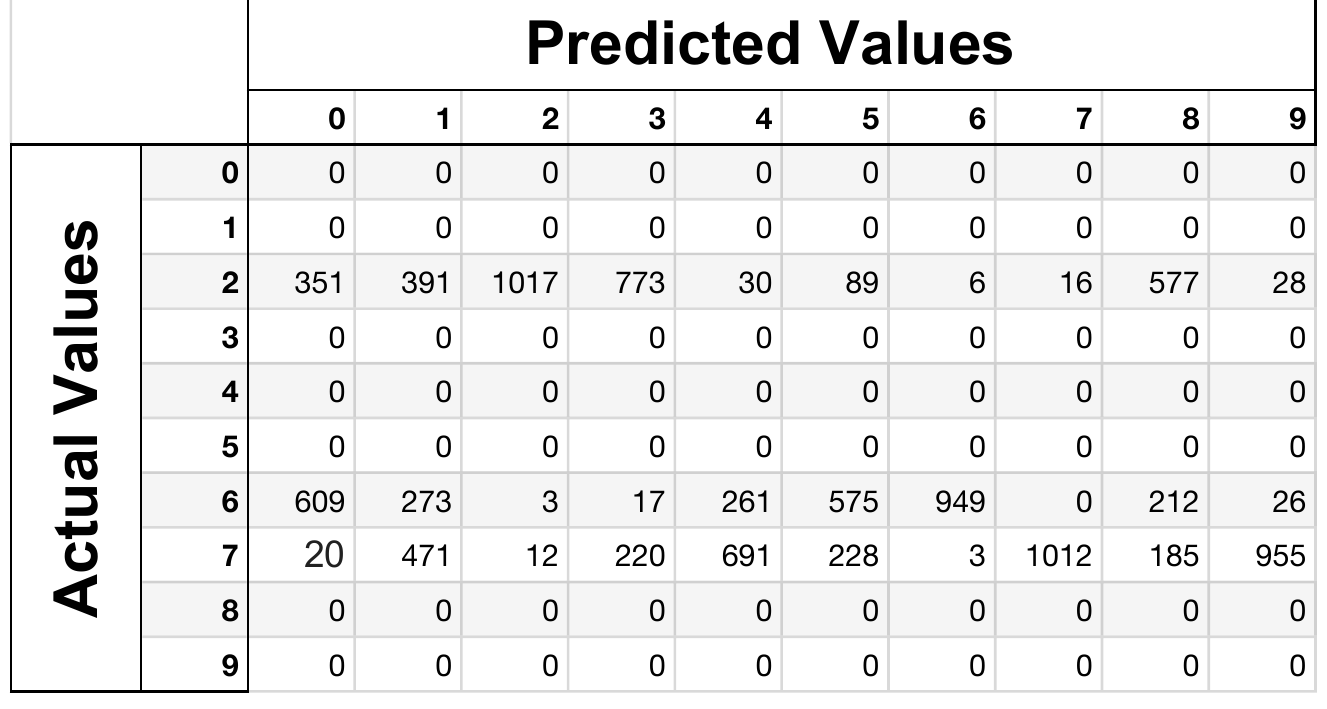}
	\includegraphics[scale=0.5]{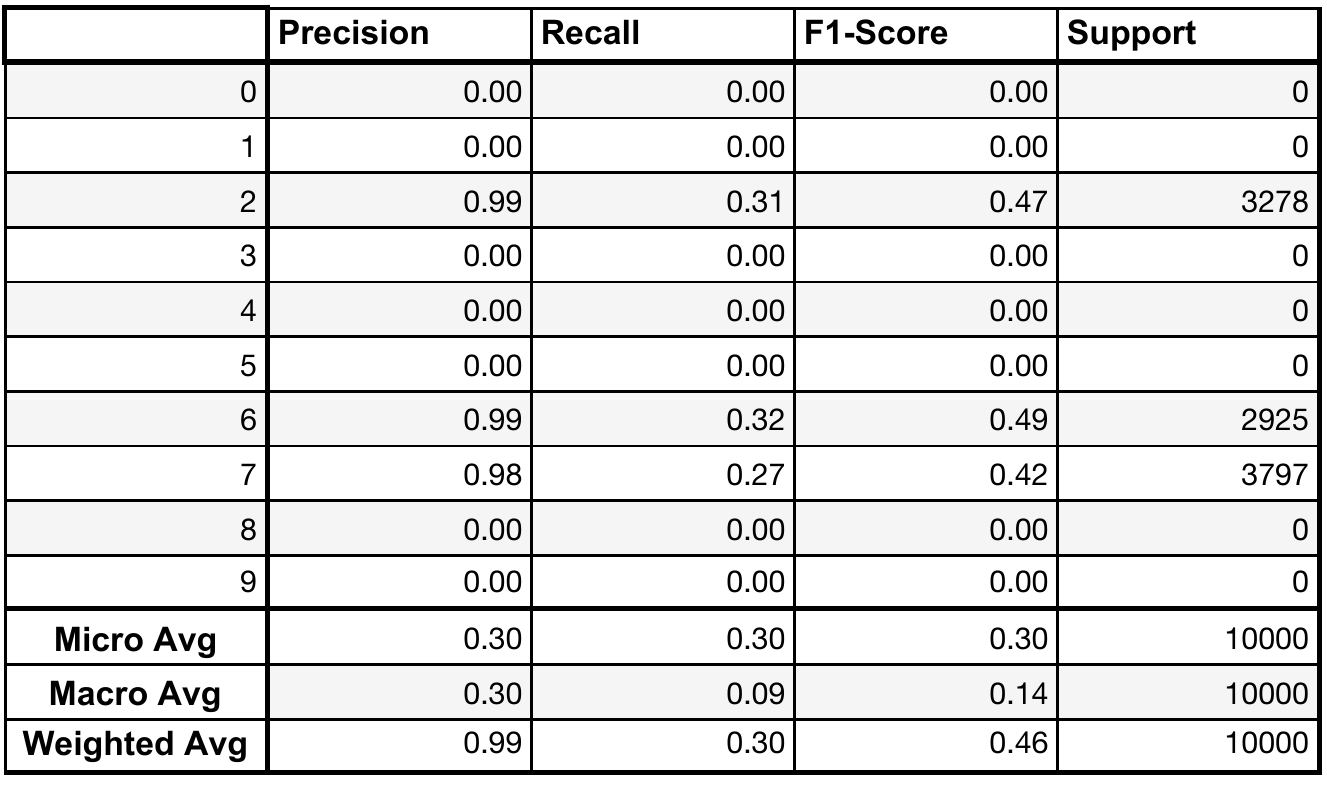}
	\caption{Confusion matrix and classification report of the neural network model with denoising autoencoder after basic iterative method attack}
	\label{fig:NN_DAE_AE_BIM}
\end{figure}


\subsection{Variational Autoencoder}

In this study, we examine variational autoencoders as the final type of autoencoder type. The variational autoencoders have an encoder and a decoder, although their mathematical formulation differs significantly. They are associated with Generative Adversarial Networks due to their architectural similarity. In summary, variational autoencoders are also generative models. Differently, from sparse autoencoders, denoising autoencoders, and vanilla autoencoders, all of which aim discriminative modeling, generative modeling tries to simulate how the data can be generated and to understand the underlying causal relations. It also considers these causal relations when generating new data.

Variational autoencoders use an estimator algorithm called Stochastic Gradient Variational Bayes for training. This algorithm assumes the data is generated by $p_\theta(x|h)$ which is a directed graphical model and $\theta$ being the parameters of decoder, in variational autoencoder's case, the parameters of the generative model. The encoder is learning an approximation of $q_\phi(h|x)$ to a posterior distribution which is showed by $p_\theta(x|h)$ and $\phi$ being the parameters of the encoder; in variational autoencoder's case, the parameters of recognition model. We will use Kullback-Leibler divergence again, showed as $D_{KL}$.

$L=(\phi,\theta,x) = D_{KL}(q_\phi(h|x)||p_\theta(h)) - \mathbb{E}_{q_\phi(h|x)}(log p_\theta(x|h))$.

Variational and likelihood distributions' shape is chosen by factorized Gaussians. The encoder outputs are $p(x)$ and $w^2(x)$. The decoder outputs are $\mu(h)$ and $\sigma^2(h)$. The likelihood term of variational objective is defined below.

$q_\phi(h|x)=N(p(x),w^2(x)I)$

$p_\theta(x|h)=N(\mu(h),\sigma^2(h)I)$

\subsubsection{Multi-Class Logistic Regression of Variational Autoencoder}
\begin{figure}[htbp!]
	\centering
	\includegraphics[width=0.5\linewidth]{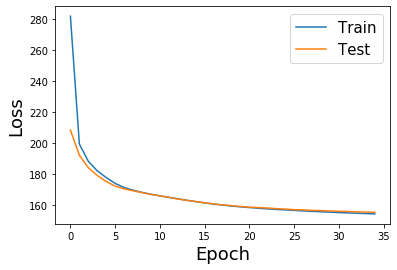}
	\caption{Optimized Relu Loss History for Variational Autoencoder}
	\label{fig:lossVAE}
\end{figure}

\begin{figure}[htbp!]
	\includegraphics[width=1.0\linewidth]{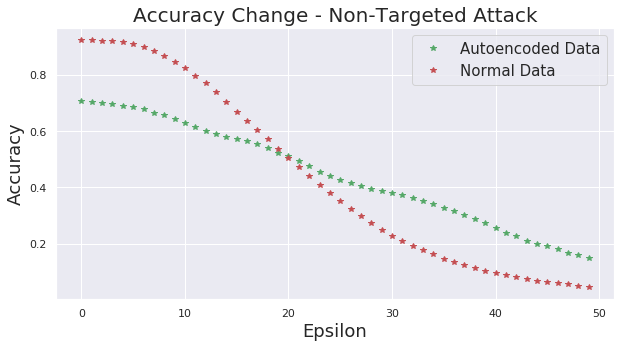}
	\caption{Comparison of accuracy with and without variational autoencoder for non-targeted attack}
	\label{fig:nontargetedAttackVAE}
\end{figure}

The findings from Figure \ref{fig:lossVAE} show that variational autoencoder indicates the best loss function result. However, Figure \ref{fig:nontargetedAttackVAE} presents that the accuracy is low, especially in low epsilon values where even autoencoded data gives worse accuracy than the normal learning process. 

Perturbation applied by variational autoencoder is not as sharp in sparse autoencoder and denoising autoencoder. It seems similar to vanilla autoencoder’s perturbation.

\begin{figure}[htbp!]
	\centering
	\includegraphics[width=0.75\linewidth]{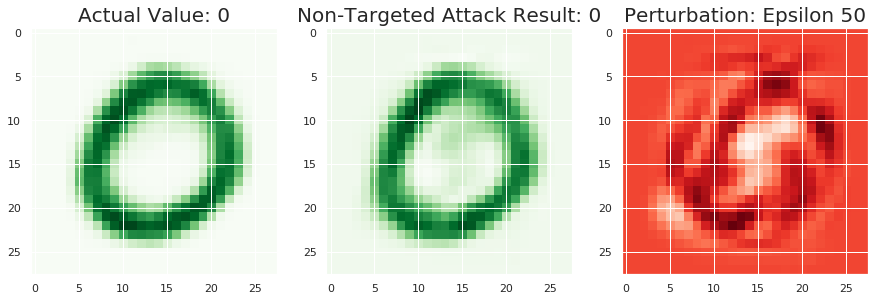}
	\caption{Value change and perturbation of a non-targeted attack on model without variational autoencoder}
	\label{fig:perturbationUnsuccessVAE}
\end{figure}

\begin{figure}[htbp!]
	\centering
	\includegraphics[width=0.75\linewidth]{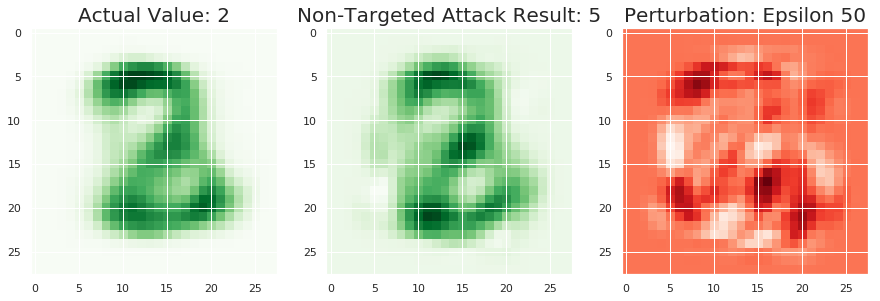}
	\caption{Value change and perturbation of a non-targeted attack on model with variational autoencoder}
	\label{fig:perturbationSuccessVAE}
\end{figure}

The variational autoencoder has the worst results. Besides, it presents bad results at the low values of epsilon, making autoencoded data less accurate and only a slight improvement compared to the normal data in high values of epsilon.

\begin{figure}[htbp!]
	\includegraphics[width=1.0\linewidth]{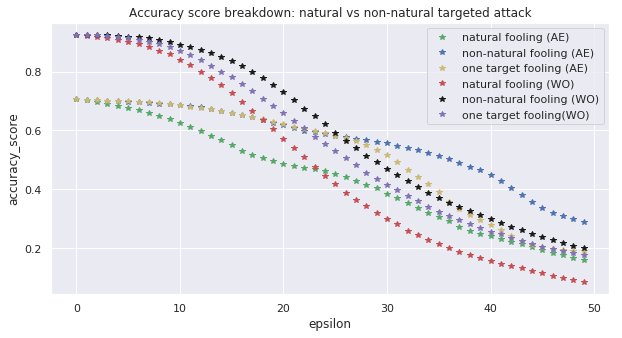}
	\caption{Comparison of accuracy with and without variational autoencoder for targeted attacks. \textit{AE stands for the models with variational autoencoder, WO stands for models without autoencoder}}
	\label{fig:targetedAttacksVAE}
\end{figure}

\begin{figure}[htbp!]
	\centering
	\includegraphics[width=0.5\linewidth]{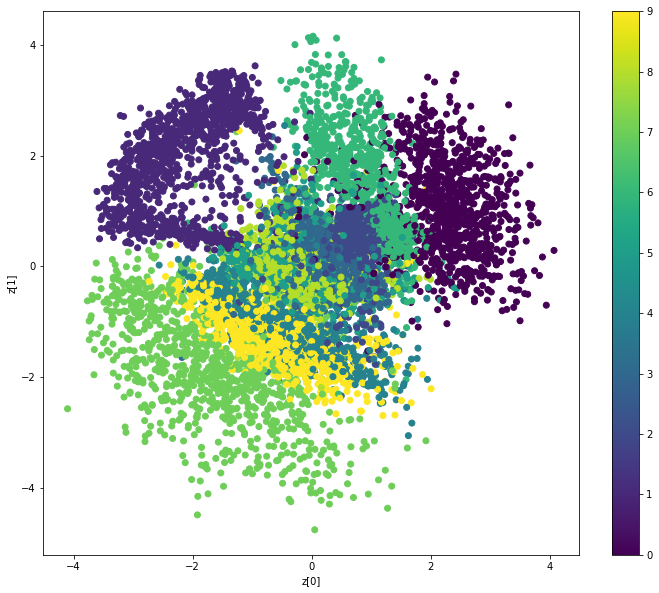}
	\caption{Because of Mnist dataset, our latent space is two-dimensional. One is to look at the neighborhoods of different classes on the latent 2D plane. Each of these colored clusters are a type of digit. Close clusters are digits that are structurally similar, they are digits that share information in the latent space.}
	\label{fig:clustersVAEMCLR}
\end{figure}

\begin{figure}[htbp!]
	\centering
	\includegraphics[width=0.5\linewidth]{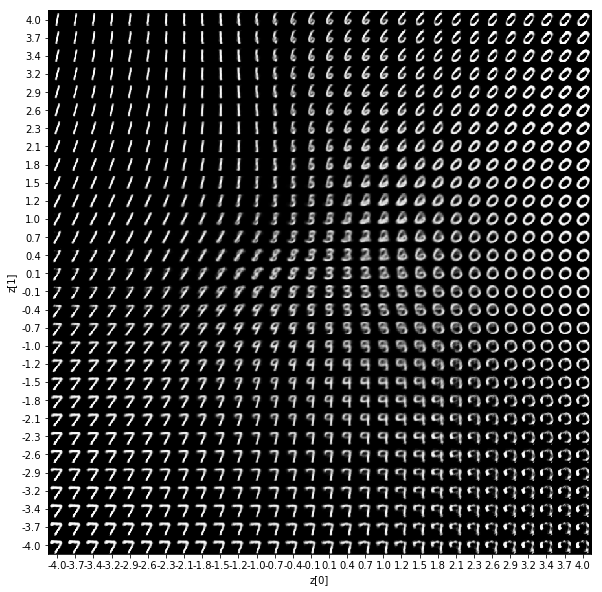}
	\caption{Due to VAE is a generative model, we can also generate new Mnist digits using latent plane, sampling latent points at regular intervals, and generating the corresponding digit for each of these points.}
	\label{fig:GANVAEMCLR}
\end{figure}

\subsubsection{Neural Network of Variational Autoencoder}

Variational autoencoder with neural networks also illustrates the worst results compared to other autoencoder types, where the accuracy for autoencoded data against an attack has around between 0.96 and 0.99 accuracies, variational autoencoder has around between 0.65 and 0.70 accuracies.

\begin{figure}[htbp!]
	\centering
	\includegraphics[scale=0.56]{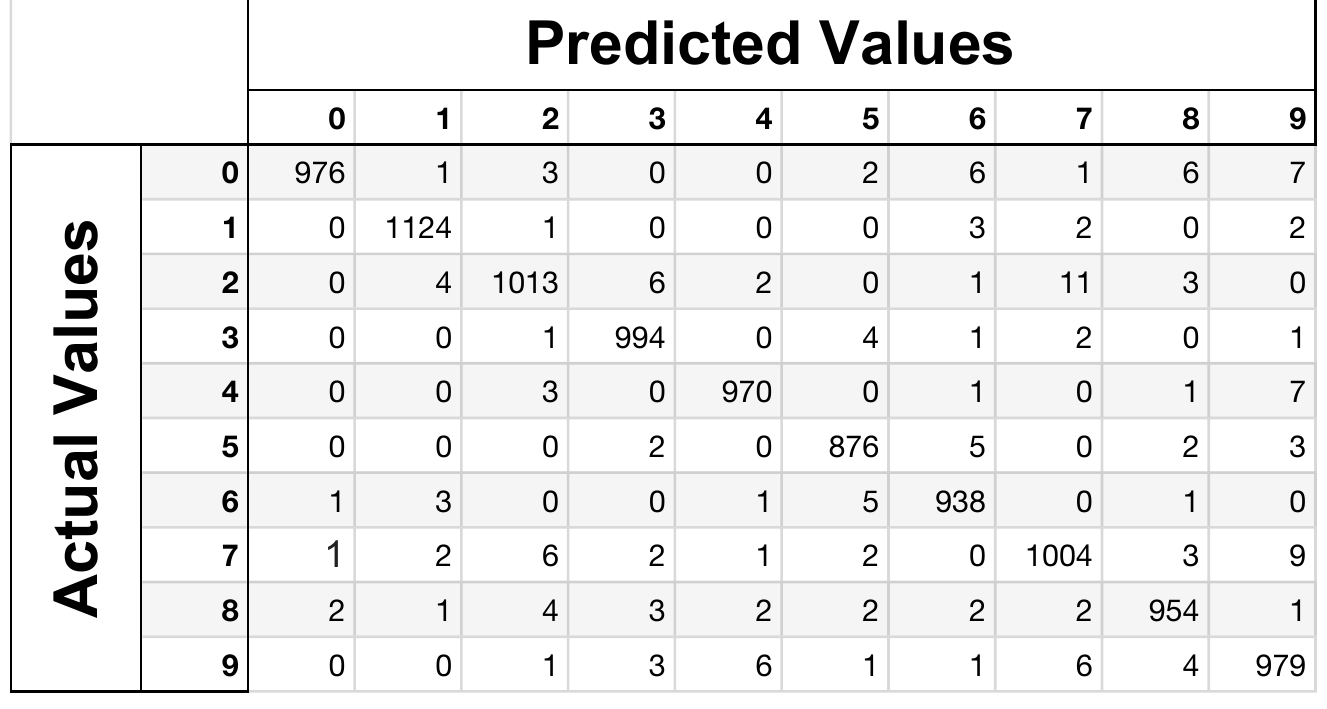}
	\includegraphics[scale=0.5]{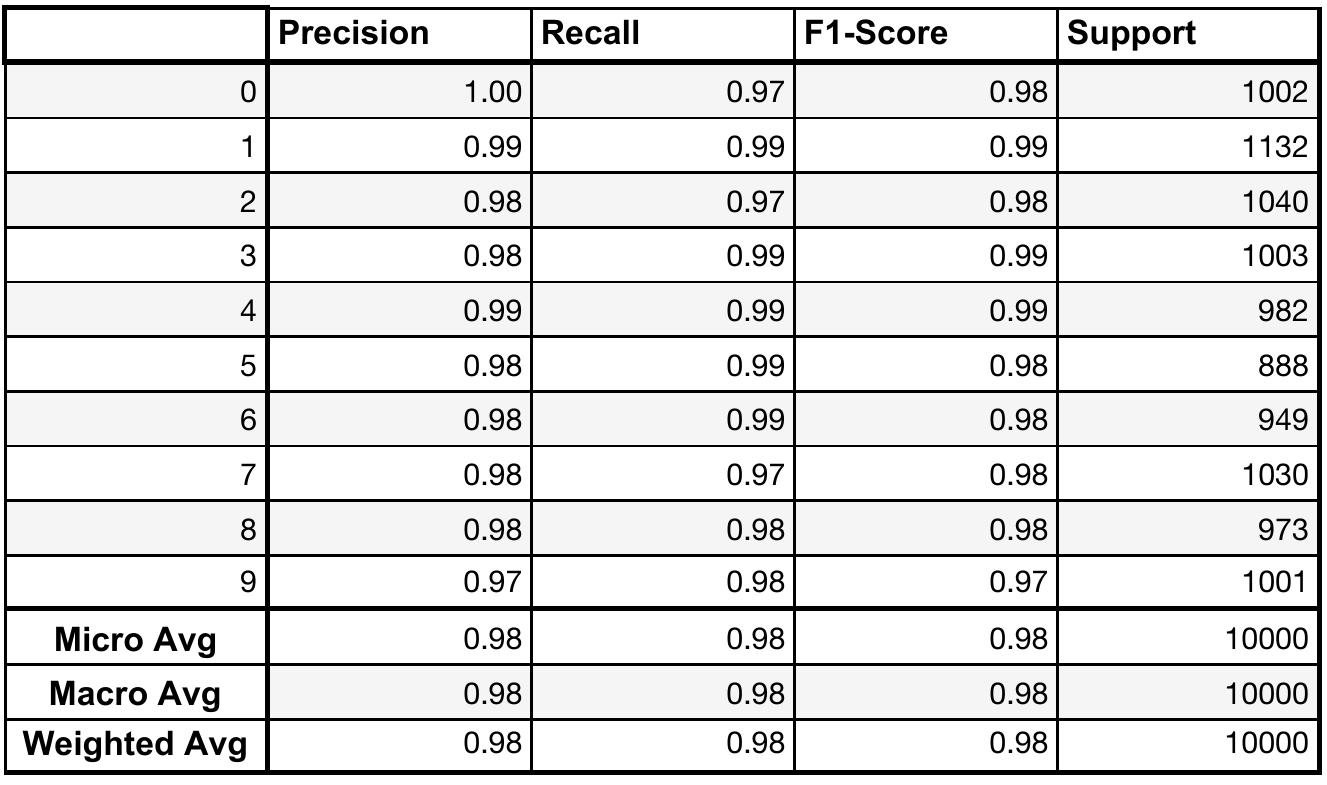}
	\caption{Confusion matrix and classification report of the neural network model without variational autoencoder}
	\label{fig:NN_VAE_WOAE}
\end{figure}

\begin{figure}[htbp!]
	\centering
	\includegraphics[scale=0.56]{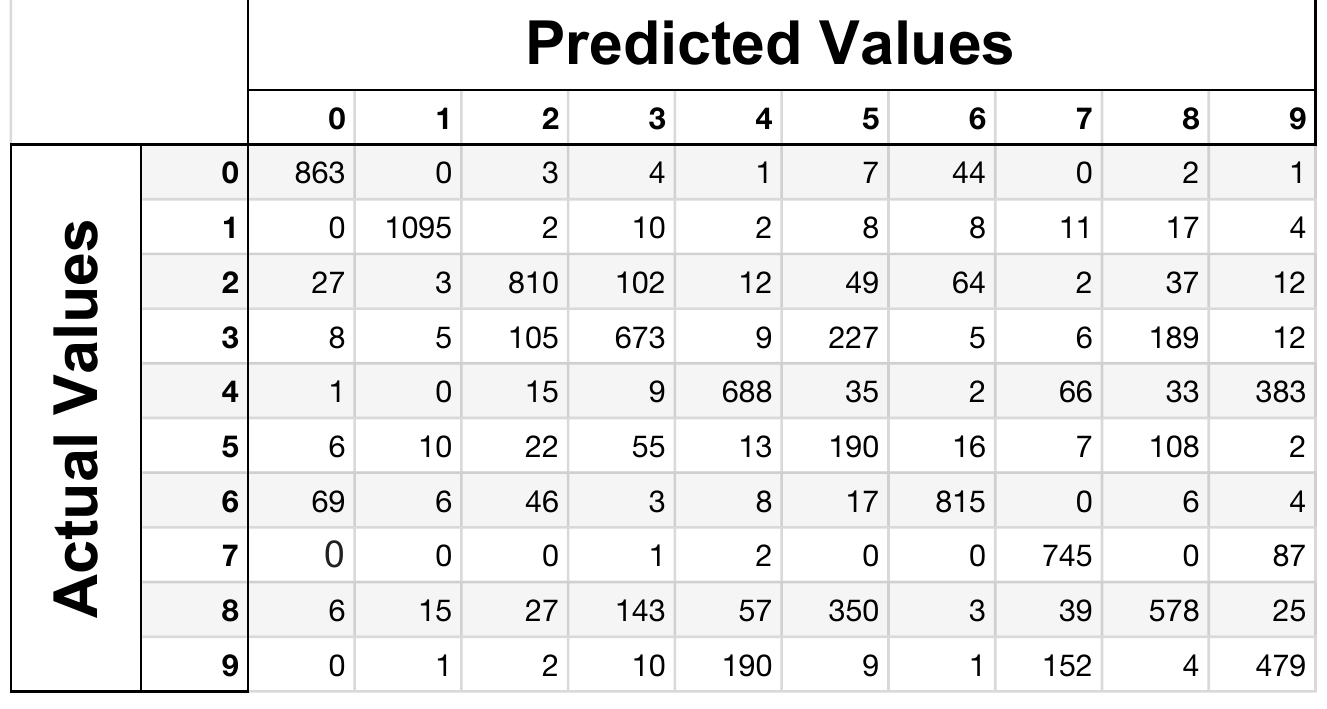}
	\includegraphics[scale=0.5]{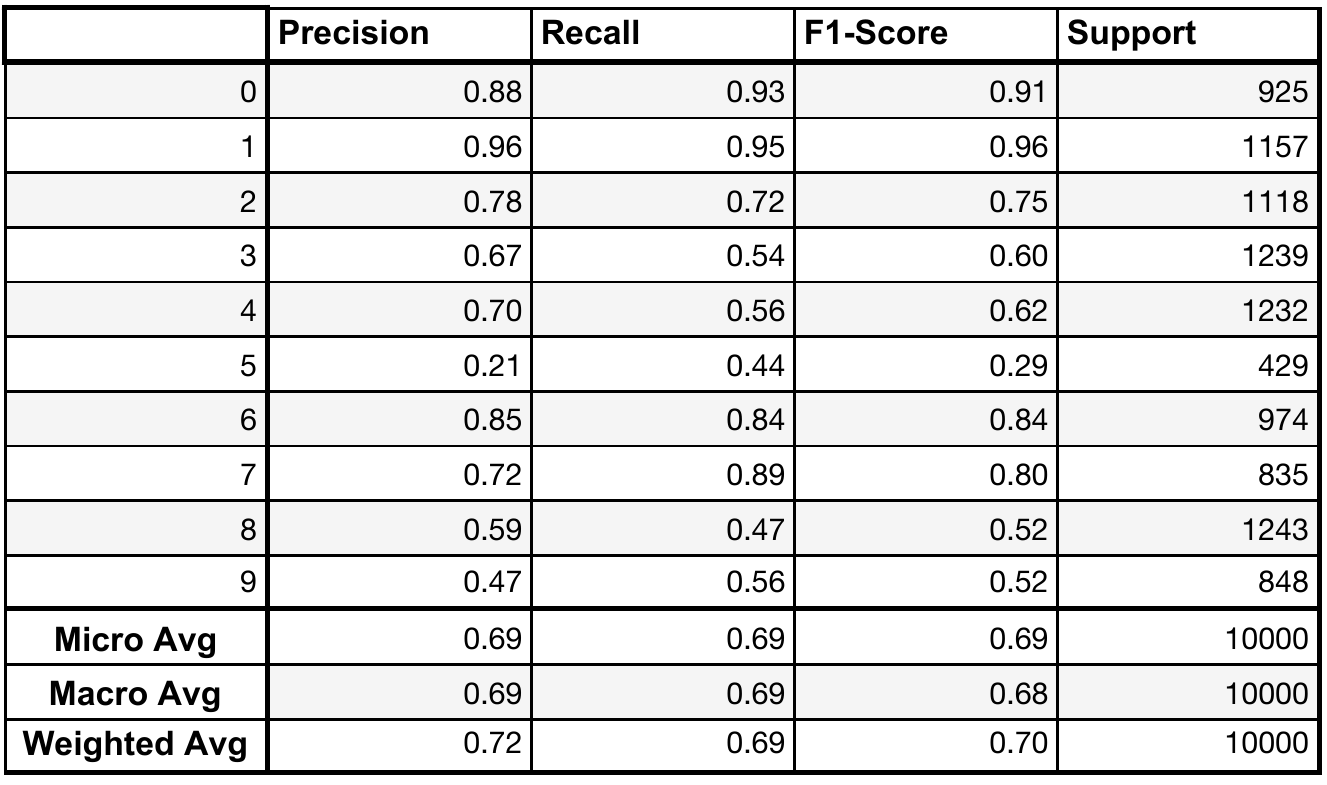}
	\caption{Confusion matrix and classification report of the neural network model with variational autoencoder}
	\label{fig:NN_VAE_AE}
\end{figure}

\begin{figure}[htbp!]
	\centering
	\includegraphics[scale=0.56]{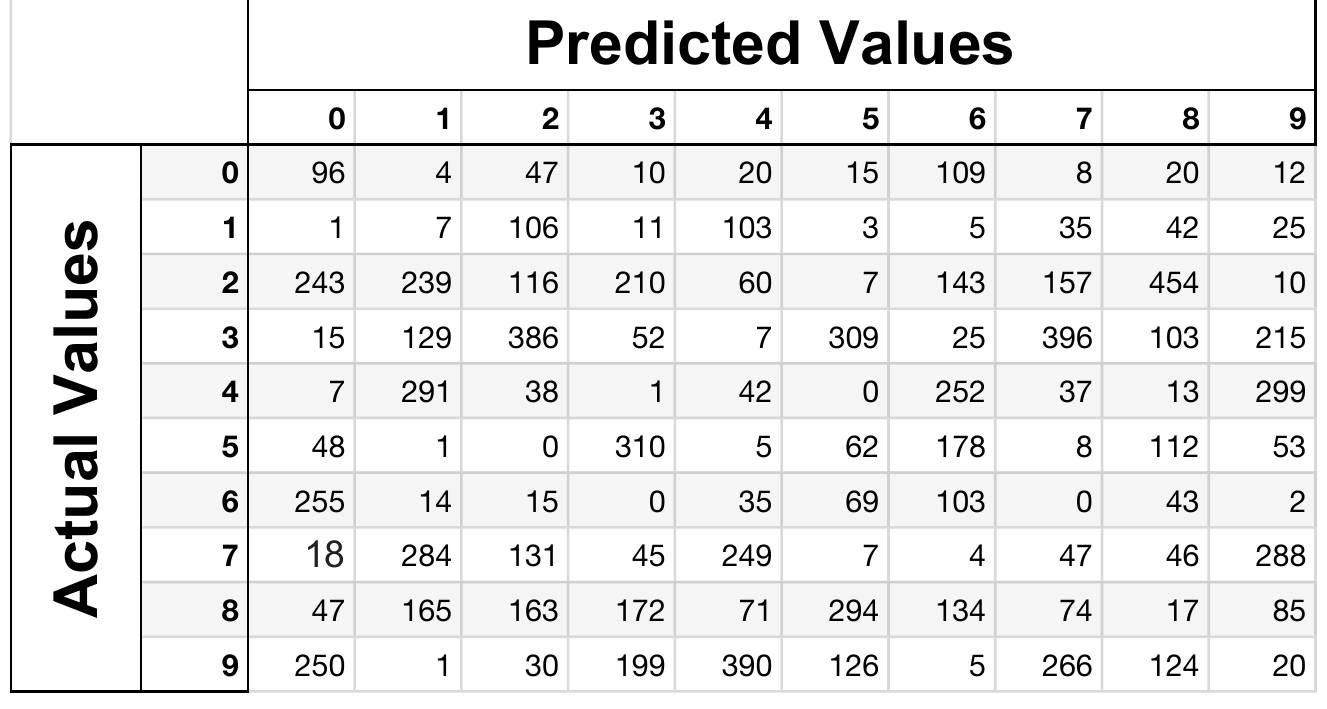}
	\includegraphics[scale=0.5]{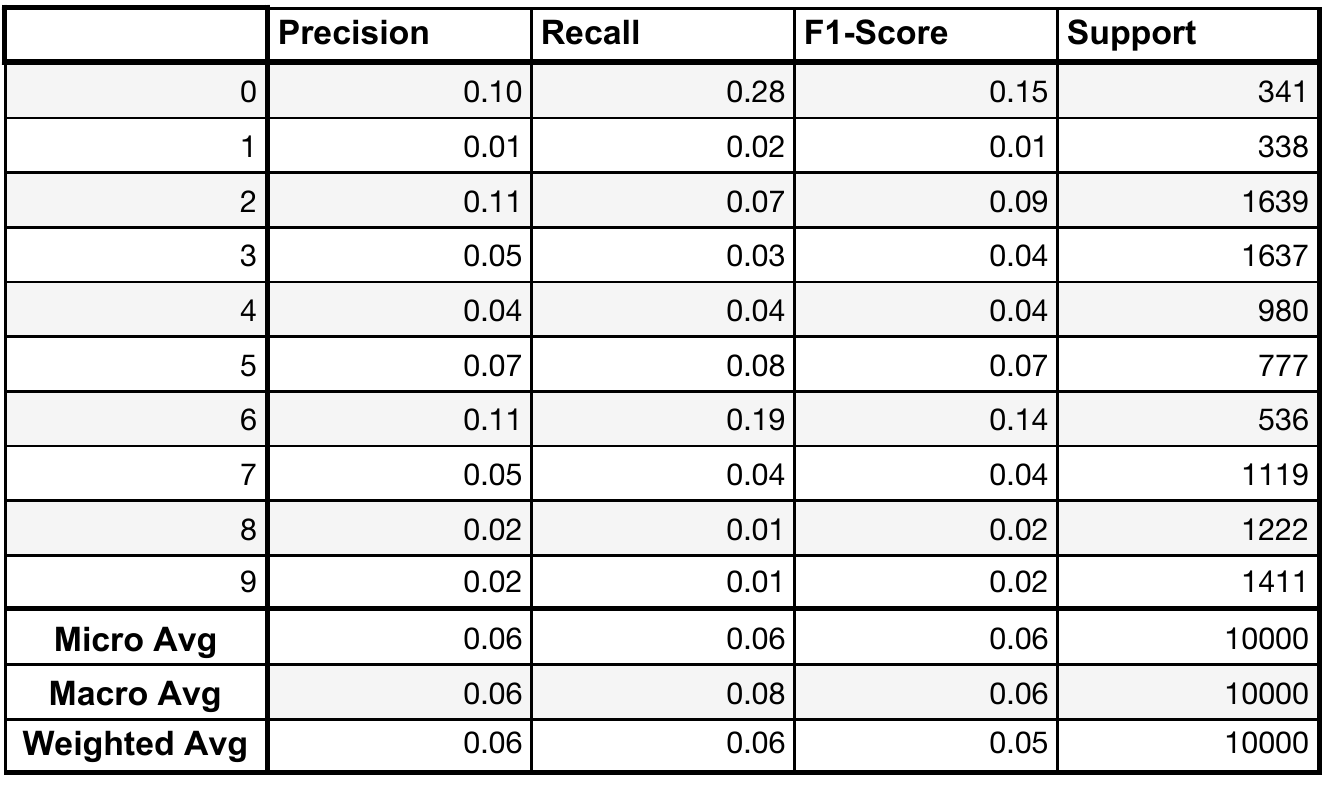}
	\caption{Confusion matrix and classification report of the neural network model without variational autoencoder after FGSM attack}
	\label{fig:NN_VAE_WOAE_FGSM}
\end{figure}

\begin{figure}[htbp!]
	\centering
	\includegraphics[scale=0.56]{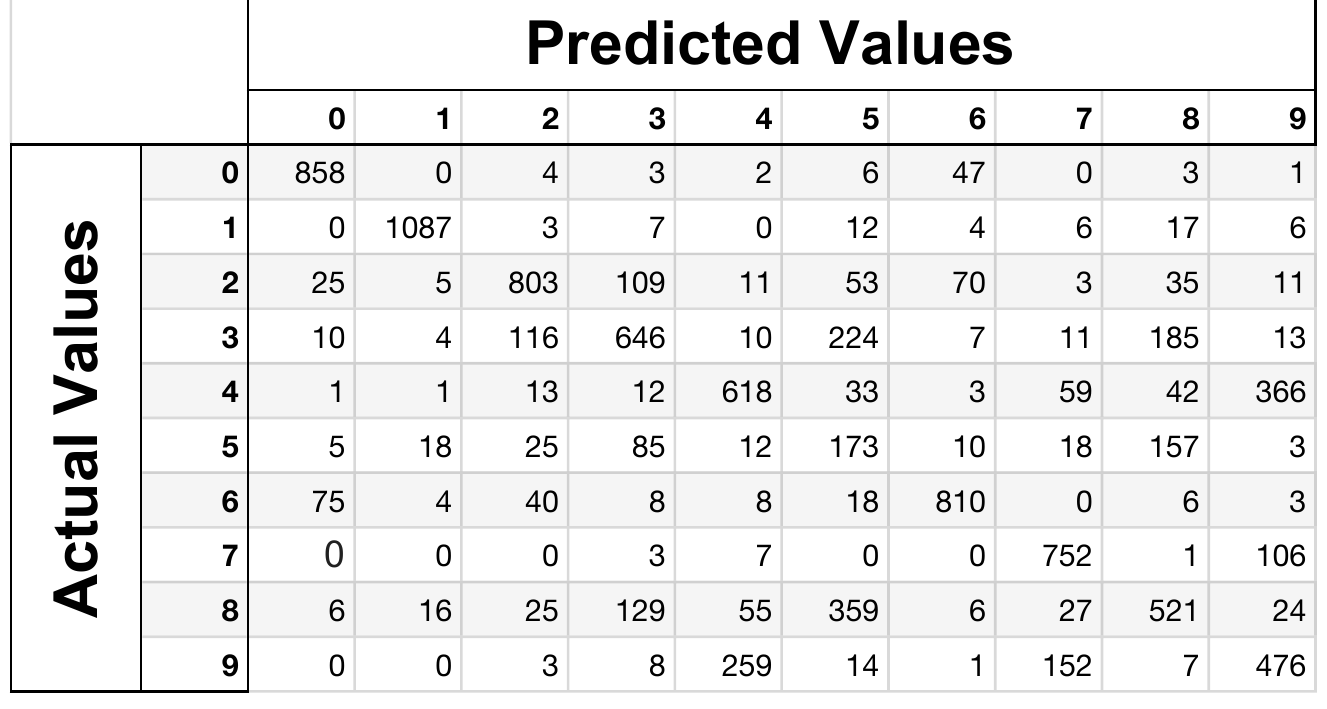}
	\includegraphics[scale=0.5]{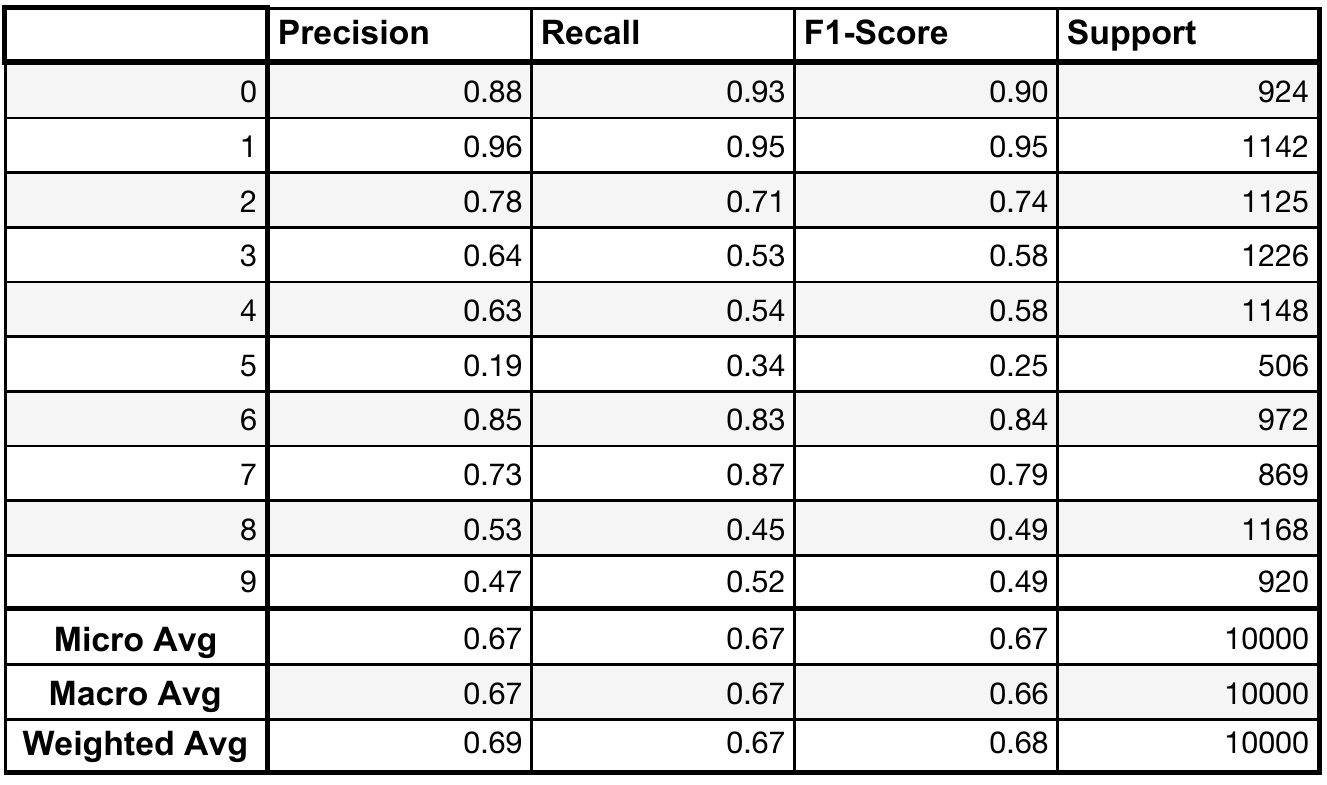}
	\caption{Confusion matrix and classification report of the neural network model with variational autoencoder after FGSM attack}
	\label{fig:NN_VAE_AE_FGSM}
\end{figure}

\begin{figure}[htbp!]
	\centering
	\includegraphics[scale=0.56]{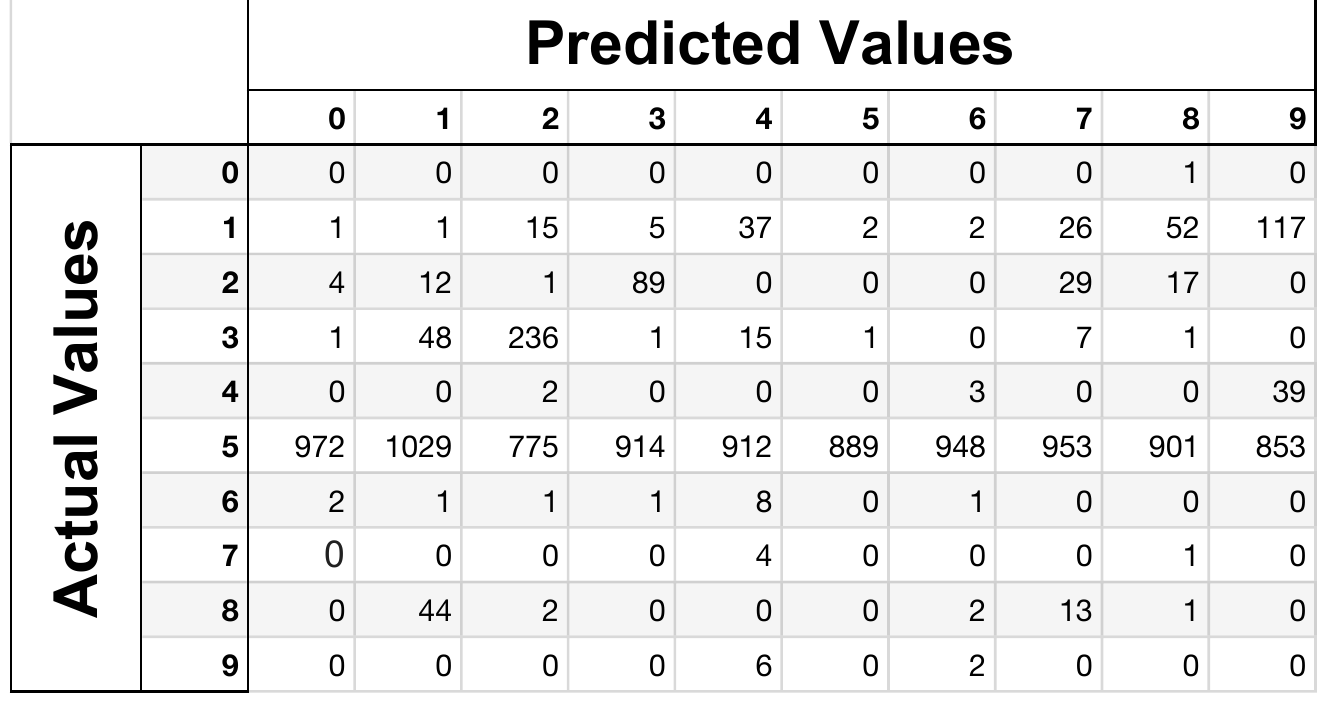}
	\includegraphics[scale=0.5]{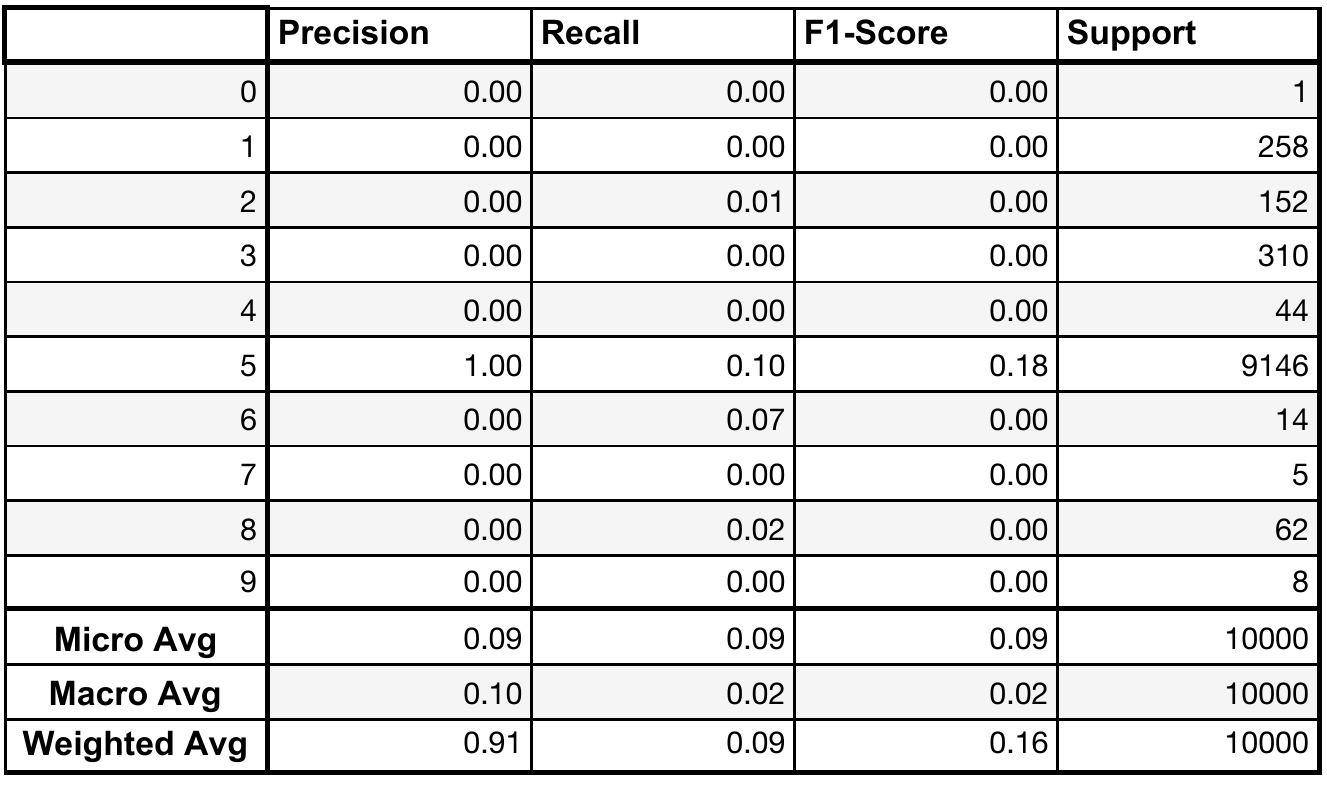}
	\caption{Confusion matrix and classification report of the neural network model without variational autoencoder after T-FGSM attack}
	\label{fig:NN_VAE_WOAE_TFGSM}
\end{figure}

\begin{figure}[htbp!]
	\centering
	\includegraphics[scale=0.56]{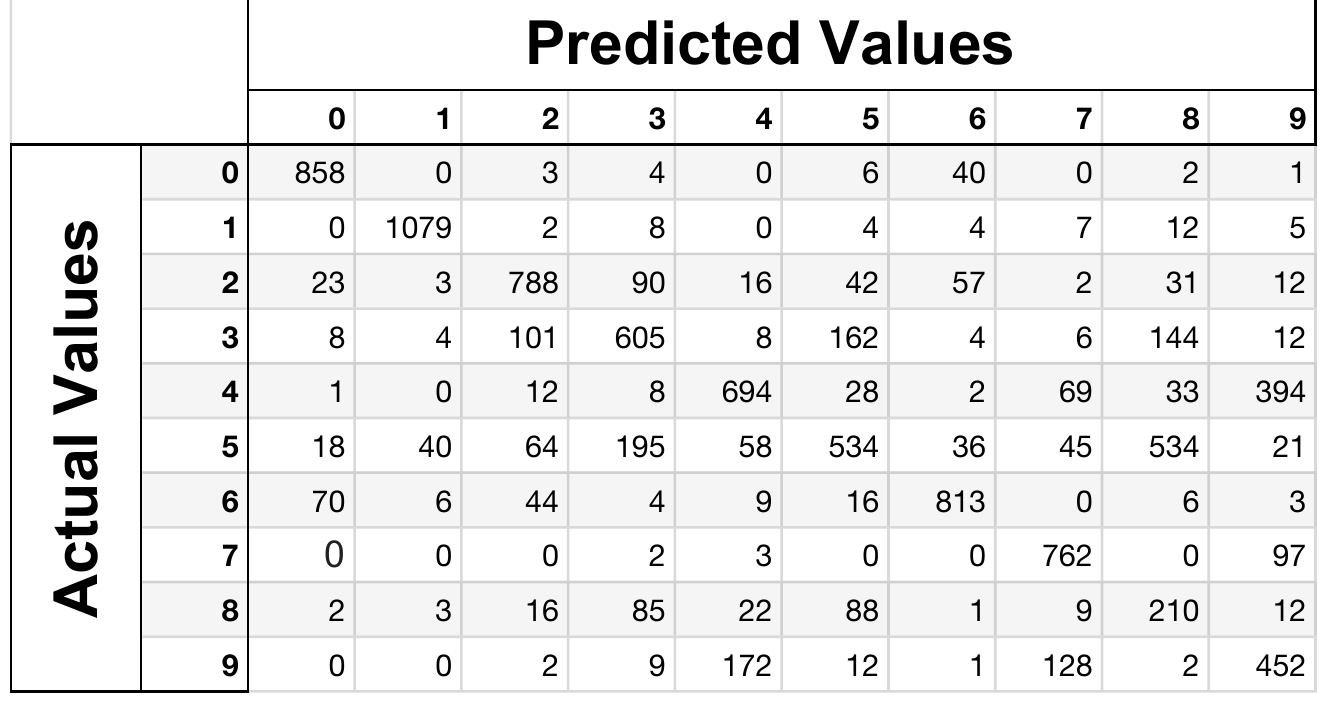}
	\includegraphics[scale=0.5]{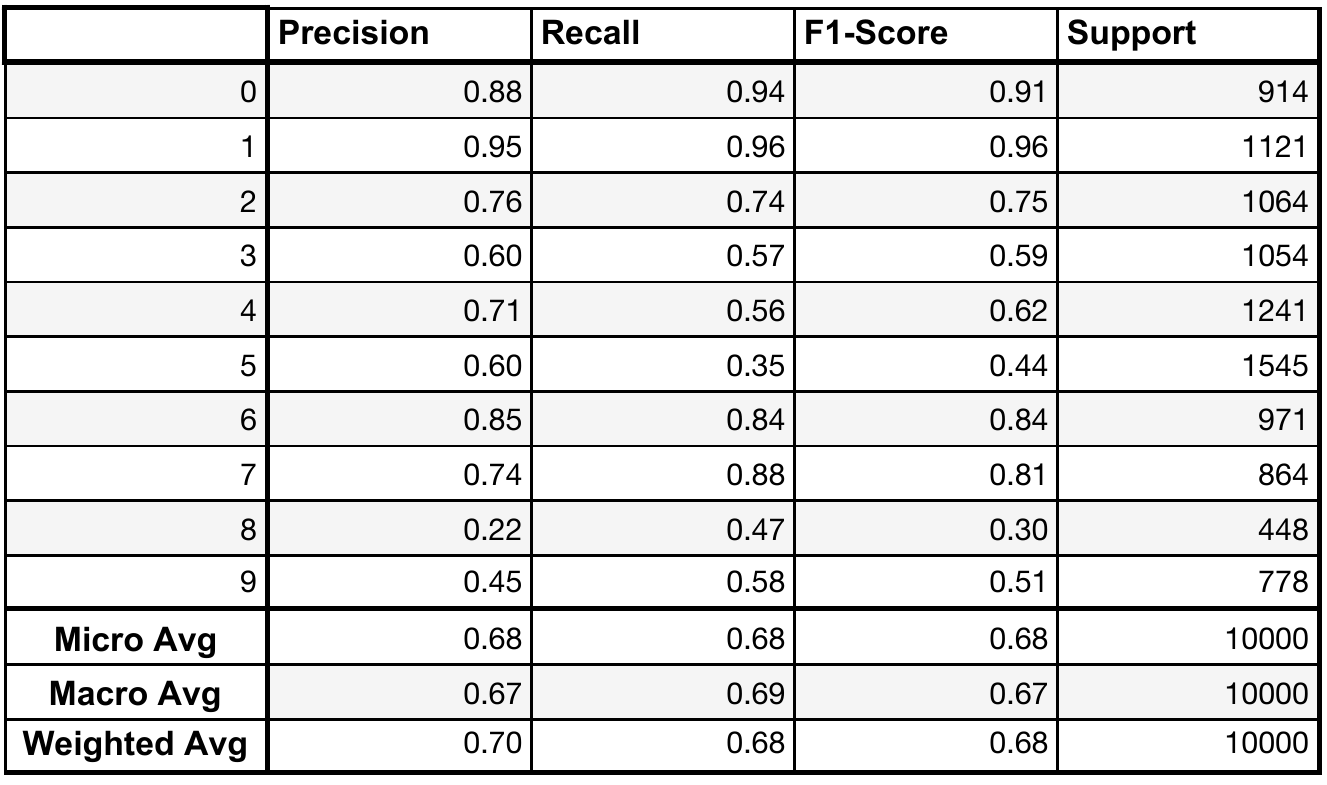}
	\caption{Confusion matrix and classification report of the neural network model with variational autoencoder after T-FGSM attack}
	\label{fig:NN_VAE_AE_TFGSM}
\end{figure}

\begin{figure}[htbp!]
	\centering
	\includegraphics[scale=0.56]{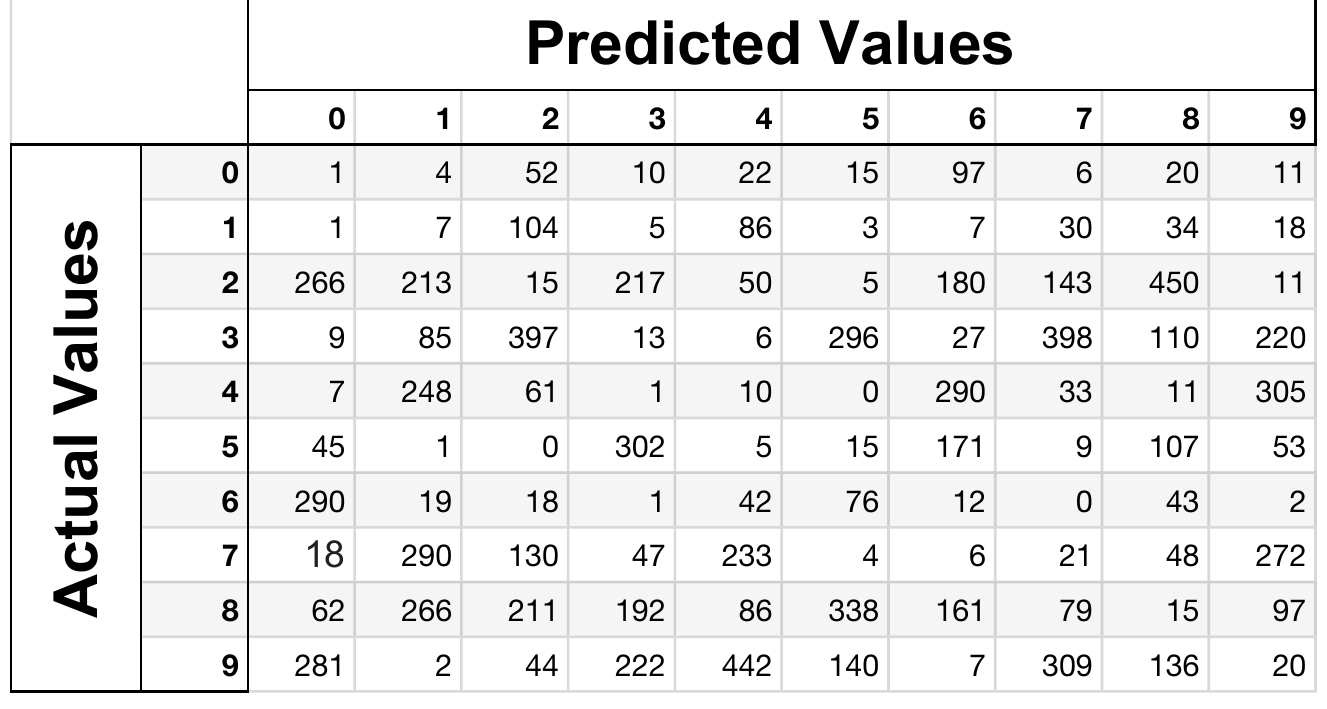}
	\includegraphics[scale=0.5]{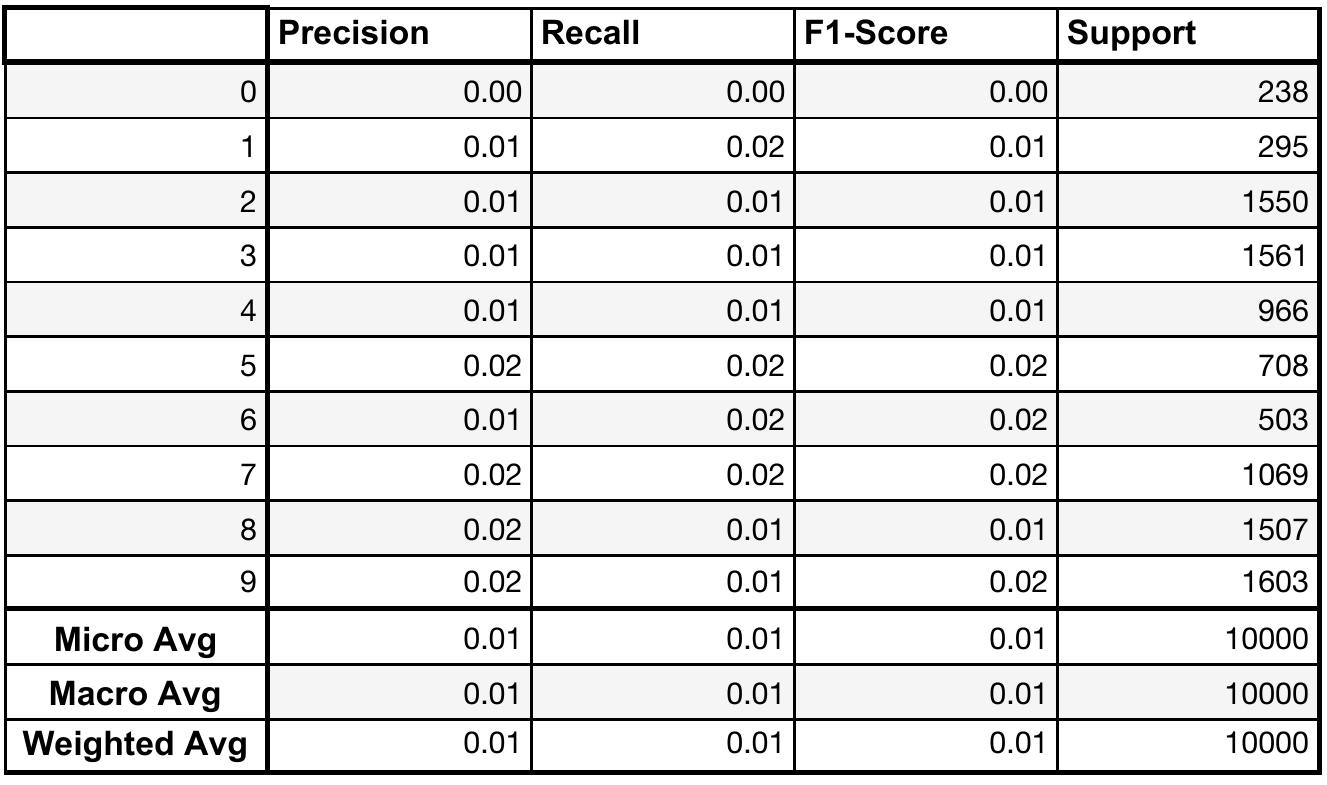}
	\caption{Confusion matrix and classification report of the neural network model without variational autoencoder after basic iterative method attack}
	\label{fig:NN_VAE_WOAE_BIM}
\end{figure}

\begin{figure}[htbp!]
	\centering
	\includegraphics[scale=0.56]{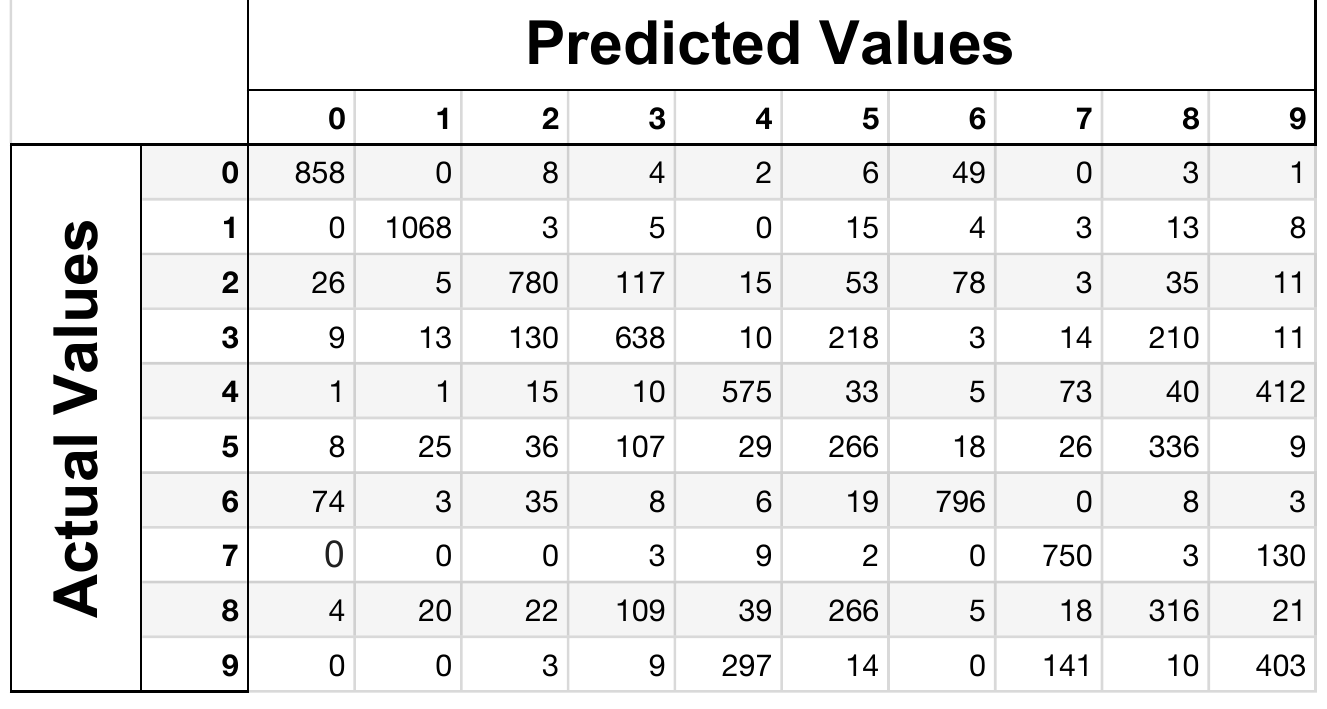}
	\includegraphics[scale=0.5]{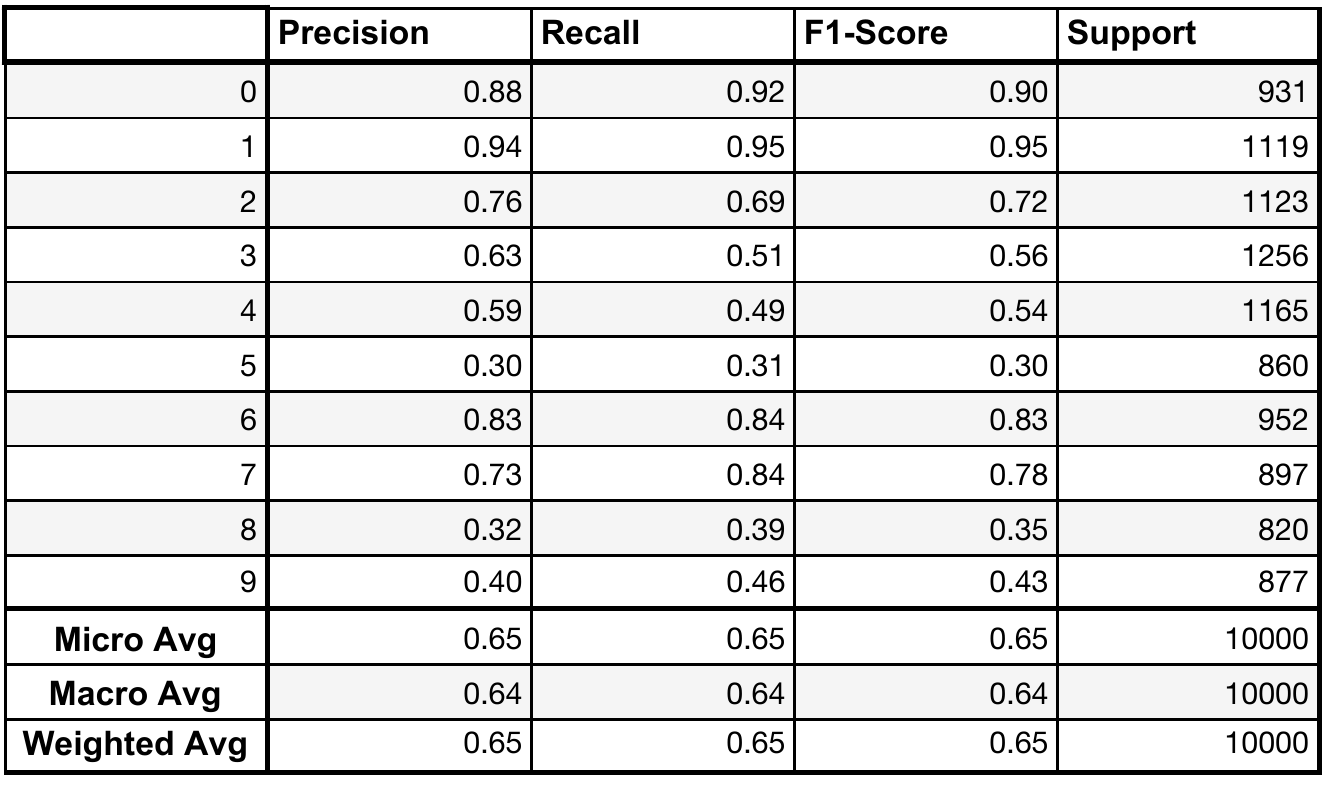}
	\caption{Confusion matrix and classification report of the neural network model with variational autoencoder after basic iterative method attack}
	\label{fig:NN_VAE_AE_BIM}
\end{figure}

\begin{figure}[htbp!]
	\centering
	\includegraphics[width=0.5\linewidth]{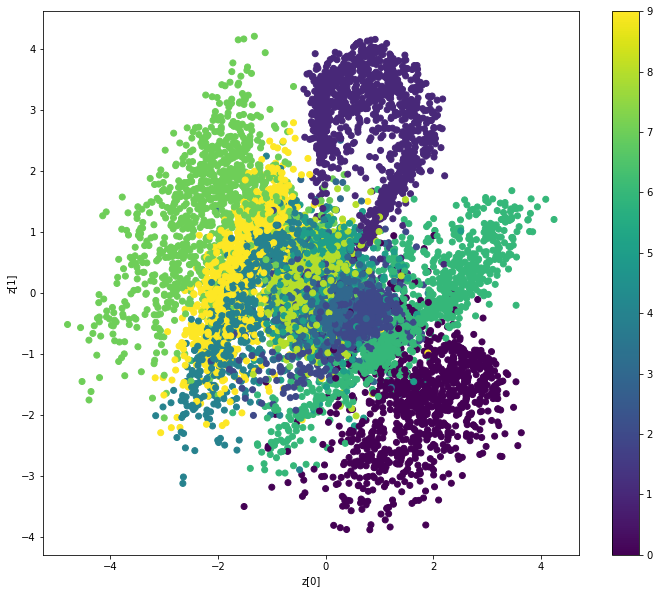}
	\caption{Because of MNIST dataset, our latent space is two-dimensional. One is to look at the neighborhoods of different classes on the latent 2D plane. Each of these colored clusters are a type of digit. Close clusters are digits that are structurally similar, they are digits that share information in the latent space.}
	\label{fig:clustersVAENN}
\end{figure}

\begin{figure}[htbp!]
	\centering
	\includegraphics[width=0.5\linewidth]{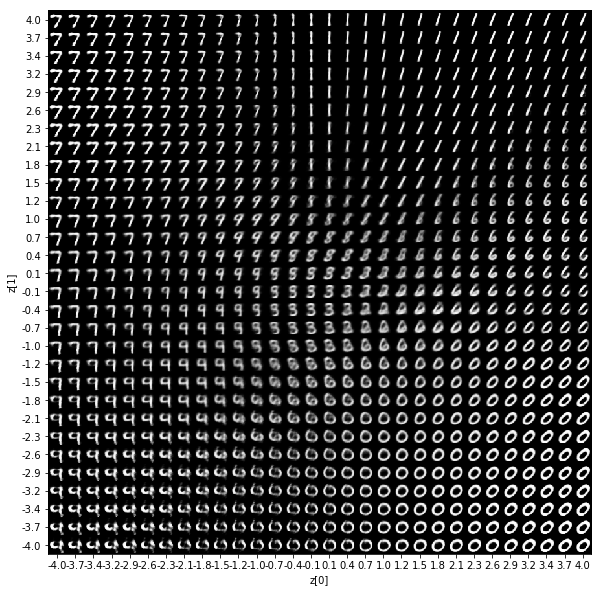}
	\caption{Due to VAE is a generative model, we can also generate new Mnist digits using latent plane, sampling latent points at regular intervals, and generating the corresponding digit for each of these points.}
	\label{fig:GANVAENN}
\end{figure}

\section{Conclusion}\label{sec:conclusion}
In this paper, we have presented the results for pre-filtering the data with an autoencoder before sending it to the machine learning model against adversarial machine learning attacks. We have investigated that the classifier accuracy changes for linear and neural network machine learning models. We have also applied non-targeted and targeted attacks to multi-class logistic regression. Besides, FGSM, T-FGSM, and BIM attacks have been applied to the neural network machine learning model. The effects of these attacks on implementing autoencoder as a filter have been analyzed for both machine learning models. We have observed that the robustness provided by autoencoder after adversarial attacks can be seen by accuracy drop between 0.1 and 0.2 percent while the models without autoencoder suffered tremendous accuracy drops hitting accuracy score between 0.6 and 0.3 in some cases even 0.1. We have proposed general, generic, and easy to implement protection against adversarial machine learning model attacks. It will be beneficial to remind that all autoencoders in this study were trained with the epoch of 35 with 1024 sized batches, so the results can be improved by increasing the number of epochs. In conclusion, we have discussed that autoencoders provide robustness against adversarial machine learning attacks to machine learning models for both linear models and neural network models. We have examined the other types of autoencoders, which are mostly called vanilla autoencoders, give the best results. The second most accurate autoencoder type is sparse autoencoders, and the third most accurate is denoising autoencoders, which gives similar results with the sparse autoencoders. We have observed that the worst autoencoder type for this process is variational autoencoders because variational autoencoders are generative models used in different areas.


In summary, the natural practice of implementing an autoencoder between data and machine learning models can provide considerable defense and robustness against attacks. These autoencoders can be easily implemented with libraries such as TensorFlow and Keras. Through the results of this review, it is evident that autoencoders can be used in any machine learning model easily because of their implementation as a separate layer.

\subsection*{Acknowledgement}
Acknowledgement text.

\bibliographystyle{ieeetr}
\bibliography{Bibliography}

\end{document}